\documentclass[lettersize,journal]{IEEEtran}
\usepackage[dvipsnames]{xcolor}
\usepackage{amsmath,amsfonts}
\usepackage{algorithmic}
\usepackage{algorithm}
\usepackage{array}
\usepackage[caption=false,font=normalsize,labelfont=sf,textfont=sf]{subfig}
\usepackage{textcomp}
\usepackage{stfloats}
\usepackage{url}
\usepackage{verbatim}
\usepackage{graphicx}
\usepackage{cite}
\usepackage{mathtools}
\usepackage{pdfpages}

\usepackage[latin1,utf8]{inputenc} 
\usepackage[T1]{fontenc}    
\usepackage{hyperref}       
\usepackage{url}            
\usepackage{booktabs}       
\usepackage{nicefrac}       
\usepackage{microtype}      
\usepackage{lipsum}
\usepackage{fancyhdr}       
\graphicspath{{media/}}     

\usepackage{amssymb}
\usepackage{nccmath}
\usepackage{multirow}
\usepackage{hhline}
\usepackage{rotating}
\usepackage{float}
\usepackage{enumitem}
\usepackage{mathrsfs}
\usepackage{mathtools}
\usepackage{nomencl}
\makenomenclature

\newcommand{\bth}{\boldsymbol{\theta}}

\newcommand{\bu}{\mathbf{u}}

\newcommand{\by}{\mathbf{y}}
\newcommand{\bx}{\mathbf{x}}
\newcommand{\bX}{\mathbf{X}}
\newcommand{\bb}{\boldsymbol{\beta}}
\newcommand{\hbb}{\hat{\boldsymbol{\beta}}}
\newcommand{\bR}{\mathbb{R}}
\newcommand{\bE}{\mathbb{E}}

\newcommand{\br}{\mathbf{r}}

\newcommand{\upperRomannumeral}[1]{\uppercase\expandafter{\romannumeral#1}}

\DeclareMathOperator*{\argmin}{argmin} 

\begin{document}

\title{The Adaptive $\tau$-Lasso:  Robustness and Oracle Properties}

\author{Emadaldin Mozafari-Majd,~\IEEEmembership{Member,~IEEE,} and Visa Koivunen,~\IEEEmembership{Fellow,~IEEE}}



\maketitle

\begin{abstract}
This paper introduces a new regularized version of the robust $\tau$-regression estimator for analyzing high-dimensional datasets subject to gross contamination in the response variables and covariates (explanatory variables). The resulting estimator, termed adaptive $\tau$-Lasso, is robust to outliers and high-leverage points. It also incorporates an adaptive $\ell_1$-norm penalty term, which enables the selection of relevant variables and reduces the bias associated with large true regression coefficients. More specifically, this adaptive $\ell_1$-norm penalty term assigns a weight to each regression coefficient. For a fixed number of predictors $p$, we show that the adaptive $\tau$-Lasso has the oracle property, ensuring both variable-selection consistency and asymptotic normality under fairly mild conditions. Asymptotic normality applies only to the entries of the regression vector corresponding to the true support, assuming knowledge of the true regression vector support. We characterize its robustness by establishing the finite-sample breakdown point and the influence function. We carry out extensive simulations and observe that the class of $\tau$-Lasso estimators exhibits robustness and reliable performance in both contaminated and uncontaminated data settings. We also validate our theoretical findings on robustness properties through simulations. In the face of outliers and high-leverage points, the adaptive $\tau$-Lasso and $\tau$-Lasso estimators achieve the best performance or match the best performances of competing regularized estimators, with minimal or no loss in terms of prediction and variable selection accuracy for almost all scenarios considered in this study. Therefore, the adaptive $\tau$-Lasso and $\tau$-Lasso estimators provide attractive tools for a variety of sparse linear regression problems, particularly in high-dimensional settings and when the data is contaminated by outliers and high-leverage points.  However, it is worth noting that no particular estimator uniformly dominates others in all considered scenarios.  

\end{abstract}

\vspace{-10 pt}
\section{Introduction}
\IEEEPARstart{T}{he} last two decades have witnessed a phenomenal surge in the collection and acquisition of large volumes of data with an increasing number of features or predictors \cite{hastie2015statistical,wainwright2019high,slavakis2014modeling}. 
Formidable challenges arise with high-dimensional models where the number of features $(p)$ exceeds the number of observations $(n)$, causing problems with uniqueness and identifiability. Such high-dimensional data often encompass a low-dimensional representation due to sparsity or a low-rank structure. Indeed, if there is no underlying structure, one may be unable to recover useful information from data about parameters of interest with a low sample size $n$. Thus, regularization of the ill-defined estimation problem may be necessary to find a unique solution and capture a low-dimensional representation of the high-dimensional data. In the sparse linear regression settings, it is common to regularize the empirical loss by the $\ell_1$-norm of the parameter vector to promote sparsity. 
 Regularizing the squared-error loss by $\ell_1$-norm gives rise to the celebrated Lasso estimator that combines variable selection with parameter estimation. While the $\ell_1$-norm penalty reduces the variance of the estimated values, it may overshrink the coefficient estimates \cite{zou2005regularization} when the true coefficients are large and produce very biased estimates. 
Adaptive Lasso \cite{zou2006adaptive} substitutes the $\ell_1$-norm penalty with an adaptive $\ell_1$-norm penalty where small weights are given to parameters associated with large true coefficients. This adjustment reduces the penalty imposed on these parameters, leading to estimates with lower bias. The modified Lasso estimator (adaptive Lasso), as described above, satisfies oracle properties, i.e., (i) the support of the estimated coefficient vector agrees with the support of the true coefficient vector in the asymptotic sense (variable selection consistency); (ii) asymptotic normality holds for the entries of the regression vector corresponding to the true support by knowing the correct support a priori.

An important issue in dealing with high-dimensional models is that the probability of observing outliers or high-leverage points may increase as the sample size and dimensionality grow together. Moreover, one must take special care when the distribution of covariates or additive noise is heavy-tailed. The regularized least-squares estimators exhibit poor performance in the presence of contamination or heavy-tailed noise as the squared-error loss is highly sensitive to outliers and high-leverage points, i.e., it is not statistically robust. In order to address this issue, one can replace the squared-error loss with a robust counterpart that grows more slowly for larger residuals. This approach gives smaller weights to data points with large residuals. Examples include absolute-error loss, Huberized loss, and $\ell_q$-based losses with $1 < q < 2$. These loss functions are robust in the face of outliers and other contaminations. There is a large body of methods using regularized \textit{M}-estimators with a convex empirical loss \cite{carrillo2016robust,fansymmetry2016,wang2007robust,khan2007robust,li2011nonconcave,lambert2011robust,lambert2016adaptive,sun2020adaptive,fan2014adaptive,bradic2011penalized,bradic2016robustness,fan2018lamm,loh2017statistical}. Despite strong theoretical guarantees, these estimators may fail to limit the influence of high-leverage points in the covariates and hence may lead to a significant decrease in performance. Several procedures have been proposed to mitigate the influence of high-leverage points, particularly in \cite{sun2020adaptive,loh2017statistical}. These procedures either down-weight the observations with outlying predictor values or truncate all predictor values via univariate winsorization, even when only one predictor in the corresponding observation is an outlier. Neither of these characteristics is desirable in high-dimensional models. Additionally, there is a collection of regularized robust linear estimators introduced in \cite{carrillo2016robust}. Two of these replace the data-discrepancy term (empirical cost term) with non-convex losses, such as $\ell_q$-based norms (with $0<q<1$) and Lorentzian-based norms of residuals, that can be tuned to grow more slowly than the absolute-error loss for larger residuals. However, they still struggle to handle high-leverage points, and in particular, $\ell_q$-norms become extremely non-smooth as $q$ approaches zero.

Motivated by this deficiency of convex empirical losses, several regularized linear regression estimators robust to high-leverage points have been developed. Regularized robust linear regression estimators, known as \textit{MM}-Lasso and adaptive \textit{MM}-Lasso, were proposed in \cite{smucler2017robust}. These estimators substitute the squared-error loss of Lasso with a non-convex loss that features a redescending score function \cite{maronna2019robust}. In addition, the authors provide a solid foundation for analyzing the behavior of these estimators by deriving the robustness properties and the asymptotic theory for consistency. Other important related works include adaptive penalized elastic net \textit{S}-estimator (adaptive PENSE) \cite{kepplinger2023robustnew} and penalized elastic net \textit{S}-estimator (PENSE) \cite{freue2019robust}, both of which demonstrate favorable robustness properties and have established the asymptotic theory for consistency.


This paper introduces the adaptive $\tau$-Lasso estimator, a low-dimensional $\tau$-estimator regularized by an adaptive $\ell_1$-norm penalty similar to adaptive lasso \cite{zou2006adaptive}. The present study extends the work of Martinez-Camara et al. \cite{martinez2015new,martinez2016regularized}, who initially proposed the $\tau$-Lasso estimator and derived its influence function. A major drawback of $\tau$-Lasso is that variable selection consistency holds under restrictive mutual incoherence assumptions on the design matrix $\bX$ present in the $\ell_1$-based methods \cite{loh2017support}. With the motivation of addressing this, we modify the $\tau$-Lasso objective function by replacing the $\ell_1$-norm penalty with an adaptive $\ell_1$-norm penalty. This modification dispenses the need for mutual incoherence conditions required by the $\tau$-Lasso estimator, thereby yielding a variable selection consistent procedure under fairly mild conditions on the loss function, the decay rate of the regularization parameter and the data distribution. Herein, we focus on regularized versions of $\tau$-estimators \cite{yohai1988high,yohai1986high} and establish robustness and asymptotic properties when data follows a sparse linear regression model. We emphasize that the asymptotic theory established in this paper is constrained to the classical fixed $p$ and diverging $n$ and holds even for very heavy-tailed errors such as Cauchy-distributed noise/errors.

We now summarize the main new contributions of the paper as follows:

\begin{itemize}
    \item The asymptotic theory for consistency of the $\tau$-Lasso estimator, including strong and root-$n$ consistency, is derived. 
    
    \item The adaptive $\tau$-Lasso estimator is introduced, and its oracle properties are established under relatively mild conditions. To do so, we first show the estimator's asymptotic normality for the entries of the regression vector corresponding to the true support when the correct support is known a priori. We then prove variable selection consistency without imposing any conditions on the moments of the error distribution. This is an advantage over its predecessor $\tau$-Lasso, which does not achieve variable consistency unless under stringent conditions on the design matrix.

    \item The global robustness properties of the adaptive $\tau$-Lasso estimator, as measured by its finite-sample breakdown, are characterized. Moreover, we provide a lower and upper bound on the finite-sample breakdown point. Subsequently, the estimator's high breakdown point is validated via computer simulations.

    \item We derive the influence function of the adaptive $\tau$-Lasso estimator. We also verify from our simulation results that the resulting influence function agrees with its finite-sample version, standardized sensitivity curve, for one-dimensional toy data.
   
    \item Extensive simulation studies are conducted where the finite-sample performance of $\tau$-Lasso and adaptive $\tau$-Lasso is compared to that of other competing state-of-the-art regularized robust linear regression estimators on synthetic and real data.

    \item The influence of pilot estimates on the adaptive $\tau$-Lasso's performance has been investigated. Additionally, we present an example encountered in practical settings, where good leverage points are found on truly irrelevant predictors, and showcase adaptive $\tau$-Lasso significantly outperforms $\tau$-Lasso in variable selection (provided in Supplemental Material). Moreover, the phenomenon of overshrinkage in $\tau$-Lasso is studied and confirmed through simulations that adaptive $\tau$-Lasso effectively remedies this issue.

\end{itemize}

\subsection{Related work} There are close connections between our adaptive $\tau$-Lasso estimator and several other high-dimensional regularized robust linear estimators: the adaptive \textit{MM}-Lasso estimator \cite{smucler2017robust}, the adaptive penalized elastic-net \textit{S}-estimator (adaptive PENSE) \cite{kepplinger2023robustnew}, and the $\tau$-Lasso estimator \cite{martinez2016regularized}, \cite{martinez2015new}, all of which deal with high-leverage points in the predictors. We highlight the similarities and differences between our adaptive $\tau$-Lasso estimator and these estimators.

\begin{itemize}
    \item Adaptive $\tau$-Lasso, $\tau$-Lasso and adaptive \textit{MM}-Lasso estimators can simultaneously achieve a high breakdown point and high normal efficiency.
    \item Moreover, the adaptive \textit{MM}-Lasso and adaptive PENSE estimators, like the adaptive $\tau$-Lasso, enjoy oracle properties for a fixed number of covariates $p$  under fairly mild conditions. This is a by-product of substituting the $\ell_1$-norm penalty with an adaptive $\ell_1$-norm penalty.
    \item While our proposed estimator enjoys oracle properties even for heavy-tailed errors under fairly mild conditions, the $\tau$-Lasso estimator suffers from the overshrinkage problem associated with the $\ell_1$-norm penalty. In fact, $\tau$-Lasso achieves variable selection consistency under restrictive mutual incoherence conditions, which is not trivial.

    \item The adaptive \textit{MM}-Lasso requires a preliminary robust estimate of the scale of the additive errors. High efficiency under normal errors can be attained if bias in \textit{M}-scale estimation remains small, which is often challenging in high-dimensional settings.
    \item The adaptive PENSE circumvents the need for a preliminary scale estimate of additive errors and achieves high-breakdown points, but it can not be tuned for high normal efficiency.
    \item In contrast, our adaptive $\tau$-Lasso achieves both high breakdown and high normal efficiency without requiring a preliminary estimate of the scale of additive errors, as it is naturally associated with an estimate of the error scale, known as $\tau$-scale.
\end{itemize}
However, it is important to acknowledge that in the high-dimensional regime, if bias associated with \textit{M}-scale of the residuals is large, the high normal efficiency of the adaptive $\tau$-Lasso may be compromised, particularly under contamination in finite-sample settings.

\subsection{Organization}
The remainder of this paper is organized as follows.  Section \ref{sec:notation} details the basic notation used throughout the paper. Section \ref{sec:data_model} describes the data model used in this article in detail. Section \ref{sec:adapative_tau} introduces the background material on the adaptive $\tau$-Lasso estimator. We then provide the main results in Section \ref{sec:consistency_oracle}, including theorems and propositions on the asymptotic properties of the $\tau$-Lasso and adaptive $\tau$-Lasso estimators. Section \ref{sec:robustness} is devoted to characterizing the robustness properties such as finite-sample breakdown point and influence function for the adaptive $\tau$-Lasso estimator. In Section \ref{sec:simulations}, we illustrate the simulation results for the $\tau$-Lasso and the adaptive $\tau$-Lasso estimators and compare the results with other competing regularized robust estimators for several scenarios. We then conclude the paper in Section \ref{sec:conclusion}. We defer full proofs and technical derivations of the theoretical results outlined in this paper to the Supplemental Material, given their lengthy and tedious nature.

\vspace{-10 pt}
\section{Notation}
\label{sec:notation}

Given the true regression parameter vector (coefficient vector) $\bb_0 \in \bR^p$, we denote by $\mathcal{A}\coloneqq \{j \in \{1,\cdots,p\} | \beta_{0,j}\neq 0\}$ the true active set corresponding to the set of predictors associated with true non-zero coefficients and $\mathcal{A}^c \coloneqq \{j \in \{1,\cdots,p\} | \beta_{0,j}=0\}$ the true inactive set corresponding to the set of predictors associated with true zero coefficients. To simplify the matters with proofs of theorems, without loss of generality, we assume that the first $k_0>0$ elements of $\bb_0$ are non-zero and thus the true active set $\mathcal{A}\coloneqq \{1,2,\cdots,k_0 \}$. We write $\bX_{\mathcal{A}}$ to denote the regression matrix whose columns are those predictors in $\mathcal{A}$. For a vector $\bb \in \bR^p$, we denote by $\bb_{\mathcal{A}}$ the first $k_0$ coefficients of $\bb$ and $\bb_{\mathcal{A}^c}$ the remaining $p-k_0$ coefficients of $\bb$. Furthermore, we note the $\ell_1$-norm as
\begin{equation}
    \|\bb\|_{\ell_1}=\sum_{j=1}^p |\beta_j|.
\end{equation}
To avoid confusion, we provide the reader with a list of notations that will be consistently used within the body of the proofs.


\nomenclature[A]{$\hbb_{\text{PT}}$}{ $\tau$-Lasso estimator.}
\nomenclature[B]{$\hbb_{\text{AT}}$}{ Adaptive $\tau$-Lasso estimator.}
\nomenclature[C]{$\hbb_{\text{T}}$}{ $\tau$-estimator.}
\nomenclature[D]{$\tilde{\bb}$}{ Pilot estimate of $\bb_0$ employed to compute the adaptive weights $w_j$.}
\nomenclature[E]{$\partial_{\bb}(f)$}{ Generalized gradient of $f$ with respect to $\bb$.}
\nomenclature[F]{$\underline{\mathcal{L}}_n(\bb)$}{ Objective function of the $\tau$-Lasso estimator evaluated at $\bb$.}
\nomenclature[G]{$\mathcal{L}_n(\bb)$}{ Objective function of the adaptive $\tau$-Lasso estimator evaluated at $\bb$.}
\nomenclature[H]{$\bx$}{ $p$-dimensional random vector of predictors.}
\nomenclature[I]{$y$}{ Random response variable.}
\nomenclature[J]{$u$}{ Random measurement noise variable.}
\nomenclature[K]{$r(\bb_0)$}{ True error, $r(\bb_0)\coloneqq y-\bx^T\bb_0=u$ (r.v.).} 
\nomenclature[L]{$\tau_n(\br(\bb))$}{ $\tau$-scale estimate of the residual vector $\br(\bb)$.}
\nomenclature[M]{$\tau(\bb_0)$}{ Population $\tau$-scale of the true error, $\tau(r(\bb_0))$, we allow a small violation of notation for simplicity.}
\nomenclature[N]{$s_n(\br(\bb))$}{ \textit{M}-scale estimate of the residual vector $\br(\bb)$.}
\nomenclature[O]{$s(\bb_0)$}{ Population \textit{M}-scale of the true error, $s(r(\bb_0))$, we allow a small violation of notation for simplicity.}
\nomenclature[P]{$\delta$}{ Tuning parameter set to a desired value for controlling the asymptotic breakdown point of the estimator.}
\nomenclature[Q]{$\delta^*$}{ Constant defining the desired asymptotic breakdown point of the estimator.}
\nomenclature[R]{$\zeta^*$}{ Constant defining the desired normal efficiency.}

\nomenclature[S]{$c_0$}{ Tuning parameter of $\rho_0(\cdot)$ adjusted to achieve the desired asymptotic breakdown point.}
\nomenclature[T]{$c_1$}{ Tuning parameter of $\rho_1(\cdot)$ adjusted to achieve the desired normal efficiency for a known $c_0$.}
\nomenclature[U]{$\underline{\lambda}_n$}{ Regularization parameter of the $\tau$-Lasso estimator, assumed to vary with sample size $n$ as required for the asymptotic analysis.}
\nomenclature[V]{$\lambda_n$}{ Regularization parameter of the adaptive $\tau$-Lasso estimator, assumed to vary with sample size $n$ as required for the asymptotic analysis.}
\nomenclature[W]{$\bE_{F}(\cdot)$}{ Expected value with respect to distribution $F$.}

\nomenclature[X]{$\theta_n=o_P(a_n)$}{ Implies that the sequence $\lVert \theta_n \rVert_{\ell_2} /\lvert a_n \rvert$ converges to zero in probability \cite{shao2003mathematical}.}
\nomenclature[Y]{$\theta_n=O_P(a_n)$}{ Implies that the sequence $\lVert \theta_n \rVert_{\ell_2} /\lvert a_n \rvert$ is \textbf{bounded in probability} (or \textbf{uniformly tight}) \cite{shao2003mathematical}.}
\nomenclature[Z]{$\mathbb{P}(\cdot)$}{ Probability of an event.}

\printnomenclature

\vspace{-5pt}
\section{Data Model}
\label{sec:data_model}
Throughout this article, we assume data follows a sparse linear regression model 

\begin{equation}
\label{eq:lin_reg}
    \by=\bX\bb_0+\bu,
\end{equation}
where $\by=[y_1,\cdots,y_n]^{T} \in \bR^n$ denotes the response vector, $\bX=[\bx_{[1]},\cdots,\bx_{[n]}]^T \in \bR^{n \times p}$ is the regression matrix and data pairs $\{(y_i,\bx_{[i]})\}_{i=1}^n$ are $n$ i.i.d. realizations of random variable $(y,\bx) \in \bR \times \bR^p$. $\bu=[u_1,\cdots,u_n]^T \in \bR^n$ is the measurement noise vector where $\{u_i \in \bR \}_{i=1}^n $ are $n$ i.i.d. realizations of random variable $u \in \bR$. The goal is to estimate the unknown sparse parameter vector $\bb_0 \in \bR^p$ based on a sample of $n$ observations $\big(\by,\bX \big)$. We assume sparsity such that $k_0 < p$ coefficients of $\bb_0$ are non-zero, and their corresponding indices and $k_0$ are not known in advance.  

Suppose the measurement errors $u_i$ follow some distribution $F$ and are independent of the explanatory variables $\bx_{[i]}$ with distribution $G$. Then, the joint distribution $H$ of $(y_i,\bx_{[i]})$ satisfy

\begin{equation}
\label{eq:dist_joint}
    H(y,\bx)=G(\bx)F(y-\bx^T \bb_0).
\end{equation}

Here, it is crucial to distinguish between $y$, $\mathbf{x}$, and $u$, which are random variables and their corresponding realizations $y_i$, $\mathbf{x}_{[i]}$, and $u_i$, which are deterministic. This distinction is of paramount importance as it significantly impacts the subsequent statistical analysis discussed in this article.
\section{The Adaptive $\tau$-Lasso}
\label{sec:adapative_tau}
\subsection{Definition}

The $\tau$-Lasso estimator \cite{martinez2015new,martinez2016regularized} is a regularized robust estimator whose objective function comprises a regularization term to deal with high-dimensional models and a robust empirical loss to deal with outliers and high-leverage points. While $\ell_1$-norm regularization of the $\tau$-Lasso promotes sparsity by setting some coefficients to zero, which is a desired property, it also severely shrinks the estimated coefficients associated with larger true coefficients. In order to remedy the overshrinkage of the $\tau$-Lasso estimates for these coefficients, one can assign properly chosen weights to different regression coefficients similarly to the adaptive Lasso \cite{zou2006adaptive}. We now introduce the adaptive $\tau$-Lasso as

\begin{equation}
\label{eq:tau_adap}
    \hbb_{\text{AT}}=\argmin_{\bb \in \bR^p} \mathcal{L}_n(\bb)=\argmin_{\bb \in \bR^p} \biggl\{\tau_n^2(\br(\bb))+\lambda_n \sum_{j=1}^{p}w_j|\beta_j|\biggl\}
\end{equation}
where $\lambda_n$ is a nonnegative regularization parameter controlling the amount of shrinkage induced by the adaptive penalty term, and adaptive weights are given by $w_j=1/|\tilde{\beta}_j|^{\gamma}$. $\tilde{\bb}$ denotes a pilot estimate 
of $\bb_0$. The choice of $\gamma >0 $ influences the variable selection consistency and asymptotic normality of the estimator and should be carefully selected to attain the desired results. We will, later on, show that if the pilot estimate $\tilde{\bb}$ is a strongly consistent estimate of $\bb_0$, the adaptive $\tau$-Lasso estimator enjoys the root-$n$ consistency and oracle properties with a proper selection of $\lambda_n$ under fairly mild conditions. Note that $\lambda_n$ varies with $n$ in the asymptotic analysis. Let $\br(\bb)=\by-\bX \bb$ be the vector of residuals, and let  $\rho_0(\cdot)$ and $\rho_1(\cdot)$ represent bounded-$\rho$-functions, as described in \textit{Assumption} $1$ in the following lines. Then $\tau_n(\br(\bb))$ is an efficient $\tau$-scale \cite{yohai1988high,yohai1986high} defined as follows:
 \begin{equation}
 \label{eq:tauscale}
     \tau_n^2(\br(\bb))=s_n^2(\br(\bb))\frac{1}{n}\sum_{i=1}^n \rho_1 \bigg(\frac{r_i(\bb)}{s_n(\br(\bb))}\bigg)
 \end{equation}
with $r_i(\bb)$'s denoting the residuals $\{y_i-\bx_{[i]}^T\bb \}$ and $s_n(\br(\bb))$ is an \textit{M}-scale estimate of residuals $\br(\bb)$ defined as the solution to
\begin{equation}
\label{eq:mscale}
    \frac{1}{n}\sum_{i=1}^n \rho_0 \bigg(\frac{r_i(\bb)}{s_n(\br(\bb))}\bigg)= \delta,
\end{equation}
where $\delta$ is tuned to control the asymptotic breakdown point of the estimator; to simplify notation, we will write $\tau_n=\tau_n(\br(\bb))$ and $s_n=s_n(\br(\bb))$. To ensure the robustness of our estimates, we need to impose several assumptions on our bounded $\rho$-functions, $\rho_0(\cdot)$ and $\rho_1(\cdot)$, as follows: \\

\textit{Assumption} 1: the $\rho$-functions in this work satisfy the following conditions \cite{maronna2019robust,kepplinger2023robustnew}

\begin{enumerate}
\item  $\rho(\cdot)$ is real, even, continuous and $\rho(0)=0$.
\item  $\rho(t)$ is bounded where $\rho(t)=1$ for $|t|\geq c$ with $0<c<\infty$, and is strictly increasing in $|t|$ elsewhere.
\item \label{R4} $\rho(t)$ is continuously differentiable with both $t\rho^{'}(t)$ and $\rho^{'}(t)$ being bounded.
\end{enumerate}

We note $\rho_i(t)=\rho(t\: ;c_i)$ where $i \in 0,1$. Detailed instructions on tuning the constants $c_0$ and $c_1$ to achieve high breakdown point and high normal efficiency are given in Section S.I of the Supplemental Material.

\textit{Assumption} 2:
\begin{equation}
    2\rho_1(t)-\psi_1(t)t \geq 0,
\end{equation}
where $\psi_1(t)=\partial \rho_1(t)/\partial t$. If the above condition holds, we can treat the $\tau$-Lasso estimator as an \textit{M}-Lasso estimator with a $\psi(t)=\overline{W} \psi_0(t) + \psi_1(t)$, which is a weighted sum of $\psi_0(t)$ and $\psi_1(t)$. Here, $\psi_0(t)$ is defined analogously to $\psi_1(t)$, and $\overline{W}$ is a scalar weight given by 
\begin{align}
  \overline{W}=\big( 2\bE[\rho_1(t)]-\bE[\psi_1(t)t] \big)/\bE[\psi_0(t)t]. \nonumber 
\end{align}

\textit{Remark 1}: Setting $w_j=1$ for $j=1,\cdots,p$ transforms the adaptive $\tau$-Lasso into the $\tau$-Lasso. We now define the $\tau$-Lasso estimator as
\begin{equation}
\label{eq:tau_lasso}
    \hbb_{\text{PT}}=\argmin_{\bb \in \bR^p}  \underline{\mathcal{L}}_n(\bb)=\argmin_{\bb \in \bR^p} \biggl\{ \tau_n^2(\br(\bb))+\underline{\lambda}_n \sum_{j=1}^{p}|\beta_j|\biggl\}.
\end{equation}
To distinguish between different levels of regularization that can be employed by the $\tau$-Lasso and the adaptive $\tau$-Lasso estimators, we use a different notation for the regularization parameter of the $\tau$-Lasso, denoted as $\underline{\lambda}_n$. Further details and background on the $\tau$-Lasso estimators can be found in Section
S.VIII of the Supplemental Material.
\subsection{Computation of the adaptive $\tau$-Lasso estimates}
One can solve the adaptive $\tau$-Lasso problem by rewriting it as a $\tau$-Lasso estimation problem as follows:
\begin{equation}
\label{eq:rewrite_atau}
    \hbb=\argmin_{\bb \in \bR^p} \biggl\{ \tau_n^2\Big( \by-\sum_{j=1}^p\underline{\bx}_{j}\beta_j\Big)+\lambda_n \|\bb\|_{\ell_1} \bigg\}
\end{equation}
where $\underline{\bx}_{j}=\bx_{j}/w_j$ and $\hat{\beta}_{\text{AT},j}=\hat{\beta}_j/w_j$. Hence, we replace the $j^{th}$ predictor $\bx_j \in \bR^n$ with its weighted counterpart $\underline{\bx}_j$ and treat the adaptive $\tau$-Lasso estimation problem as a $\tau$-Lasso estimation problem. Eventually, $\hat{\beta}_j$ are scaled by the adaptive weights, and the result equals $\hat{\beta}_{\text{AT},j}$. 

\subsection{Choice of $\rho$-function}
A popular choice of $\rho$-function in robust regression meeting the above conditions is Tukey's bisquare family of functions:
\begin{equation}
\label{eq:rho_def}
    \rho(t\: ; c)=1-\Big(1-\big(\frac{t}{c}\big)^2\Big)^3 \mathbf{1}_{|t|\leq c}
\end{equation}
with indicator function
\begin{equation*}
    \mathbf{1}_{|t|\leq c}=\begin{cases} 1, & |t|\leq c, \\
    0, & \text{otherwise},
    \end{cases}
\end{equation*}
where $c$ is a tuning parameter. This choice of $\rho$-function satisfies the conditions outlined in \textit{Assumption} $1$ for selecting an appropriate bounded $\rho$-function. While other choices of $\rho(\cdot)$, such as the Huber loss function, grow at a slower rate for large residuals compared to the squared-error loss, they introduce significant bias in the estimation and can result in the complete breakdown of the estimator, particularly in the presence of high-leverage points in predictors. Furthermore, choices like Huber loss do not satisfy the necessary conditions given in \textit{Assumption} $1$, which are essential for the correctness of theoretical guarantees provided in this work. In general, the choice of $\rho$-functions influences the robustness and efficiency of the adaptive $\tau$-Lasso estimators. Hence, the choice of $\rho$-functions also affects the estimator's variance. 

\subsection{Choice of pilot estimator}
\label{subsec:choice_pilot}
Our recommendation for the pilot estimates $\tilde{\bb}$ is to select a robust and strongly consistent estimate of $\bb_0$, ensuring that the adaptive $\tau$-Lasso demonstrates root-$n$ consistency and oracle properties with a proper choice of $\lambda_n$ under fairly mild conditions. If the pilot estimate contains zero coefficients, $\hat{\beta}_{\text{AT},j}$ is set to $0$ for $\tilde{\beta}_{j}=0$, thereby removing the associated predictors from the set of active variables. While this may significantly reduce computational complexity, the impact on variable selection depends on the quality of the pilot estimates. It might boost variable selection by discarding truly irrelevant variables, but it could also incorrectly omit truly relevant ones, thus deteriorating variable selection performance. To somewhat alleviate this drawback, one can set the zero coefficients of the pilot estimate $\tilde{\bb}$ to a very small value $\epsilon$ and adjust the weights as $w_j=1/\max(\epsilon,|\tilde{\beta}_j|)$, or alternatively use the formula presented in \cite{zou2009adaptive}. One can expect a loss of performance in variable selection and estimation if the pilot estimator is not consistent for $\bb_0$ and has a large associated estimation error. The impact of the pilot estimator on the adaptive $\tau$-Lasso predictive and variable selection performance, including the failure to identify true non-zero coefficients, failure to identify zero coefficients, and estimation bias, is studied in detail in S.IV of the Supplemental Material.

Throughout this article, we use the \textit{S}-Ridge as the pilot estimator for calculating the adaptive $\tau$-Lasso estimates. The regularization parameter of the \textit{S}-Ridge is selected via five-fold cross-validation using a $\tau$-scale of the residuals and simulations are performed using the R package \href{https://cran.r-project.org/web/packages/pense/index.html}{\texttt{pense}}. This choice is primarily motivated by the strong consistency and robustness of the \textit{S}-Ridge and its ability to produce non-sparse estimates. It avoids prematurely omitting the truly relevant variables and, in turn, results in improved variable selection. For convenience in interpretation, the $\tau$-Lasso estimator is picked as the pilot estimator only for validating theoretical results regarding the influence function.
\\


\vspace{-10 pt}
\section{Consistency and Oracle Properties}
\label{sec:consistency_oracle}

In statistics, a desirable property of a point estimator $T_n$ is consistency. We call a point estimator root-$n$ consistent, $T_n-\boldsymbol{\theta}=O_P(1/\sqrt{n})$, if the estimation error of $T_n$ converges in probability to zero at a rate of $n^{-1/2}$. In sparse linear regression, a regularized estimator is considered to have the oracle property if it fulfills two important properties. Firstly, the probability of correctly identifying the true non-zero coefficients of $\bth$ converges to 1. Secondly, we would have the same asymptotic (normal) distribution for the estimated coefficients corresponding to the non-zero entries of $\bth$ if we had applied the unregularized estimator solely to the truly active variables. In order to establish consistency and oracle properties for the class of $\tau$-Lasso estimators, we impose several additional conditions on the $\rho$-functions, the probability distribution of errors $F$, and the design matrix. We provide a detailed description along with the intuition behind these conditions in \textit{Assumption} 3 of the Supplemental Material.

This article aims to characterize the asymptotic behavior of $\tau$-Lasso and adaptive $\tau$-Lasso estimators under the above assumptions for fixed dimensionality $p$. In particular, we establish strong and root-$n$ consistency of both the $\tau$-Lasso and the adaptive $\tau$-Lasso estimators. Furthermore, we prove that the adaptive $\tau$-Lasso estimator enjoys the oracle property under fairly mild conditions. Before we proceed, it is worth acknowledging that despite the extensive effort involved, the theoretical derivations and proofs in this section largely build upon the theoretical results and machinery developed in \cite{kepplinger2023robustnew} and \cite{smucler2017robust} and to a lesser extent, \cite{yohai1988high}. Our proofs follow a similar line of reasoning and give credit to these pioneering works.


\textit{Proposition 1:} Suppose $\big(y_i,\bx_{[i]}\big), \ i=1,\cdots,n$ are i.i.d. observations with distribution $H$ given by (\ref{eq:dist_joint}). Under Assumptions $1$-$3$, except for the first and fourth conditions of Assumption $3$, if $\underline{\lambda}_n \rightarrow 0$, then the $\tau$-Lasso estimator $\hbb_{\text{PT}}$ defined by (\ref{eq:tau_lasso}) is a strongly consistent estimator of $\bb_0$,

\begin{equation}
   \hbb_{\text{PT}} \xrightarrow{a.s.} \bb_0. 
\end{equation}
\textit{Proof:} See the Supplemental Material.\\

\textit{Proposition 2:} Suppose $\big(y_i,\bx_{[i]}\big) , \ i=1,\cdots,n$ are i.i.d. observations with distribution $H$ given by (\ref{eq:dist_joint}). Under Assumptions $1$-$3$, except for the first and fourth conditions of Assumption $3$, if $\lambda_n \rightarrow 0$ and $\underline{\lambda}_n \rightarrow 0$ (in order to retain the strong consistency property for the pilot estimate $\tilde{\bb}=\hbb_{\text{PT}}$), then the adaptive $\tau$-Lasso estimator $\hbb_{\text{AT}}$ as defined by (\ref{eq:tau_adap}) is a strongly consistent estimator of $\bb_0$,

\begin{equation}
   \hbb_{\text{AT}} \xrightarrow{a.s.} \bb_0. 
\end{equation}
\textit{Proof:} See the Supplemental Material.\\

\textit{Remark 2:}
Propositions 1 and 2 ensure that $\tau$-Lasso and adaptive $\tau$-Lasso converge almost surely to the true coefficient vector $\bb_0$ under fairly mild conditions as the sample size $n$ grows. This convergence is achieved by reducing the regularization parameters at the rates specified in Propositions 1 and 2, balancing between regularization (to promote sparsity) and consistency (to accurately approximate $\bb_0$). \\

We now investigate the convergence rate of the $\tau$-Lasso and the adaptive $\tau$-Lasso estimators and prove their root-$n$ consistency. The estimation error of both the $\tau$-Lasso and the adaptive $\tau$-Lasso converges in probability to zero at a rate of $n^{-1/2}$.\\

\textit{Theorem 1:} Suppose $\big(y_i,\bx_{[i]}\big) , i=1,\cdots,n$ denote i.i.d. observations with distribution $H$ given in (\ref{eq:dist_joint}). Under Assumptions $1$-$3$, except for the first condition of Assumption $3$, if $\underline{\lambda}_n=O(1/\sqrt{n})$, then the $\tau$-Lasso estimator $\hbb_{\text{PT}}$ as defined by (\ref{eq:tau_lasso}) is a root-$n$ consistent estimator of $\bb_0$,

\begin{equation}
   \hbb_{\text{PT}}-\bb_0=O_P(1/\sqrt{n}).
\end{equation}
\textit{Proof:} Refer to the Supplemental Material.\\

\textit{Theorem 2:} Suppose $\big(y_i,\bx_{[i]}\big) , i=1,\cdots,n$ denote i.i.d. observations with distribution $H$ given by (\ref{eq:dist_joint}). Under Assumptions $1$-$3$, except for the first condition of Assumption $3$, if $\lambda_n=O(1/\sqrt{n})$ and $\underline{\lambda}_n \rightarrow 0$ (in order to retain the strong consistency property for the pilot estimate $\tilde{\bb}=\hbb_{\text{PT}}$), then the adaptive $\tau$-Lasso estimator $\hbb_{\text{AT}}$ defined by (\ref{eq:tau_adap}) is a root-$n$ consistent estimator of $\bb_0$,

\begin{equation}
    \hbb_{\text{AT}}-\bb_0=O_P(1/\sqrt{n}).
\end{equation}
\textit{Proof:} See the Supplemental Material.\\

\textit{Remark 3:}
Theorems 1 and 2 guarantee that the estimation errors $\hbb_{\text{PT}}-\bb_0$ and $\hbb_{\text{AT}}-\bb_0$ converge in probability to zero at the optimal rate of $1/\sqrt{n}$ under fairly mild conditions as the sample size $n$ grows. This convergence is achieved by diminishing the regularization parameters at appropriate rates, trading off regularization (to promote sparsity) with statistical efficiency (to accurately approximate $\bb_0$). \\
 
 \textit{Theorem 3:} Suppose $\big(y_i,\bx_{[i]}\big), i=1,\cdots,n$ denote i.i.d. observations with distribution $H$ given by (\ref{eq:dist_joint}). Under Assumptions $1$-$3$, if $\underline{\lambda}_n=O(1/\sqrt{n})$, $\lambda_n=O(1/\sqrt{n})$, and $\lambda_n n^{\gamma/2} \rightarrow \infty$, then the adaptive $\tau$-Lasso estimator $\hbb_{\text{AT}}$ defined by (\ref{eq:tau_adap}) is a variable selection consistent estimator:
\begin{equation}
    \mathbb{P}([\hbb_{\text{AT}}]_{\mathcal{A}^c}=\mathbf{0}_{p-k_0}) \rightarrow 1 \mbox{ as } n\rightarrow \infty
\end{equation}
where $\mathcal{A}^c\coloneqq \{k_0+1,\cdots,p\}$ denotes the true inactive set and $k_0$ is the number of non-zero coefficients of the true parameter vector $\bb_0$. \\

\textit{Proof:} See the Supplemental Material.\\

\textit{Remark 4:} Theorem 3 assures that, under fairly mild conditions, the adaptive $\tau$-lasso estimator correctly estimates the coefficients corresponding to true zeros as zero, with probability converging to one, as the sample size goes to infinity. In other words, it can effectively identify the true zeros, indicating a correct selection of the support. This behavior can be achieved by appropriately choosing the decay rates of the regularization parameters.\\

\textit{Theorem 4:} Suppose $\big(y_i,\bx_{[i]}\big), i=1,\cdots,n$ denote i.i.d. observations with distribution $H$ given by (\ref{eq:dist_joint}). Under Assumptions $1$-$3$, if the regularization parameter of $\tau$-Lasso $\underline{\lambda}_n=O(1/\sqrt{n})$, the regularization parameter of adaptive $\tau$-Lasso $\lambda_n=O(1/\sqrt{n})$, $\lambda_n n^{\gamma/2} \rightarrow \infty$ and $\sqrt{n} \lambda_n \rightarrow 0$, then the asymptotic distribution of adaptive $\tau$-lasso estimator for true non-zero coefficients of the parameter vector $[\hbb_{\text{AT}}]_{\mathcal{A}} \in \bR^{k_0}$ is multivariate Gaussian as follows:

\begin{align}
    &\sqrt{n}\Big([\hbb_{\text{AT}}]_{\mathcal{A}}-[\bb_{0}]_{\mathcal{A}}\Big) \nonumber \\
    &\qquad \qquad \xrightarrow{d}
    \mathcal{N}\Big(\mathbf{0}_{k_0},s^2(\bb_0) \frac{\mathbb{E}_{F}\big[ \psi^2\big(\frac{u}{s(\bb_0)}\big)\big]}{\big(\mathbb{E}_{F}\big[ \psi^{'}\big(\frac{u}{s(\bb_0)}\big)\big]\big)^2 } \mathbf{V}_{\mathcal{A}}^{-1}\Big)
\end{align}

where:
\begin{itemize}
    \item $\mathbf{V}_{\mathcal{A}}$ is the covariance matrix of truly active predictors.
    \item $\psi(\cdot)=W^0\psi_0(\cdot)+\psi_1(\cdot)$, $\psi^{'}(\cdot)=W^0\psi_0^{'}(\cdot)+\psi_1^{'}(\cdot)$, with scalar weight \begin{align}
        W^0=\frac{\bE_F\big[2\rho_1\left(\frac{u}{s(\bb_0)}\right)-\psi_1\left(\frac{u}{s(\bb_0)}\right)\frac{u}{s(\bb_0)}\big]}{\bE_F\big[\psi_0\left(\frac{u}{s(\bb_0)}\right)\frac{u}{s(\bb_0)}\big]}.\nonumber
    \end{align}
    \item $s(\bb_0)$ denotes the population \textit{M}-scale of true error defined as follows:
\begin{equation}
    s(\bb_0)=\inf \{s>0:\mathbb{E}_{F}\big[ \rho_0\big(u/s\big)\big] \leq \delta \}.
\end{equation} 
\end{itemize}

\textit{Proof:} See the Supplemental Material.\\

\textit{Remark 5:} Theorem 4 ensures that under relatively mild conditions, the adaptive $\tau$-Lasso estimator has the same asymptotic distribution for the coefficient estimates of the true non-zero components as the unregularized $\tau$-estimator applied only to data with predictors corresponding to true non-zero coefficients, provided that regularization parameters decay at appropriate rates specified in Theorem 4. The efficiency under normal errors is controlled by the fraction term in the covariance matrix, which reflects the influences of the error distribution and $\psi$-function.\\
\section{Robustness}
\label{sec:robustness}

This section focuses on studying the \textit{statistical robustness} for the adaptive $\tau$-Lasso estimates. In particular, we establish the local and global robustness properties of the adaptive $\tau$-Lasso estimator. As for global robustness, we analyze the finite-sample breakdown point measuring the largest fraction of arbitrarily contaminated observations (outliers and high-leverage points) that can be introduced into the sample without causing an arbitrarily large maximum bias in the estimator. This definition can be generalized to estimators taking values in a bounded parameter set (space), where it quantifies the maximum fraction of contaminated observations that can be tolerated without causing the estimates to leave the parameter set (space), that is, to remain bounded away from its boundary. Moreover, we assess the local properties of robustness for the adaptive $\tau$-Lasso estimator via the influence function, which measures the influence of infinitesimal contamination on the asymptotic value of the estimator.

\subsection{Finite-sample breakdown point}
Let $\mathbf{Z}$ be a collection of $n$ observations consisting of response values $y_i$ and the associated vector of predictors $\bx_{[i]}$. The replacement finite-sample breakdown point $\varepsilon^{*}(T_n;\mathbf{Z})$ of a regression estimator $T_n$ is defined as follows \cite{zoubir2018robust,maronna2019robust}:
 
\begin{equation}
    \varepsilon^{*}(T_n;\mathbf{Z})=\max_m \{ \frac{m}{n}: \sup_{\mathbf{Z}_m \in \mathcal{Z}_m} \|T_n(\mathbf{Z}_m)\|_{\ell_2} < \infty \}
\end{equation}
 where the set $\mathcal{Z}_m$ includes all possible datasets $\mathbf{Z}_m$ generated by replacing $m$ ($0 \leq m<n$) out of $n$ observations in $\mathbf{Z}$ with arbitrary values. $T_n(\mathbf{Z}_m)$ denotes the regression estimator based on the contaminated dataset $\mathbf{Z}_m$. Note that the bounded supremum of the $\ell_2$-norm term in the above definition implies a bounded maximum bias. The following theorem aims to characterize the global robustness properties of the adaptive $\tau$-Lasso estimator via the concept of finite-sample breakdown point. Before proceeding, it's worth mentioning that the proofs concerning the finite-sample breakdown point follow to some extent a similar line of reasoning as those of Theorem 1 in \cite{mozafari2022two}, Theorem 1 in \cite{freue2019robust}. Additionally, they also leverage results on the connection between $\tau$-scale  and \textit{M}-scale from \cite{yohai1986high}. \\
 
\textit{Theorem 5:} Let $m(\delta)$ represent the largest integer smaller than $n \min (\delta,1-\delta)$ for a dataset $\mathbf{Z}=\big(\by,\bX \big) \in \bR^{n \times (p+1)}$. Furthermore, $\delta$ determines the estimator's asymptotic breakdown point as defined by equation (\ref{eq:mscale}). Under certain regularity conditions, the asymptotic breakdown point essentially equals the limit of the finite-sample breakdown as $n$ approaches $\infty$. $\tilde{\bb}$ denotes a pilot estimate of $\bb_0$ obtained through the $\tau$-Lasso estimator. Then, the finite-sample breakdown point of the adaptive $\tau$-Lasso estimator retains the finite-sample breakdown point of the $\tau$-Lasso estimator as follows:

\begin{equation}
    \frac{m(\delta)}{n} \leq \varepsilon^{*}(\hbb_{\text{AT}};\mathbf{Z}) \leq \delta
\end{equation}
where $\hbb_{\text{AT}}$ denotes the adaptive $\tau$-Lasso estimator and $\tilde{\bb}=\hbb_{\text{PT}}$ . \\

\textit{Proof:} The proof is provided in the Supplemental Material.\\

\textit{Remark 6:} Theorem 5 guarantees that the finite-sample breakdown point of adaptive $\tau$-Lasso is at least as high as that of the unregularized $\tau$-estimator. The parameter $\delta$ controls the asymptotic breakdown point of the estimator and is associated with \textit{M}-scale of residuals. Since the finite-sample breakdown point of the unregularized $\tau$-estimator is given by $\lfloor n \min(\delta,1-\delta) \rfloor /n$. This property allows for the calibration of robustness by tuning $\delta$, which makes it particularly attractive. Specifically, setting $\delta=0.5$, the estimator achieves the desirable property of maximal robustness in the face of outliers. In practical terms, this means that as long as less than half of the observations are arbitrarily contaminated, the estimator remains robust and bounded away from the boundary of the parameter space.

\subsection{Influence function (IF)}

Before proceeding with the influence function derivation, we provide a brief introduction to \textit{statistical functional} required to derive the influence function. A \textit{statistical functional} $T:\mathcal{H} \mapsto \boldsymbol{\Theta}$ is defined as a mapping from a distribution space $\mathcal{H}$ into the parameter space $\boldsymbol{\Theta}$, which is an open subset of $\bR^d$ (in our case, $d=p+1$).  We will denote by $\boldsymbol{\theta}_{\infty}$ the asymptotic value of the estimator, which is a functional of the underlying distribution $H$, i.e., $\boldsymbol{\theta}_{\infty}=T(H)$. Let $\mathbf{Z} $ be a sample of $n$ observations $\{\mathbf{z}_i \in \mathscr{Z} | i=1,\hdots,n\}$ drawn from $H$; we can approximate the underlying distribution $H$ by the empirical distribution $H_n$. Hence, we define the estimator $\hat{\boldsymbol{\theta}}=T(H_n)$ as a surrogate for the asymptotic value $\boldsymbol{\theta}_{\infty}=T(H)$.

\subsubsection{Definition} In robust statistics, the influence function provides a theoretical framework that allows us to study the local robustness properties of estimators. Consider a  statistical functional $T(H)$ that is G\^{a}teaux differentiable \cite{van2000asymptotic}, we define the \textit{influence function} of $T(H)$ at point $\mathbf{z}_0 \in \mathscr{Z}_0$  for a distribution $H \in \mathcal{H}$ as 
\begin{equation}
\label{eq:IF_def}
\text{IF}(\mathbf{z}_0\;;H,T)=\frac{dT(H+\epsilon(\Delta_{\mathbf{z}_0}-H))}{d\epsilon}|_{\epsilon=0}
\end{equation}
where $\Delta_{\mathbf{z}_0}$ denotes a point mass with probability one at $\mathbf{z}_0$ and zero elsewhere.

In modern statistics, we often deal with non-differentiable regularized estimators. In order to derive the influence functions of such regularized estimators, a new framework that allows us to cope with non-differentiable risk functions is required. Avella-Medina \cite{avella2017influence} developed a rigorous framework for two-stage non-differentiable regularized \textit{M}-estimators, which defines the influence function as the limiting influence function of approximating estimators. Nonetheless, defining the general regularized \textit{M}-estimators is instructive before addressing the influence function of adaptive $\tau$-Lasso estimators. 

\subsubsection{Regularized M-estimators}
\label{subsubsec:regularized-M}
Suppose $\bE_{H_n}[\mathcal{F}(\mathbf{z},\boldsymbol{\theta})]$ (the data discrepancy term) measures the fit between a parameter vector $\boldsymbol{\theta} \in \bR^{p+1}$ and observations, and $q(\boldsymbol{\theta}\;;\lambda)$ denotes a penalty term with regularization parameter $\lambda$. We then call any estimator $T(H_n)$ satisfying the implicit equation 
\begin{equation}
\label{eq:es_eq_rlasso}
    \Bigg[\bE_{H_n}[\Psi(\mathbf{z},\boldsymbol{\theta})]+\frac{\partial q(\boldsymbol{\theta}\;;\lambda)}{\partial \boldsymbol{\theta} } \Bigg]_{\boldsymbol{\theta}=T(H_n)}=\mathbf{0}
\end{equation}
a {regularized \textit{M}-estimator}. The notation $\Psi(\mathbf{z},\boldsymbol{\theta}) \in \bR^{p+1}$ stands for the gradient of $\mathcal{F}(\mathbf{z},\boldsymbol{\theta}) \in \bR$ with respect to $\boldsymbol{\theta}$. Moreover, the notation $q^{'}(\boldsymbol{\theta}\;;\lambda)$ may be used interchangeably with $\frac{\partial q(\boldsymbol{\theta}\;;\lambda)}{\partial \boldsymbol{\theta} }$ to represent the derivative of $q(\boldsymbol{\theta}\;;\lambda)$ with respect to $\boldsymbol{\theta}$.

\textit{Remark 7}: Throughout the article, we will use both the estimator $\hat{\boldsymbol{\theta}}$ and its functional representation $T(H_n)$ interchangeably for the same estimator.

\subsubsection{Necessary mathematical notations}
All necessary notations are provided to understand and follow the influence function derivation fully.
\begin{itemize}
    \item We denote by $\boldsymbol{\theta}_{\infty}=T(H)$ a functional of the underlying distribution $H$, which represents the asymptotic value of an adaptive $\tau$-Lasso estimator in the standard form of a regularized \textit{M}-estimator, as follows:
    \begin{equation} \label{eq:TA-functional}
    T(H)=\begin{bmatrix}
     S(H) \\ T_{\bb}(H) 
    \end{bmatrix}=\begin{bmatrix}
     s_{\infty}\\ \bb_{\infty}
    \end{bmatrix}.
    \end{equation}
    \item We define $S(H)$ as a functional of the underlying distribution $H$. This represents the asymptotic value of the \textit{M}-scale estimator of the residual vector obtained through the adaptive $\tau$-Lasso, denoted as $s_{\infty}$.
    \item We use $T_{\bb}(H)$ to represent a functional of the underlying distribution $H$. This functional corresponds to the asymptotic value of the adaptive $\tau$-Lasso estimator for the regression parameter vector, denoted as $\bb_{\infty}=[\beta_{1,\infty},\cdots,\beta_{p,\infty}]^T$.
    \item We define $\bth$ by augmenting the scale parameter $s_{\sigma}$ with the vector of regression parameters $\bb$,
    
    \begin{equation}
        \bth=\begin{bmatrix}
     s_{\sigma}\\ \bb
    \end{bmatrix},
    \end{equation}
    
   where $s_{\sigma}$ and $\bb$ serve as optimization variables and parameterize the regularized \textit{M}-estimator model.
    
    \item We denote by $\tilde{r}(\bth)$ the standardized residual parameterized by $\bth$, such that $ \tilde{r}(\bth)=(y-\bx^T\bb)/s_{\sigma}$.
    \item $k_s$ denotes the number of non-zero entries in the asymptotic value of the regression estimates for the parameter vector, represented by $\bb_{\infty}$.
\end{itemize}

\textit{Remark 8}: The regularized \textit{M}-estimator formulation of the $\tau$-Lasso involves the notations $\underline{\bth}_{\infty}$, $\underline{s}_{\infty}$, and $\underline{\bb}_{\infty}$,  along with their functional representations $\underline{T}(H)$, $\underline{S}(H)$, and $\underline{T}_{\bb}(H)$, which correspond to their respective counterparts used in the adaptive $\tau$-Lasso estimator where the same relations apply with different notations. Specifically, $\underline{T}(H)$ can be defined as an augmented vector, given by
\begin{equation} \label{eq:T-functional}
    \underline{T}(H)=\begin{bmatrix}
     \underline{S}(H) \\ \underline{T}_{\bb}(H) 
    \end{bmatrix}=\begin{bmatrix}
     \underline{s}_{\infty}\\ \underline{\bb}_{\infty}
    \end{bmatrix}.
    \end{equation}
    
\textit{Remark 9}: Likewise, the notations $\underline{\bth}$, $\underline{s}_{\sigma}$, $\underline{\bb}$, and $\underline{k}_s$ in the regularized \textit{M}-estimator formulation of the $\tau$-Lasso are analogous to the corresponding notations used in the adaptive $\tau$-Lasso counterparts. In this context, we use the underscore to indicate $\tau$-Lasso estimator notations.

\textit{Remark 10}: We define $\underline{\bth}$ by augmenting $\underline{s}_{\sigma}$ with $\underline{\bb}$, in a similar manner as done for $\bth$.

\textit{Remark 11}: For convenience, we may assume without loss of generality that both $\bb_{\infty}$ and $\underline{\bb}_{\infty}$ are sparse vectors, with only the first $k_s$ and $\underline{k}_s$ entries being non-zero, respectively, and all the remaining entries are zero.

 \subsubsection{Theorems on the influence function of adaptive $\tau$-Lasso estimator}
Before stating the theorems, it should be noted that to derive the influence function of adaptive $\tau$-Lasso estimators, we shall express it in the standard form of two-stage regularized \textit{M}-estimators and then transform it into the population version. We will then calculate the influence function by leveraging the existing tools and results from Proposition 2 and 5 of \cite{avella2017influence}. For more information on this procedure, please see the Supplemental Material. We now derive the influence function of the $\tau$-Lasso estimator. Later in this subsection, we will also derive the influence function of the adaptive $\tau$-Lasso estimator. 

\textit{Theorem 6 (Influence Function of $\tau$-Lasso)}: Consider we are given a joint distribution $H(y,\bx)$, defined by equation (\ref{eq:dist_joint}), from which observations are generated. Let $\underline{T}(H)$ be a population $\tau$-Lasso estimate of the column vector $\bth_0 \coloneqq (s(\bb_0),\bb_0) \in \bR^{(p+1)}$, with $\underline{k}_s$ non-zero entries in $\underline{\bb}_{\infty}$ as in equation (\ref{eq:T-functional}). Then under the assumptions stated in \cite{avella2017influence} for the one-stage regularized \textit{M}-estimators, the influence function of the $\tau$-Lasso estimator $\underline{T}(H)$ at $\mathbf{z}_0=(y_0,\bx_{[0]})$ has the following form:

\begin{align}
\label{eq:Theorem_6_IF_tau}
        \text{IF}(\mathbf{z}_0\;;H,\underline{T})=-\begin{bmatrix}
     M^{-1} & \mathbf{0}_{(\underline{k}_s+1)\times(p-\underline{k}_s)}\nonumber\\
     \mathbf{0}_{(p-\underline{k}_s)\times(\underline{k}_s+1)} & \mathbf{0}_{(p-\underline{k}_s)\times (p-\underline{k}_s)}
    \end{bmatrix}  \nonumber\\ \times
    \Big(\Psi(\mathbf{z}_0,\underline{T}(H))+\underline{q}^{'} (\underline{T}(H)\;;\underline{\lambda}_n)\Big) 
\end{align}
where 
\begin{subequations}
\begin{align}
 \Psi(\mathbf{z}_0,\underline{T}(H))&=\begin{bmatrix}
 \rho_0\Big(\frac{y_0-\bx_{[0]}^T \underline{\bb}_{\infty}}{´\underline{s}_{\infty}}\Big)-\delta \\
     -\psi \Big(\frac{y_0-\bx_{[0]}^T \underline{\bb}_{\infty}}{´\underline{s}_{\infty}}\Big)\bx \underline{s}_{\infty} 
\end{bmatrix},\\
     \underline{q}^{'} (\underline{T}(H)\;;\underline{\lambda}_n)&=\begin{bmatrix}
      0
      \\
      \underline{\lambda}_n\text{sgn}(\underline{\bb}_{\infty})
     \end{bmatrix}, \quad \text{and}\\
      M&=\begin{bmatrix}
     \overbrace{M_{11}}^{\text{ scalar}} & \overbrace{M_{12}}^{(1 \times \underline{k}_s)\text{ row vector}}\\
      \underbrace{M_{21}}_{(\underline{k}_s \times 1)\text{ column vector}} &  \underbrace{M_{22}}_{(\underline{k}_s \times \underline{k}_s)\text{ matrix}}
    \end{bmatrix},
\end{align}
\end{subequations}
 with 
 \begin{subequations}
 \begin{align}
 M_{11}&=-\frac{1}{\underline{s}_{\infty}}\bE_H[\psi_0(\tilde{r}(\underline{T}(H)))\tilde{r}(\underline{T}(H))],\\
     M_{12}&=-\frac{1}{\underline{s}_{\infty}} \bE_H[\psi_0(\tilde{r}(\underline{T}(H)))\bx_{\underline{\Gamma}}^T], \\
     M_{21}&=- \bE_H[(\underline{s}_{\infty} \frac{\partial \psi(\tilde{r}(\underline{T}(H)))}{\partial \underline{s}_{\sigma}} +\psi(\tilde{r}(\underline{T}(H))))\bx_{\underline{\Gamma}} ],
     \end{align}
 \end{subequations}
  and $M_{22}$ referring to a $\underline{k}_s\times \underline{k}_s$ submatrix of $-(\bE_H[\bx \underline{s}_{\infty}\partial \psi(\tilde{r}(\underline{T}(H)))/\partial \underline{\bb}])$ indexed by the set $\underline{\Upsilon}=\{1,\cdots,\underline{k}_s\} \times \{1,\cdots,\underline{k}_s\} $. $\bx_{\underline{\Gamma}}$ denotes a subvector of elements indexed by $\underline{\Gamma}=\{1,\cdots,\underline{k}_s\}$. Moreover, $M$ reflects the impact of data-generating distribution on the influence function and remains unchanged by $\mathbf{z}_0$.
 \\
 
 \textit{Proof:} See the Supplemental Material.
 
 \textit{Theorem 7 (Influence Function of Adaptive $\tau$-Lasso):} Consider we are given a joint distribution $H(y,\bx)$, defined by equation (\ref{eq:dist_joint}), from which observations are generated. Let $\underline{T}(H)$ be a pilot $\tau$-Lasso estimate of $\boldsymbol{\theta}_0$, with $\underline{k}_s$ non-zero entries in $\underline{\bb}_{\infty}$ as given in equation (\ref{eq:T-functional}), and suppose that we denote by $T(H)$ an adaptive $\tau$-Lasso estimate of $\boldsymbol{\theta}_0$, with $k_s$ non-zero entries in $\bb_{\infty}$ as given in equation (\ref{eq:TA-functional}). Then under the assumptions stated in \cite{avella2017influence} for the two-stage regularized \textit{M}-estimators, the influence function of the adaptive $\tau$-Lasso estimator $T(H)$ at $\mathbf{z}_0=(y_0,\bx_{[0]})$ has the following form:

\begin{align} \label{eq:IF-theorem7-adaptive-tau}
        \text{IF}(\mathbf{z}_0\;;H,T)=-&\begin{bmatrix}
     N^{-1} & \mathbf{0}_{(k_s+1)\times(p-k_s)}\nonumber \\
     \mathbf{0}_{(p-k_s) \times (k_s+1)} & \mathbf{0}_{(p-k_s)\times (p-k_s)}
    \end{bmatrix}  \nonumber \\
   &\times \Bigg(\Psi(\mathbf{z}_0,T(H))+q^{'} (T(H),\underline{T}(H)\;;\lambda_n)\nonumber\\
      &- \textbf{diag}\Big(\boldsymbol{\Phi},\mathbf{0}_{p-\underline{k}_s}\Big)\times \text{IF}(\mathbf{z}_0\;;H,\underline{T})\Bigg) 
\end{align}
where 
\begin{subequations}
\begin{align}
 \Psi(\mathbf{z}_0,T(H))&=\begin{bmatrix}
 \rho_0\Big(\frac{y_0-\bx_{[0]}^T \bb_{\infty}}{s_{\infty}}\Big)-\delta \\
     -\psi \Big(\frac{y_0-\bx_{[0]}^T \bb_{\infty}}{s_{\infty}}\Big)\bx s_{\infty} 
\end{bmatrix},\\
      \boldsymbol{\Phi}&=\begin{bmatrix}
      0 
      \\
    \lambda_n \frac{\text{sgn}(\beta_{1,\infty})\text{sgn}(\underline{\beta}_{1,\infty})}{|\underline{\beta}_{1,\infty}|^2}
    \\
    \vdots
    \\
        \lambda_n \frac{\text{sgn}(\beta_{\underline{k}_s,\infty})\text{sgn}(\underline{\beta}_{\underline{k}_s,\infty})}{|\underline{\beta}_{\underline{k}_s,\infty}|^2}
     \end{bmatrix},  \\
     q^{'} (T(H),\underline{T}(H)\;;\lambda_n)&=\begin{bmatrix}
      0
      \\
      \lambda_n\frac{\text{sgn}(\beta_{1,\infty})}{|\underline{\beta}_{1,\infty}|}
      \\
      \vdots
      \\
    \lambda_n\frac{\text{sgn}(\beta_{\underline{k}_s,\infty})}{|\underline{\beta}_{\underline{k}_s,\infty}|}\\ \mathbf{0}_{p-\underline{k}_s}
     \end{bmatrix}, \quad \text{and}\\
     N&=\begin{bmatrix}
          \overbrace{N_{11}}^{\text{ scalar}} & \overbrace{N_{12}}^{(1 \times k_s)\text{ row vector}}\\
      \underbrace{N_{21}}_{(k_s \times 1)\text{ column vector}} &  \underbrace{N_{22}}_{(k_s \times k_s)\text{ matrix}}
    \end{bmatrix},
\end{align}
\end{subequations}
with
\begin{subequations}
\begin{align}
    N_{11}&=-\frac{1}{s_{\infty}}\bE_H[\psi_0(\tilde{r}(T(H)))\tilde{r}(T(H))], \\
    N_{12}&=-\frac{1}{s_{\infty}} \bE_H[\psi_0(\tilde{r}(T(H)))\bx_{\Gamma}^T],\\
    N_{21}&=- \bE_H[(s_{\infty} \frac{\partial \psi(\tilde{r}(T(H)))}{\partial s_{\sigma}} +\psi(\tilde{r}(T(H))))\bx_{\Gamma} ],
\end{align}  
\end{subequations}
and $N_{22}$ referring to a $k_s\times k_s$ submatrix of $-(\bE_H[\bx s_{\infty}\partial \psi(\tilde{r}(T(H)))/\partial \bb])$ indexed by the set $\Upsilon=\{1,\cdots,k_s\} \times \{1,\cdots,k_s\} $. $\bx_{\Gamma}$ denotes a subvector of elements indexed by $\Gamma=\{1,\cdots,k_s\}$. Moreover, $N$ captures the impact of data-generating distribution on the influence function and remains unchanged by $\mathbf{z}_0$.
\\

\textit{Proof:} See the Supplemental Material.\\

\textit{Remark 12:} The influence functions derived in Theorems 6 and 7 quantify the effect of an infinitesimal $\epsilon$ (i.e., extremely small) perturbation applied to the data distribution on the (asymptotic) values of the estimators. Consequently, they serve as valuable tools for assessing the relative influence of an individual observation on the estimator's value, giving an insight into the local stability of the estimators. Since both adaptive $\tau$-Lasso and $\tau$-Lasso have bounded influence functions, no single observation can lead to arbitrarily large changes in estimates. Therefore, these influence functions can be used to downweight or reject outliers, identify influential observations and outliers, as well as quantify the extent to which the estimators' values would change if some observations were removed or added.

\textit{Remark 13:} In order to compare the local robustness of $\tau$-Lasso and adaptive $\tau$-Lasso via their influence functions, one should first choose a norm such as $\ell_2$-norm or $\ell_{\infty}$-norm to quantify the magnitude of these influence functions for a meaningful comparison. Since the adaptive $\tau$-Lasso utilizes the pilot estimator (e.g., $\tau$-Lasso), its influence function is partly driven by the influence function of the pilot estimator and the chosen family of $\rho$-functions along with the selection of tuning constants (which also govern the breakdown point and normal efficiency). Consequently, determining which estimator is more robust in terms of the influence function involves evaluating the IF of both estimators over a range of $\mathbf{z}_0$ values. However, since both estimators possess bounded influence functions, they both exhibit local robustness.

\vspace{-7 pt}
\section{Simulation results}
\label{sec:simulations}

In this section, we conduct extensive simulations to verify the derived robustness analytical results and compare the finite-sample performance of the $\tau$-Lasso and the adaptive $\tau$-Lasso estimators with other state-of-the-art robust and non-robust linear regression estimators using $\ell_1$-norm regularization, as well as oracle estimators applied to the relevant variables only. To do so, we evaluate the estimators' model selection and prediction performance in the presence of outliers and high-leverage points. The employed quantitative performance criteria are the (out-of-sample) prediction root-mean-squared error (RMSE), false-negative error rate (FNR), false-positive error rate (FPR), and classification error rate (CER). The latter metric measures the proportion of misclassified instances (false positives and false negatives) out of the total instances, thus capturing both FPR and FNR in one metric. Specifically, 
\begin{align}
    \text{CER}=\frac{\text{Number of FPs}+\text{Number of FNs}}{\text{Total Number of Instances}}.
\end{align}
Here, the total number of instances is equal to $p$.

We then proceed by generating a training sample and a test sample of $n$ observations, each independently. We use the training sample to estimate regression coefficients and the test sample to evaluate the (out-of-sample) prediction accuracy. Across all simulations involving synthetic datasets, both samples are randomly drawn from a standard linear regression model defined by equation (\ref{eq:lin_reg}). Moreover, we extend the simulations by assuming three possible distributions for the measurement errors: a zero-mean Gaussian distribution, a Student's $t$-distribution with three degrees of freedom (heavy-tailed), and a Student's $t$-distribution with one degree of freedom (very heavy-tailed), which is heavy-tailed Cauchy distribution. If the additive errors are normally distributed, the variance $\sigma_u^2$ is set to $\|\bX\bb_0\|_{\ell_2}^2 10^{-\text{SNR}/10}/n$ ($\text{SNR in dB}$). It is worth noting that the Cauchy distribution has infinite variance; consequently, SNR can not be defined for such data. In cases, where the additive errors are heavy-tailed, we use the median of the absolute value of the prediction residuals (MAD) to quantify the prediction accuracy of the estimators. We use five-fold cross-validation using a $\tau$-scale of residuals to select the regularization parameter for $\tau$-Lasso and adaptive $\tau$-Lasso estimators. Details can be found in Section S.II of the Supplemental Material.

 In the following subsections, we first present a robust scheme that will be used for data standardization throughout the paper. We then present the values chosen for the tuning constants $c_0$ and $c_1$ to attain the desired breakdown point and normal efficiency for $\tau$-Lasso and adaptive $\tau$-Lasso estimators. We describe various scenarios under which synthetic datasets are generated. We briefly mention the competing state-of-the-art methods used in this study. In the remainder of this section, the simulation results will be presented, including robustness and prediction performance of the $\tau$-Lasso and the adaptive $\tau$-Lasso estimators compared to other competing state-of-the-art estimators, and the phenomenon of overshrinkage in $\tau$-Lasso and how adaptive $\tau$-Lasso mitigates this issue, as well as a comparison of the theoretical influence function of the adaptive $\tau$-Lasso estimator with its finite-sample counterpart, the standardized sensitivity curve.

\subsection{Standardization of data} \label{sssec:data_stand}
Across all simulations, we assume the data-generating model assumes an intercept term equal to zero. The data standardization is carried out by centering all columns of the augmented regression matrix $[\mathbf{1}, \mathbf{X}]$ except for the first one using a bisquare location estimator and scaling the resulting columns using a bisquare scale estimator. The response vector $\by$ is then centered using a bisquare location estimator \cite{maronna2011robust}. 

\subsection{Choice of tuning constants $c_0$ and $c_1$}
Herein, the main goal is to attain a $ 25 \%$ breakdown point and $95 \%$ Gaussian efficiency in the absence of regularization by tuning parameters $c_0$ and $c_1$ for both the $\tau$-Lasso estimator and the adaptive $\tau$-Lasso estimator. To do so, we shall set $c_0=2.9370$ and $c_1=5.1425$ so that we can simultaneously attain the desired robustness against outliers and high normal efficiency.

\subsection{Simulation scenarios}
Here, we consider five different scenarios for which synthetic datasets are created as follows:

\begin{itemize}
    \item \textbf{Scenario 1}: We chose the simulation setup in this scenario as $n=50$, $p=10$ with a moderately high ratio of $p/n=0.2$, $\text{SNR}=5 \text{ dB}$, and 
    \begin{equation}
        \bb_0=[4, 2, 0, 0, 3, 0, 0, 0 ,0, 0]^T.
    \end{equation}
    Each row of the regression matrix $\bX$ is independently drawn from a multivariate Gaussian distribution $\mathcal{N}(\mathbf{0}, \boldsymbol{\Sigma})$ with a Toeplitz covariance structure $\Sigma_{ij}=\rho^{|i-j|}$ with $\rho=0.5$. 
    
    \item \textbf{Scenario 2}: We set the simulation parameters in this scenario as follows: $n=40$, $p=500$ with  $p/n>1$ (under-determined system), $\text{SNR}=15 \text{ dB}$, and 
    \begin{equation}
        \bb_0=[2, 2, 2, 2, 2, 2, 2, 2,\mathbf{0}_{492}^T]^T.
    \end{equation}
    Each row of $\bX$ is a $p$-dimensional vector of covariates, independently drawn from a multivariate Gaussian distribution $\mathcal{N}(\mathbf{0}, \boldsymbol{\Sigma})$ with $\Sigma_{ij}=\rho^{|i-j|}$ with $\rho=0.5$. 
    
    \item \textbf{Scenario 3}: We chose the simulation setup in this scenario as follows: $n=100$, $p=30$ with a high ratio of $p/n=0.3$, $\text{SNR}=25 \text{ dB}$, and 
    \begin{equation}
        \bb_0=[\underbrace{2.5, \cdots, 2.5}_{5 \text{ entries}}, \underbrace{1.5, \cdots, 1.5}_{5 \text{ entries}},\underbrace{ 0.5, \cdots, 0.5}_{5 \text{ entries}}, \mathbf{0}_{15}^T]^T.
    \end{equation}
    Each row of the regression matrix $\bX$ is independently drawn from a multivariate Gaussian distribution $\mathcal{N}(\mathbf{0}, \boldsymbol{\Sigma})$ with $\Sigma_{ij}=\rho^{|i-j|}$ with $\rho=0.95$. 
    
    \item \textbf{Scenario 4}: We chose the simulation setup in this scenario as follows: $n=100$, $p=200$ with $p/n>1$ (under-determined system), $\text{SNR}=25 \text{ dB}$, and 
    \begin{equation}
        \bb_0=[\underbrace{2.5, \cdots, 2.5}_{5 \text{ entries}}, \underbrace{1.5, \cdots, 1.5}_{5 \text{ entries}},\underbrace{ 0.5, \cdots, 0.5}_{5 \text{ entries}}, \mathbf{0}_{185}^T]^T.
    \end{equation}
    The first $15$ covariates and the remaining $185$ covariates are assumed to be independent. Each row of the regression matrix $\bX$ is independently drawn from a multivariate Gaussian distribution $\mathcal{N}(\mathbf{0}, \boldsymbol{\Sigma})$ with $\Sigma_{ij}=\rho^{|i-j|}$ with $\rho=0.95$ for $i,j=1,\cdots,15$ and $i,j=16,\cdots,200$, and $\Sigma_{ij}=0$ elsewhere. 
    
  \item \textbf{Scenario 5}: In this scenario, we chose the same simulation setup as in scenario 4, except that
     \begin{equation}
        \bb_0=[\underbrace{2.5, \cdots, 2.5}_{5 \text{ entries}}, 0, 1.5, 1.5,  \mathbf{0}_{192}^T]^T.
    \end{equation}
     The above setup provides some insight into the impact of reducing the sparsity level $k_0/p$.
    
\end{itemize}

In all scenarios, the synthetic datasets are contaminated by outliers and high-leverage points. We introduce outliers by setting $10 \%$ of entries in the response vector $\by$ to random draws from $\mathcal{N}(100,1)$. Likewise, we introduce high-leverage points by setting $10 \%$ of observations in the regression matrix $\bX$ to random draws from a multivariate Gaussian $\mathcal{N}(30,\mathbf{I}_p)$. 

\subsection{Competing state-of-the-art methods}
We compare adaptive $\tau$-Lasso and $\tau$-Lasso estimators' performance with several state-of-the-art methods, including adaptive \textit{MM}-Lasso, \textit{MM}-Lasso \cite{smucler2017robust}, adaptive PENSE \cite{kepplinger2023robustnew}, sparse-LTS \cite{alfons2013sparse}, ESL-Lasso \cite{wang2013robust}, LAD-Lasso \cite{wang2007robust}, Lasso \cite{tibshirani96} and Oracle estimator (which requires the knowledge of true support and is used for benchmarking). Details on calibrating regularization parameters and the libraries used for implementation can be found in Section S.III of the Supplemental Material.

\subsection{Results}
\label{subsec:results}
We present the simulation results using the datasets generated in the five scenarios discussed above. For each scenario, we run a Monte-Carlo study of $500$ trials where a random realization of $\by$ and $\bX$ is used at each trial. In the presence of contamination, the Monte-Carlo experiment is carried out by adding a random realization of outliers in $\by$ and the fixed high-leverage points in $\bX$ at each trial. We then report the simulation results by averaging the mentioned performance measures at the beginning of Section \ref{sec:simulations} across $500$ trials. 

Note that we can only calculate ESL-Lasso for settings with $p/n<1$, which includes scenarios 1 and 3. In the face of contamination from outliers and high-leverage points, the simulation results for ESL-Lasso are not reported for scenarios 1 and 3 as the \href{https://ysph.yale.edu/c2s2/software/elasso/}{\texttt{eLASSO}} MATLAB code does not compile. We summarize the simulation results shown in \textbf{Tables \ref{tab:tab1}} to \textbf{\ref{tab:tab4}} as follows:
\begin{itemize}
    \item Except for a few cases, we obtain almost similar results for the adaptive $\tau$-Lasso and the $\tau$-Lasso estimators compared to the adaptive \textit{MM}-Lasso and the \textit{MM}-Lasso estimators, and our results are generally close to oracle estimators. The adaptive $\tau$-Lasso and $\tau$-Lasso estimators exhibit good performance for all three error distributions across all scenarios. While in scenario 2, they perform slightly worse than the adaptive \textit{MM}-Lasso and the \textit{MM}-Lasso, respectively, the reverse holds true in terms of RMSE in scenario 1. Moreover, in the presence of outliers and high-leverage points for scenarios 3, 4, and 5, the adaptive $\tau$-Lasso and the $\tau$-Lasso show better predictive and model selection behavior than the adaptive \textit{MM}-Lasso and \textit{MM}-Lasso. The adaptive $\tau$-Lasso and $\tau$-Lasso show remarkable performance for all but one of the scenarios in the presence of contamination, with $\tau$-Lasso achieving the best predictive performance in scenarios 3 and 4, and its adaptive version having the best predictive performance and the second best variable selection performance in scenario 5, marginally behind the adaptive PENSE. In the above comparisons between the class of $\tau$-Lasso estimators and the class of \textit{MM}-Lasso estimators, we compare adaptive $\tau$-Lasso with adaptive \textit{MM}-Lasso and $\tau$-Lasso with \textit{MM}-Lasso. 
    
    \item The sparse-LTS estimator shows promising predictive performance across all scenarios, except for scenario 2, which performs poorly without contamination. Furthermore, it shows the worst overall model selection performance for scenarios 1 and 2, except for one instance in scenario 1 with 
    Cauchy errors.
    When outliers and high-leverage points contaminate the data, it performs poorly and shows the worst and second-worst overall model selection results for scenarios 1 and 2, respectively. In contrast, the sparse-LTS estimator exhibits the best model selection performance compared to other estimators for scenario 3, in the presence and absence of contamination.

    \item Except for scenario 5 with Cauchy errors, the adaptive $\tau$-Lasso demonstrated better predictive performance than the adaptive PENSE in all considered scenarios. The adaptive $\tau$-Lasso performs slightly better than the adaptive PENSE in model selection in scenarios 1, 3, and 4 with normal and moderately heavy-tailed errors. In the presence of contamination, adaptive $\tau$-Lasso outperforms adaptive PENSE in model selection for scenarios 1 and 3 and in predictive performance for scenarios 3, 4 and 5. Conversely, the adaptive PENSE marginally outperforms the adaptive $\tau$-Lasso in a few of the remaining scenarios.
    
    \item The ESL-Lasso estimator performs relatively well in scenario 1. However, it shows extremely poor performance in scenario 3.
    
    \item The LAD-Lasso performs well in all scenarios when there is no contamination. However, it performs extremely poorly for certain cases, such as scenarios 3 and 4, when outliers and high-leverage points contaminate the data. In these cases, its performance is significantly worse than the adaptive $\tau$-Lasso and $\tau$-Lasso. This phenomenon may be associated with the sensitivity of absolute-error loss to such contaminations.
    
    \item The Lasso estimator shows a relatively good model selection performance and remarkable predictive performance under normal errors and to a lesser degree, under moderately heavy-tailed errors, which closely agrees with the RMSE and MAD obtained by oracle estimators. When errors are extremely heavy-tailed, such as Cauchy distribution, it may perform significantly worse than other estimators due to the lack of robustness in squared-error loss, for instance, in scenarios 4 and 5. The same issue can arise when the data is contaminated by high-leverage points and outliers, for instance, in scenarios 3, 4, and 5.
    
    \item We observe that the adaptive $\tau$-Lasso tends to have a lower false-positive rate but a higher false-negative rate than $\tau$-Lasso. A similar conclusion can be made for the adaptive \textit{MM}-Lasso and \textit{MM}-Lasso estimator.
    
    \item We observe that none of the nine estimators, excluding the oracle estimator, can outperform the other eight competing estimators in all considered scenarios. However, the classes of $\tau$-Lasso and \textit{MM}-Lasso and adaptive PENSE have an overall reliable performance in all scenarios.

\end{itemize}

In conclusion, our study suggests that the class of $\tau$-Lasso estimators demonstrate a reliable performance in both the presence and absence of contamination, achieving either the best or matching the best performances with minimal or no loss across various settings, except for the oracle estimators where it is assumed that the true support of parameter vector $\bb_0$ is known. Our results highlight the robustness of the class of $\tau$-Lasso, including the $\tau$-Lasso and the adaptive $\tau$-Lasso, and their usefulness in high-dimensional settings.

\vspace{-6pt}
\begin{table*}
\caption{\label{tab:tab1} Simulation results showing the root mean squared-error (the median absolute deviation for heavy-tailed errors), false-negative rate, and false-positive rate of the adaptive $\tau$-Lasso, $\tau$-Lasso, and their competitors for scenarios 1, 2, 3, 4, and 5  with all three error distributions in the absence of contamination, averaged over 500 trials. We observe from the following results that the adaptive $\tau$-Lasso and $\tau$-Lasso estimators exhibit good performance for all three error distributions and across all scenarios, with their RMSE and MAD generally close to that of the oracle estimator. The \colorbox{Lavender}{best}, \colorbox{Periwinkle}{second-best} and \colorbox{YellowOrange}{third-best} values are highlighted with colorcoding, with the oracle estimator excluded from the comparison.}
\resizebox{0.99\textwidth}{!}{
\begin{tabular}{lllllllllllr}
\hline
Scenario  & Normal& & & &$t(3)$ & & & &$t(1)$& & \\
\cline{2-4}\cline{6-8} \cline{10-12} 
 & $\text{RMSE}$ & $\text{FNR}$ & $\text{FPR}$ & & $\text{MAD}$ & $\text{FNR}$ & $\text{FPR}$ & &$\text{MAD}$ & $\text{FNR}$ & $\text{FPR}$\\
 \hline
 1 \\
 \hline
  Adaptive $\tau$-Lasso& \colorbox{Periwinkle}{$3.8621$} & $0.0187$ & $0.2569$ & & \colorbox{Lavender}{$0.8378$} & $0$ & $0.1834$ & &\colorbox{Periwinkle}{$1.1827$} & $0$ & $0.2040$\\
  $\tau$-Lasso &\colorbox{YellowOrange}{$3.8742$} & $0.0033$ & $0.4391$ & & $0.8664$ & $0$ & $0.4657$ & &$1.2351$ & $0$ & $0.4694$\\
  Adaptive \textit{MM}-Lasso& $3.9096$ & $0.0307$ & $0.1960$ & & \colorbox{YellowOrange}{$0.8517$} & $0$ & $0.1754$ & &\colorbox{Lavender}{$1.1798$} & $0.0013$ & $0.1403$\\
  \textit{MM}-Lasso& $3.9493$ & $0.0060$ & $0.4431$ & & $0.8870$ & $0$ & $0.4571$ & &$1.2526$ & $0$ & $0.4357$\\
 Adaptive PENSE& $4.0165$ & $0.0287$ & $0.3063$ & & \colorbox{Periwinkle}{$0.8504$} & $0$ & $0.2194$ & &\colorbox{YellowOrange}{$1.1972$} & $0.0013$ & $0.2694$\\
  Sparse-LTS & $4.4895$ & $0.0280$ & $0.6789$ & & $0.9399$ & $0$ & $0.5200$ & &$1.2529$ & $0.00067$ & $0.4571$\\
  ESL-Lasso & $4.8880$ & $0.2747$ & $0.1146$ & & $0.9542$ & $0.0107$ & $0.0877$ & &$1.2519$ & $0.0173$ & $0.0514$\\
  LAD-Lasso & $4.0021$ & $0.0173$ & $0.4894$ & & $0.8872$ & $0$ & $0.4611$ & &$1.2481$ & $0$ & $0.4846$\\
  Lasso & \colorbox{Lavender}{$3.8175$} & $0.0047$ & $0.3700$ & & $0.9247$ & $0.00067$ & $0.4151$ & &$3.0092$ & $0.2967$ & $0.3026$\\
  Oracle & $3.6428$ & $0$ & $0$ & & $0.8038$ & $0$ & $0$ & &$1.0761$ & $0$ & $0$\\
  \hline
    2 \\
  \hline
   Adaptive $\tau$-Lasso& $4.5574$ & $0.1848$ & $0.0368$ & & $2.8488$ & $0.1858$ & $0.0372$ & &\colorbox{Periwinkle}{$4.0363$} & $0.2603$ & $0.0424$\\
  $\tau$-Lasso & $4.8986$ & $0.2380$ & $0.0264$ & & $3.0789$ & $0.2288$ & $0.0260$ & & $4.2530$ & $0.3335$ & $0.0353$\\
  Adaptive \textit{MM}-Lasso& \colorbox{YellowOrange}{$3.4002$} & $0.1713$ & $0.0097$ & &\colorbox{YellowOrange}{$2.1688$} & $0.1695$ & $0.0091$ && \colorbox{YellowOrange}{$4.0382$} & $0.4148$ & $0.0146$\\
  \textit{MM}-Lasso& $3.8334$ & $0.1273$ & $0.0404$ & & $2.4655$ & $0.1335$ & $0.0394$ & &$4.1209$ & $0.3230$ & $0.0352$\\
  Adaptive PENSE& $5.4250$ & $0.2845$ & $0.0333$ & & $3.4340$ & $0.2828$ & $0.0336$ & &$4.1631$ & $0.3025$ & $0.0348$\\
  Sparse-LTS & $7.5244$ & $0.5035$ & $0.0477$ & & $4.8378$ & $0.5000$ & $0.0478$ & &$5.4039$ & $0.5165$ & $0.0484$\\
  LAD-Lasso & \colorbox{Periwinkle}{$3.0763$} & $0.0353$ & $0.0493$ & & \colorbox{Periwinkle}{$1.9045$} & $0.0410$ & $0.0477$ & &\colorbox{Lavender}{$3.7147$} & $0.2420$ & $0.0367$\\
  Lasso & \colorbox{Lavender}{$2.6520$} & $0.0060$ & $0.0421$ & & \colorbox{Lavender}{$1.7161$} & $0.0145$ & $0.0428$ & &$5.3161$ & $0.5105$ & $0.0270$\\
  Oracle & $1.7833$ & $0$ & $0$ & & $0.9061$ & $0$ & $0$ & &$1.3729$ & $0$ & $0$\\
 \hline
  3 \\
 \hline
  Adaptive $\tau$-Lasso& $1.3186$ & $0.1964$ & $0.0483$ & & \colorbox{YellowOrange}{$0.9163$} & $0.1936$ & $0.0544$ & &$1.3017$ & $0.2324$ & $0.1225$\\
  $\tau$-Lasso &\colorbox{Lavender}{$1.2642$} & $0.0784$ & $0.1789$ & & \colorbox{Lavender}{$0.8765$} & $0.0888$ & $0.1724$ & &\colorbox{Lavender}{$1.2595$} & $0.1399$ & $0.1939$\\
  Adaptive \textit{MM}-Lasso& $1.3606$ & $0.1963$ & $0.1265$ & & $0.9382$ & $0.2131$ & $0.0853$ & &$1.3676$ & $0.2931$ & $0.0697$\\
  \textit{MM}-Lasso& \colorbox{YellowOrange}{$1.3036$} & $0.0916$ & $0.2371$ & & \colorbox{Periwinkle}{$0.8916$} & $0.0955$ & $0.1941$ & &\colorbox{Periwinkle}{$1.2703$} & $0.1503$ & $0.1625$\\
  Adaptive PENSE& $1.3560$ & $0.2051$ & $0.0677$ & & $0.9288$ & $0.1995$ & $0.0695$ & &$1.3147$ & $0.2273$ & $0.1301$\\
  Sparse-LTS & $1.5138$ & $0.1263$ & $0.0693$ & & $1.0468$ & $0.1295$ & $0.0647$ & &$1.3266$ & $0.1624$ & $0.0732$\\
  ESL-Lasso & $9.5278$ & $0.6803$ & $0.1512$ & & $6.1236$ & $0.6713$ & $0.1491$ & &$6.5134$ & $0.6896$ & $0.1291$\\
  LAD-Lasso & $1.3202$ & $0.0943$ & $0.2516$ & & $0.9192$ & $0.1004$ & $0.2288$ & &\colorbox{YellowOrange}{$1.2736$} & $0.1441$ & $0.2123$\\
  Lasso & \colorbox{Periwinkle}{$1.3013$} & $0.0765$ & $0.2675$ & & $1.0214$ & $0.1163$ & $0.2975$ & &$4.6165$ & $0.5227$ & $0.2464$\\
  Oracle & $1.2352$ & $0$ & $0$ & & $0.8694$ & $0$ & $0$ & &$1.2301$ & $0$ & $0$\\
  \hline
   4 \\
 \hline
  Adaptive $\tau$-Lasso& \colorbox{Lavender}{$1.2629$} & $0.0955$ & $0.0044$ & & \colorbox{Lavender}{$0.8911$} & $0.1068$ & $0.0088$ & &\colorbox{Lavender}{$1.3034$} & $0.1616$ & $0.0233$\\
  $\tau$-Lasso &\colorbox{Periwinkle}{$1.3021$} & $0.0732$ & $0.0368$ & & \colorbox{Periwinkle}{$0.9188$} & $0.0841$ & $0.0388$ & &\colorbox{Periwinkle}{$1.3300$} & $0.1356$ & $0.0404$\\
  Adaptive \textit{MM}-Lasso& $1.4396$ & $0.2267$ & $0.0191$ & & $1.0076$ & $0.2264$ & $0.0188$ & &$1.4250$ & $0.2967$ & $0.0141$\\
  \textit{MM}-Lasso& $1.4275$ & $0.1124$ & $0.0644$ & & $1.0004$ & $0.1111$ & $0.0640$ & &$1.3956$ & $0.1548$ & $0.0485$\\
  Adaptive PENSE& $1.4945$ & $0.2177$ & $0$ & & $1.0405$ & $0.2173$ & $0.00001$ & &$\colorbox{YellowOrange}{1.3531}$ & $0.2245$ & $0.0006$\\
  Sparse-LTS & $1.5063$ & $0.1120$ & $0.0103$ & & $1.0622$ & $0.1127$ & $0.0126$ & &$1.3875$ & $0.1433$ & $0.0193$\\
  LAD-Lasso & \colorbox{YellowOrange}{$1.3660$} & $0.0921$ & $0.0602$ & & \colorbox{YellowOrange}{$0.9648$} & $0.0960$ & $0.0449$ & &$1.3652$ & $0.1473$ & $0.0345$\\
  Lasso & $1.4568$ & $0.0736$ & $0.2329$ & & $1.2192$ & $0.1223$ & $0.2251$ & &$13.423$ & $0.5621$ & $0.1204$\\
  Oracle & $1.2387$ & $0$ & $0$ & & $0.8764$ & $0$ & $0$ & &$1.2264$ & $0$ & $0$\\
   \hline
   5 \\
 \hline
   Adaptive $\tau$-Lasso& \colorbox{Lavender}{$0.8569$} & $0.0006$ & $0.0087$ & & \colorbox{Lavender}{$0.8439$} & $0.0057$ & $0.0145$ & &\colorbox{Periwinkle}{$1.2129$} & $0.0303$ & $0.0271$\\
  $\tau$-Lasso & \colorbox{Periwinkle}{$0.8832$} & $0$ & $0.0373$ & & \colorbox{Periwinkle}{$0.8723$} & $0.0037$ & $0.0419$ & &\colorbox{YellowOrange}{$1.2371$} & $0.0246$ & $0.0458$\\
  Adaptive \textit{MM}-Lasso& \colorbox{YellowOrange}{$0.8835$} & $0.0100$ & $0.0086$ & & $0.9011$ & $0.0466$ & $0.0134$ & &$1.2664$ & $0.1143$ & $0.0107$\\
  \textit{MM}-Lasso& $0.9084$ & $0.0006$ & $0.0455$ & & $0.9104$ & $0.0134$ & $0.0480$ & &$1.2800$ & $0.0317$ & $0.0453$\\
   Adaptive PENSE& $0.9699$ & $0.0240$ & $0.0026$ & & \colorbox{YellowOrange}{$0.8842$} & $0.0303$ & $0.0029$ & &\colorbox{Lavender}{$1.1846$} & $0.0466$ & $0.0055$\\
  Sparse-LTS & $0.9994$ & $0.0014$ & $0.0165$ & & $0.9576$ & $0.0134$ & $0.0272$ & &$1.2652$ & $0.0269$ & $0.0319$\\
 LAD-Lasso & $0.9314$ & $0.00086$ & $0.0619$ & & $0.9040$ & $0.0071$ & $0.0489$ & &$1.2738$ & $0.0223$ & $0.0397$\\
 Lasso & $0.9967$ & $0.00086$ & $0.2344$ & & $1.1175$ & $0.0214$ & $0.2016$ & &$6.6167$ & $0.4700$ & $0.1031$\\
  Oracle & $0.8388$ & $0$ & $0$ & & $0.8146$ & $0$ & $0$ & &$1.0967$ & $0$ & $0$\\
\end{tabular}
}
\end{table*}

\begin{table*}  
\caption{\label{tab:tab3} Simulation results showing the root mean squared-error, false-negative rate, and false-positive rate of the adaptive $\tau$-Lasso, $\tau$-Lasso, and their competitors for scenarios 1, 2, 3, 4, and 5 with normal errors and in the presence of contamination, averaged over 500 trials. We observe from the following results that the adaptive $\tau$-Lasso and $\tau$-Lasso estimators achieve the top-three or close-to-top-three performance in terms of RMSE and variable selection accuracy for almost all scenarios. The \colorbox{Lavender}{best}, \colorbox{Periwinkle}{second-best} and \colorbox{YellowOrange}{third-best} values are highlighted with colorcoding, with the oracle estimator excluded from the comparison.}
\resizebox{\textwidth}{!}{
\setlength{\tabcolsep}{10pt}
\begin{tabular}{lllllllr}
\hline
 Scenario & $\text{RMSE}$ &$\text{FNR}$ & $\text{FPR}$ & Scenario & $\text{RMSE}$ &$\text{FNR}$ & $\text{FPR}$ \\
\cline{2-4} \cline{6-8}
 1 & & & & 2 & & & \\
 \hline
  Adaptive $\tau$-Lasso  & $4.8750$ & $0.1587$& $0.3003$ & Adaptive $\tau$-Lasso & $6.9055$ & $0.4548$ & $0.0362$   \\ 
  $\tau$-Lasso  &\colorbox{YellowOrange}{$4.8539$} & $0.0580$ & $0.4246$ & $\tau$-Lasso & $6.8249$ & $0.4493$ & $0.0291$  \\

  Adaptive \textit{MM}-Lasso & \colorbox{Periwinkle}{$4.5907$} & $0.1800$ & $0.2357$ &  Adaptive \textit{MM}-Lasso  & \colorbox{YellowOrange}{$6.0858$} & $0.5715$ & $0.0091$  \\
 
  \textit{MM}-Lasso  & $5.1054$ & $0.0940$ & $0.3823$ &  \textit{MM}-Lasso  & $6.9527$ & $0.4968$ &$0.0213$   \\

  Adaptive PENSE & $4.9093$ & $0.1553$ & $0.3180$ &  Adaptive PENSE  & $6.8788$ & $0.5020$ & $0.0222$  \\
 
  Sparse-LTS  & \colorbox{Lavender}{$4.3348$} & $0.0287$ & $0.5637$ & Sparse-LTS    & $6.4486$ & $0.3598$ & $0.0455$\\

  LAD-Lasso  & $5.2491$ & $0.1253$ & $0.3554$  & LAD-Lasso   & \colorbox{Periwinkle}{$5.5186$} & $0.2628$ & $0.0454$  \\

  Lasso &   $5.0737$ & $0.0747$ & $0.4266$ &  Lasso   & \colorbox{Lavender}{$5.2231$} & $0.1935$ & $0.0509$   \\

  Oracle  & $3.7984$ & $0$ & $0$ & Oracle & $1.8248$ & $0$ & $0$ \\
  
  \hline
  3 & & & & 4 & & & \\
 \hline
   Adaptive $\tau$-Lasso  & \colorbox{Periwinkle}{$1.3311$} &  $0.2019$ &  $0.0448$  &  Adaptive $\tau$-Lasso  & \colorbox{Periwinkle}{$1.3482$} &  $0.1472$ & $0.0165$  \\
  $\tau$-Lasso  & \colorbox{Lavender}{$1.2738$} & $0.0884$ &  $0.1239$ & $\tau$-Lasso  & \colorbox{Lavender}{$1.3241$} & $0.0825$ & $0.0312$ \\
  
  Adaptive \textit{MM}-Lasso & $3.4547$& $0.3605$  & $0.0973$ &  Adaptive \textit{MM}-Lasso &$4.2428$ &  $0.5940$ &  $0.0064$ \\
  
  \textit{MM}-Lasso  &   $5.6258$ & $0.2829$ &  $0.1597$  &   \textit{MM}-Lasso & $7.1007$ &  $0.3707$ & $0.0688$  \\

  Adaptive PENSE & \colorbox{YellowOrange}{$1.3925$} & $0.1956$ & $0.0573$ &  Adaptive PENSE  & $1.6069$ & $0.2572$ & $0.0006$  \\
  
  Sparse-LTS  &  $1.4075$ & $0.1251$ & $0.0701$  & Sparse-LTS  & \colorbox{YellowOrange}{$1.4307$} &  $0.1124$ & $0.0128$ \\
  
  LAD-Lasso & $15.6940$ &  $0.6340$ &  $0.2368$ & LAD-Lasso  &  $8.6238$ &  $0.4309$ &  $0.1218$  \\
  
  Lasso  &   $15.0113$ & $0.6639$ & $0.2525$  &   Lasso & $8.4566$ &  $0.4132$ &  $0.2017$\\
  
  Oracle  &  $1.2672$ &  $0$ &  $0$ &  Oracle & $1.2727$ &  $0$ & $0$  \\
   \hline
   5 & & & &  $---$ & & & \\
 \hline
   Adaptive $\tau$-Lasso &  \colorbox{Lavender}{$0.9337$} & $0.0117$ & $0.0164$ &$---$ &  $---$ & $---$ & $---$ \\
  $\tau$-Lasso & \colorbox{Periwinkle}{$0.9526$} & $0.0037$ & $0.0393$ & $---$ &  $---$ & $---$ & $---$\\
  Adaptive \textit{MM}-Lasso & $3.0690$ & $0.4869$ &  $0.0156$ &$---$ &  $---$ & $---$ & $---$ \\
   \textit{MM}-Lasso & $5.7360$ & $0.3323$ &  $0.0978$&$---$ &  $---$ & $---$ & $---$\\
  Adaptive PENSE & $0.9955$ & $0.0591$ & $0.0057$ &  $---$  & $---$ & $---$ & $---$  \\
    
  Sparse-LTS &  \colorbox{YellowOrange}{$0.9574$} &  $0.0017$ &  $0.0176$ &$---$ &  $---$ & $---$ & $---$ \\
   LAD-Lasso & $5.5076$ &  $0.3134$ & $0.1291$&$---$ &  $---$ & $---$  & $---$\\
     Lasso & $5.3642$ & $0.2757$ & $0.2075$  &$---$ &  $---$ & $---$ & $---$\\
  Oracle & $0.8440$ &  $0$ & $0$ &$---$ &  $---$ & $---$ & $---$ \\
\end{tabular}
}
\vspace{-21 pt}
\end{table*}

\begin{table*}  
\caption{\label{tab:tab4} Simulation results showing the classification error rates of the adaptive $\tau$-Lasso, $\tau$-Lasso, and their competitors for scenarios 1, 2, 3, 4, and 5  in the presence and absence of contamination, averaged over 500 trials. We observe from the following results that the adaptive $\tau$-Lasso and $\tau$-Lasso estimators achieve the top-three best or close-to-top-three-best performance in terms of variable selection accuracy for almost all scenarios. The \colorbox{Lavender}{best}, \colorbox{Periwinkle}{second-best} and \colorbox{YellowOrange}{third-best} values are highlighted with colorcoding.}
\resizebox{\textwidth}{!}{
\setlength{\tabcolsep}{10pt}
\begin{tabular}{lllllllllr}
 \hline
 Scenario  & $\text{CER}$ & & & &  Scenario & $\text{CER}$ & & &  \\
 \cline{2-5} \cline{7-10}  
 & $\text{Normal}$ &$t(3)$ & $t(1)$ & $\text{Contam.}$ &  &$\text{Normal}$ & $t(3)$ & $t(1)$ & $\text{Contam.}$\\
 \hline 
 1 & & & &  &  2 & & & & \\
 \hline
  Adaptive $\tau$-Lasso  & \colorbox{YellowOrange}{$0.1854$} & \colorbox{YellowOrange}{$0.1284$}& \colorbox{YellowOrange}{$0.1428$} & \colorbox{Periwinkle}{$0.2578$} & Adaptive $\tau$-Lasso & $0.0392$ & $0.0396$ & $0.0459$&  $0.0429$ \\ 
  
  $\tau$-Lasso  &$0.3084$ & $0.3260$ & $0.3286$ & $0.3146$ & $\tau$-Lasso & \colorbox{Periwinkle}{$0.0297$} & \colorbox{Periwinkle}{$0.0292$} & $0.0400$ & $0.0358$\\

  Adaptive \textit{MM}-Lasso & \colorbox{Lavender}{$0.1464$} & \colorbox{Periwinkle}{$0.1228$} & \colorbox{Periwinkle}{$0.0986$} & \colorbox{Lavender}{$0.2190$} & Adaptive \textit{MM}-Lasso  & \colorbox{Lavender}{$0.0123$} & \colorbox{Lavender}{$0.0116$} & \colorbox{Lavender}{$0.0210$} & \colorbox{Lavender}{$0.0181$} \\
 
  \textit{MM}-Lasso  & $0.3120$ & $0.3200$ & $0.3050$ & $0.2958$ & \textit{MM}-Lasso  & $0.0418$ & $0.0409$ &$0.0398$ & \colorbox{Periwinkle}{$0.0289$} \\

  Adaptive PENSE & $0.2230$ & $0.1536$ & $0.1890$ & \colorbox{YellowOrange}{$0.2692$} & Adaptive PENSE  & \colorbox{YellowOrange}{$0.0374$} & \colorbox{YellowOrange}{$0.0376$} & \colorbox{YellowOrange}{$0.0391$} & \colorbox{YellowOrange}{$0.0299$}\\
 
  Sparse-LTS  & $0.4836$ & $0.3640$ & $0.3202$ & $0.4032$& Sparse-LTS    & $0.0549$ & $0.0550$ & $0.0559$ & $0.0505$\\

  ESL-Lasso  & \colorbox{Periwinkle}{$0.1626$} & \colorbox{Lavender}{$0.0646$} & \colorbox{Lavender}{$0.0412$} & $---$ & ESL-Lasso    & $---$ & $---$ & $---$ & $---$\\

  LAD-Lasso  & $0.3478$ & $0.3228$ & $0.3392$  & $0.2864$ & LAD-Lasso   & $0.0490$ & $0.0476$ & $0.0400$& $0.0488$  \\

  Lasso &   $0.2604$ & $0.2908$ & $0.3008$ & $0.3210$ & Lasso   & $0.0416$ & $0.0423$ & \colorbox{Periwinkle}{$0.0347$}  & $0.0532$  \\
  
    \hline
 3 & & & &  &  4 & & & & \\
 \hline
   Adaptive $\tau$-Lasso  & \colorbox{Periwinkle}{$0.1223$} &  \colorbox{Periwinkle}{$0.1240$} &  $0.1775$  & \colorbox{YellowOrange}{$0.1233$} &  Adaptive $\tau$-Lasso  & \colorbox{Lavender}{$0.0113$} &  \colorbox{Lavender}{$0.0161$} & \colorbox{YellowOrange}{$0.0337$} & \colorbox{YellowOrange}{$0.0263$} \\
   
  $\tau$-Lasso  & \colorbox{YellowOrange}{$0.1287$} & \colorbox{YellowOrange}{$0.1306$} &  \colorbox{YellowOrange}{$0.1669$} & \colorbox{Periwinkle}{$0.1061$} & $\tau$-Lasso  & $0.0395$ & $0.0422$ & $0.0475$ & $0.0350$ \\
  
  Adaptive \textit{MM}-Lasso & $0.1614$& $0.1492$  & $0.1814$ & $0.2289$ & Adaptive \textit{MM}-Lasso &$0.0347$ &  $0.0344$ &  $0.0353$ & $0.0505$\\
  
  \textit{MM}-Lasso  &   $0.1643$ & $0.1448$ &  \colorbox{Periwinkle}{$0.1564$}  & $0.2213$ & \textit{MM}-Lasso & $0.0680$ &  $0.0676$ & $0.0564$  & $0.0915$\\

  Adaptive PENSE & $0.1364$ & $0.1345$ & $0.1787$ & $0.1265$&  Adaptive PENSE  & \colorbox{Periwinkle}{$0.0163$} & \colorbox{Periwinkle}{$0.0163$} & \colorbox{Lavender}{$0.0174$} & \colorbox{Lavender}{$0.0198$} \\
  
  Sparse-LTS  &  \colorbox{Lavender}{$0.0978$} & \colorbox{Lavender}{$0.0971$} & \colorbox{Lavender}{$0.1178$} & \colorbox{Lavender}{$0.0976$} & Sparse-LTS  & \colorbox{YellowOrange}{$0.0179$} &  \colorbox{YellowOrange}{$0.0201$} & \colorbox{Periwinkle}{$0.0286$} & \colorbox{Periwinkle}{$0.0203$}\\

  ESL-Lasso  &  $0.4157$ & $0.4102$ & $0.4093$ & $---$ & ESL-Lasso  & $---$ &  $---$ & $---$ & $---$\\
  
  LAD-Lasso & $0.1729$ &  $0.1646$ &  $0.1782$& $0.4354$ & LAD-Lasso  &  $0.0626$ &  $0.0487$ &  $0.0430$  & $0.1449$\\
  
  Lasso  &   $0.1720$ & $0.2069$ & $0.3845$  & $0.4582$ & Lasso & $0.2209$ &  $0.2174$ &  $0.1535$& $0.2176$\\
  
\hline
 5 & & & &  &  $---$ & & & & \\
 \hline
   Adaptive $\tau$-Lasso &  \colorbox{Periwinkle}{$0.0084$} & \colorbox{Periwinkle}{$0.0142$} & \colorbox{YellowOrange}{$0.0272$} & \colorbox{Periwinkle}{$0.0162$} & $---$ &  $---$ & $---$ & $---$ & $---$ \\
  $\tau$-Lasso & $0.0360$ & $0.0406$ & $0.0451$ & $0.0381$ & $---$ &  $---$ & $---$ & $---$& $---$\\
  Adaptive \textit{MM}-Lasso & \colorbox{YellowOrange}{$0.0086$} & \colorbox{YellowOrange}{$0.0145$} &  \colorbox{Periwinkle}{$0.0143$} &$0.0321$ &  $---$ & $---$ & $---$  & $---$ & $---$\\
   \textit{MM}-Lasso & $0.0440$ & $0.0468$ &  $0.0448$ & $0.1060$ &  $---$ & $---$ & $---$ & $---$& $---$\\
  Adaptive PENSE & \colorbox{Lavender}{$0.0033$} & \colorbox{Lavender}{$0.0038$} & \colorbox{Lavender}{$0.0069$} &  \colorbox{Lavender}{$0.0075$}  & $---$ & $---$ & $---$ & $---$ & $---$\\
    
  Sparse-LTS &  $0.0160$ &  $0.0268$ &  $0.0317$ &\colorbox{YellowOrange}{$0.0170$} &  $---$ & $---$ & $---$ & $---$ & $---$\\
   LAD-Lasso & $0.0597$ &  $0.0474$ & $0.0391$ & $0.1355$ &  $---$ & $---$  & $---$ & $---$ & $---$ \\
     Lasso & $0.2263$ & $0.1953$ & $0.1159$  &$0.2099$ &  $---$ & $---$ & $---$ & $---$ & $---$\\
\end{tabular}
}
\vspace{-21 pt}
\end{table*}

  \subsection{RMSE under contamination}
  
\begin{figure} 
  \centering
	\includegraphics[width=7.5cm,trim = 0.1cm 0.1cm 0.1cm 1cm, clip]{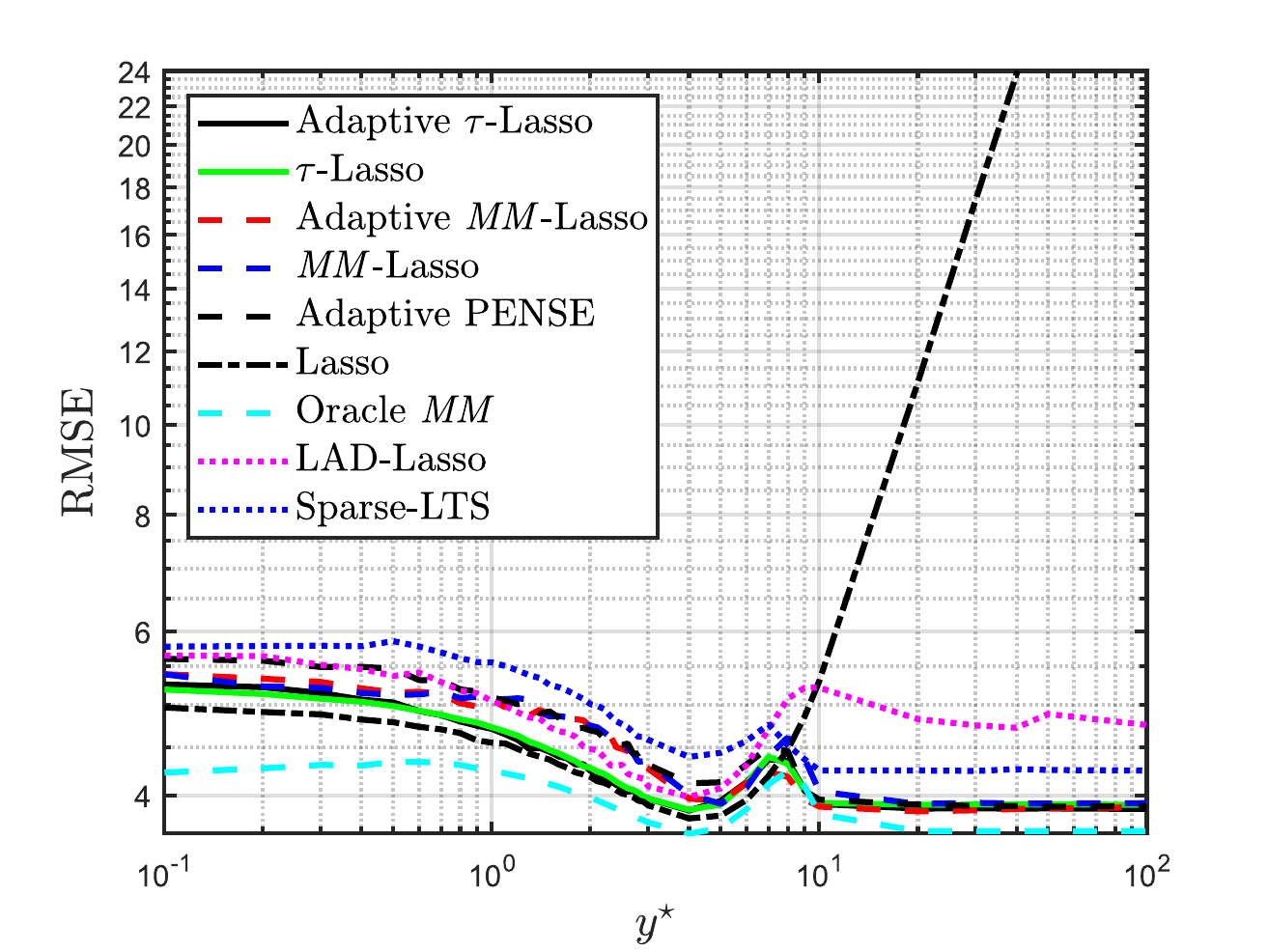}
	\vspace{-5pt}
	\caption{Plots of RMSE as a function of $y^{\star}$ outlier magnitude for each of the estimators under \textbf{Scenario} 1, averaged over 100 trials. Except for the non-robust estimator Lasso, the RMSE values of the remaining estimators for larger values of $y^{\star}$ do not exceed those of the remaining estimators for smaller values of $y^{\star}$. Moreover, both adaptive $\tau$-Lasso and $\tau$-Lasso show slightly better performance than other regularized robust estimators.  } 
	\label{fig:RMSE_plot}
 \vspace{-10pt}
\end{figure}

  Herein, we conduct a simulation study on how the RMSE criterion varies with the outlier magnitude for a range of robust and non-robust estimators. We run the simulations on the dataset in \textbf{Scenario} 1. We introduce contamination to the data model by setting $y_i=5y^{\star}$ and $\bx_{[i]}=[5,0,\cdots,0]^T$ for $i=1,\cdots,\lfloor 0.1\times n \rfloor$. We run a Monte-Carlo experiment with 100 trials, which generates a random realization of $\by$ and $\bX$ for each trial. We then plot the RMSE for each estimator against outlier magnitude $y^{\star}$, varying between $0.1$ to $100$, with results averaged over 100 trials. As shown in \textbf{Fig. \ref{fig:RMSE_plot}}, RMSE values for larger outlier magnitudes $y^{\star}$ remain lower than those for smaller outlier magnitudes $y^{\star}$ in the case of regularized robust estimators. In addition, we observe that both adaptive $\tau$-Lasso and $\tau$-Lasso exhibit slightly better overall performance than other regularized robust estimators. As expected, the RMSE values of the Lasso significantly grow as the outlier magnitude $y^{\star}$ exceeds 5.
  \vspace{-6pt}

\subsection{A remedy for overshrinkage in $\tau$-Lasso}
\label{subsec:overshrinkage}
\begin{figure*}
  \centering
	\includegraphics[width=15.5cm,trim = 1cm 0.1cm 0.1cm 0.1cm, clip]{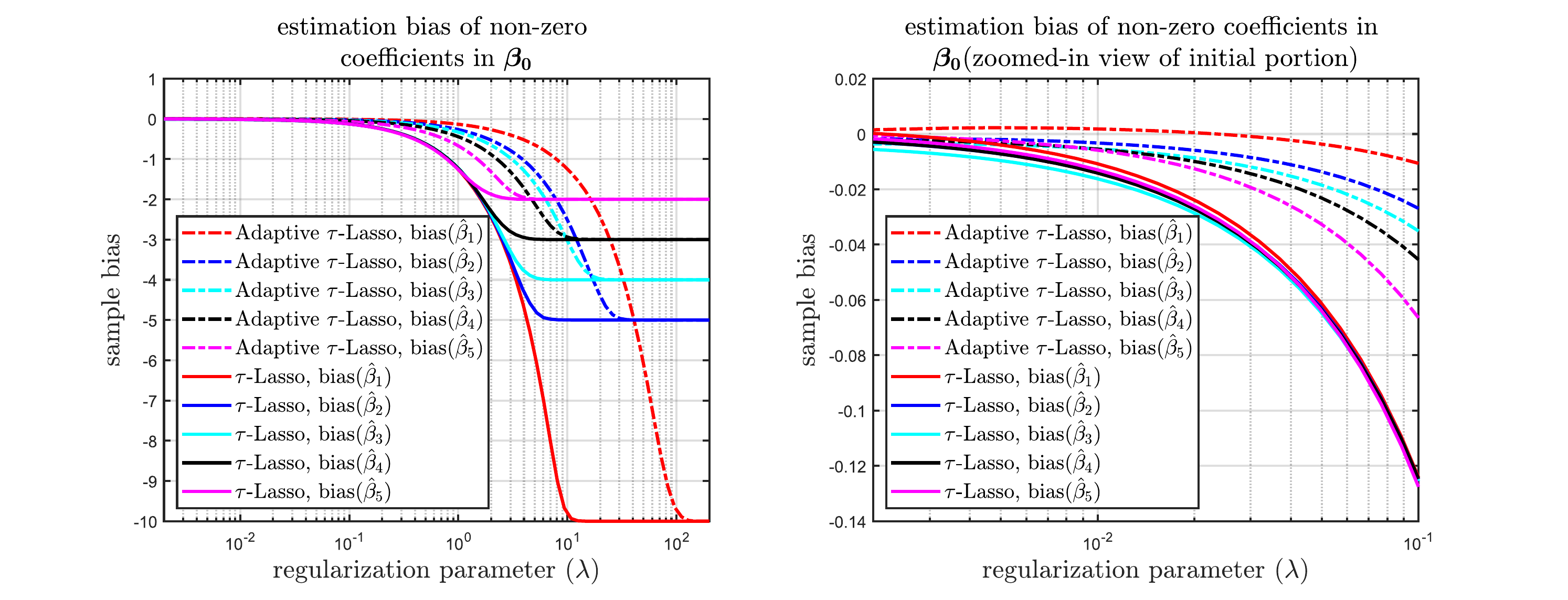}
	\vspace{-5pt}
	\caption{Plots comparing estimation bias between adaptive $\tau$-Lasso and $\tau$-Lasso. As observed from the bias paths, $\tau$-Lasso shrinks non-zero coefficients more heavily toward zero than adaptive $\tau$-Lasso. For large values of $\lambda$, $\tau$-Lasso significantly overshrinks non-zero coefficients compared to adaptive $\tau$-Lasso and produces highly biased estimates, in particular for the large coefficients of $\bb_0$. This behavior continues until $\lambda$ gets sufficiently large, at which point both estimators shrink all coefficients to zero. The right-hand panel (zoomed-in view) shows that when $\lambda$ is too small, overfitting occurs, which even causes slight overestimation for $\hat{\beta}_1$ in adaptive $\tau$-Lasso.} 
	\label{fig:RMSE_bias}
 \vspace{-10pt}
\end{figure*}

A vast literature, including seminal works in \cite{zou2006adaptive} and \cite{fan2004nonconcave}, discusses a key deficiency of Lasso: $\ell_1$-norm penalty tends to shrink estimates for large non-zero coefficients more heavily than for smaller ones. Moreover, when the mutual incoherence condition fails, Lasso-type estimators may select irrelevant variables that have a strong correlation with the relevant ones, further increasing the estimation bias associated with large true coefficients. Such issues arise because $\ell_1$-norm penalty shrinks estimates proportional to magnitude of coefficients. In order to counteract the influence of magnitude, one can assign appropriately chosen weights to regression coefficients using an adaptive $\ell_1$-norm penalty as described in Section \ref{sec:adapative_tau}. We run simulations to test whether the phenomenon of overshrinkage is present in $\tau$-Lasso and how adaptive $\tau$-Lasso remedies this issue. We set $n=50$, $p=10$, $k_0=5$, $\text{SNR}=35 \text{ dB}$, and $\bb_0=[10, 5, 4, 3, 2, 0, 0, 0, 0, 0]^T$. We generate $\bx_{[i]}$'s from a multivariate Gaussian distribution $\mathcal{N}(\mathbf{0}, \mathbf{I}_{p \times p})$, and $u_i$'s from a zero-mean normal distribution with variance as defined in Section \ref{sec:simulations}. We run 500 trials in which each trial uses a random realization of $\bX$ and $\by$. For each trial, the sample bias for each non-zero coefficient is calculated and averaged across all 500 trials. For entire grid of $\lambda$ values, we employ the \textit{S}-Ridge estimator as the pilot estimator, with its regularization parameter selected via the procedure described in Section \ref{subsec:choice_pilot}. Figure \ref{fig:RMSE_bias} shows the bias paths of non-zero coefficients, from which we see that $\tau$-Lasso shrinks the non-zero coefficients more heavily and this trend becomes more pronounced for larger coefficients. For example, the overshrinkage is significant for $\hat{\beta}_1$ associated with the largest true coefficient (10), thus confirming the presence of overshrinkage in $\tau$-Lasso. However, the adaptive $\ell_1$-norm penalty in adaptive $\tau$-Lasso more effectively reduces this bias compared to $\tau$-Lasso, particularly for higher values of the regularization parameter. Note that if $\lambda$ is too large, both adaptive $\tau$-Lasso and $\tau$-Lasso shrink all coefficients to zero. In contrast, if $\lambda$ is too small, overfitting occurs and thus overshrinkage no longer applies. Furthermore, when adaptive $\tau$-Lasso and $\tau$-Lasso employ different criteria for the selection of the regularization parameter, this phenomenon may not be observed. For example, BIC tends to favor simpler models because it penalizes complex models more heavily, while AIC tends to select more complex models, particularly for large sample sizes, due to its light penalty on model complexity, leading to models with lower bias \cite{hastie01statisticallearning}.
 
  \subsection{The empirical validation of the influence function} 
  
\begin{figure} 
\centering
	\includegraphics[width=8cm,trim = 0.1cm 0.1cm 0.1cm 0.1cm, clip]{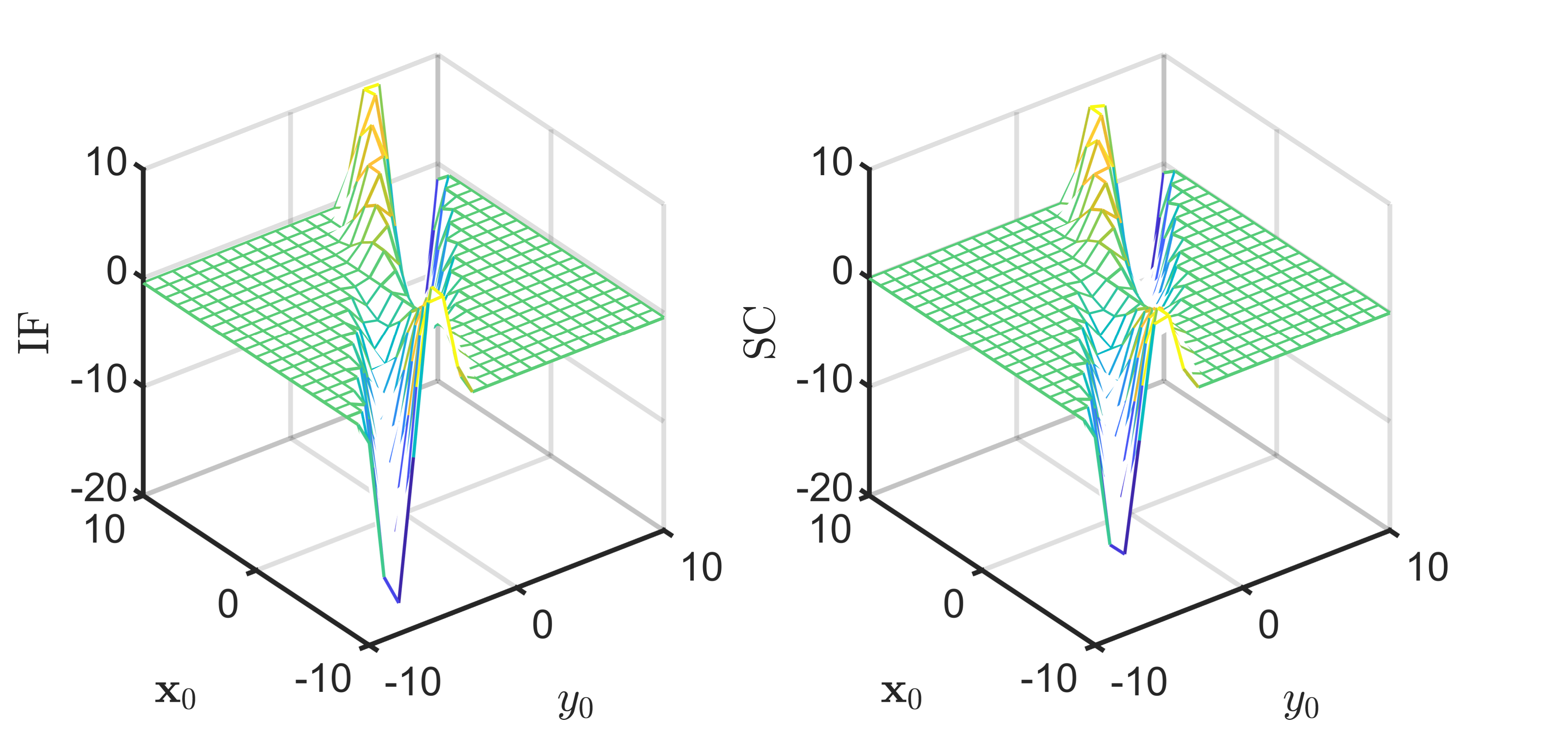}
	\vspace{-12pt}
	\caption{Plots of \textit{influence function} (IF) and \textit{standardized sensitivity curve} (SC) of the adaptive $\tau$-Lasso estimator as a function of $\mathbf{z}_0=(y_0,\bx_{[0]})$ for a one-dimensional toy example with regularization  parameter $\lambda_n=0.1/n$. As predicted, the plotted IF and SC are almost identical and bounded across the entire plotted space, which indicates the correctness of  our results about the influence function of the adaptive $\tau$-Lasso estimator derived in Theorem 7.} 
	\label{fig:IF_plot}
 \vspace{-13pt}
\end{figure}

  In order to study the local robustness properties of the adaptive $\tau$-Lasso, we carry out a simulation study on the influence function of the adaptive $\tau$-Lasso estimator. This allows us to verify the correctness of our results concerning the influence function of the adaptive $\tau$-Lasso estimator derived in Theorem 7. We run the simulations on a toy one-dimensional dataset as visualizing the influence function becomes difficult in high-dimensional problems. We generate a dataset of $n=1000$ i.i.d observations following the linear model described by equation (\ref{eq:lin_reg}) with $p=1$ where the parameter vector $\bb_0=1.5$. The rows of the regression matrix $\bX$ and the noise vector $\bu$ are randomly drawn from a Gaussian distribution with zero mean and unit variance. We calculate the influence function via the closed-form expression derived in Theorem 7, given by equation (\ref{eq:IF-theorem7-adaptive-tau}),  for the given synthetic data. We then validate the results by plotting the \textit{standardized sensitivity curve} (\textbf{SC}), which is a finite-sample version of the influence function. We define the \textit{standardized sensitivity curve} of the estimator $\hat{\theta}$ for a sample of $n$ observations $\mathbf{Z}$ at point $\mathbf{z}_0 \in \mathscr{Z}_0$ as
  
  \begin{equation}
      \text{SC}(\mathbf{z}_0 ; \hat{\theta})= \frac{\hat{\theta}(\mathbf{Z},\mathbf{z}_0)-\hat{\theta}(\mathbf{Z})}{1/(n+1)}.
  \end{equation}
  We compute the derived influence function and standardized sensitivity curve for a two-dimensional grid of $(y_0,\bx_{[0]})$ with a linear spacing of $1$, spanning from $-10$ to $10$ along each dimension for $\lambda_n=0.1/n$. We observe from \textbf{Fig. \ref{fig:IF_plot}} that the influence function and the standardized sensitivity curve are almost identical and bounded across the plotted space. 

\subsection{When does adaptive $\tau$-Lasso outperform $\tau$-Lasso?}
Here, we briefly discuss several scenarios where adaptive $\tau$-Lasso gives superior performance compared to the $\tau$-Lasso and other considered regularized robust estimators. First, when the true regression coefficients vary greatly in magnitude, the $\tau$-Lasso tends to overshrink large coefficients and introduces significant bias. In contrast, adaptive $\tau$-Lasso reduces this bias via an adaptive $\ell_1$-norm penalty. This advantage is demonstrated in Section \ref{subsec:overshrinkage}. Second, in practical scenarios where good leverage points are present on truly irrelevant predictors, the $\tau$-Lasso may produce a higher number of false positives. Adaptive $\tau$-Lasso addresses this problem and reduces the number of false positives. A detailed explanation of this phenomenon can be found
in \cite{kepplinger2023robustnew}, and simulation results demonstrating this behavior are presented in Section S.VI of the Supplemental Material. Third, when truly irrelevant variables are highly correlated with the set of truly relevant variables, the $\tau$-Lasso may fail to correctly select truly relevant variables or even select too many irrelevant variables. Moreover, adaptive $\tau$-Lasso can outperform regularized robust regression estimators with convex loss functions, such as LAD-Lasso, particularly in the presence of high-leverage points. Furthermore, in scenarios with a small sample size, adaptive $\tau$-Lasso shows better predictive and variable selection performance compared to Sparse-LTS, because Sparse-LTS discards a portion of the data to achieve robustness, which adversely impacts the accuracy of the estimation. Note that adaptive $\tau$-Lasso outperforms $\tau$-Lasso and the estimators discussed above in the scenarios considered, given that the pilot estimates are reasonably accurate. 

\section{Conclusion}
\label{sec:conclusion}

  This paper introduced the adaptive $\tau$-Lasso estimator for dealing with high-dimensional data subject to outliers and high-leverage points and discussed its favorable robustness and statistical properties. We established asymptotic theory for consistency of the $\tau$-Lasso and showed that the adaptive $\tau$-Lasso possesses the oracle properties. We then analyzed the adaptive $\tau$-Lasso estimator from a robustness perspective and derived its finite-sample breakdown point and influence function. We studied the performance of the adaptive $\tau$-Lasso estimator compared with other competing regularized robust estimators through extensive simulations. Our results indicate that even in the presence of contamination, the class of $\tau$-Lasso estimators, including adaptive $\tau$-Lasso and $\tau$-Lasso, performs reliably and achieves either the best performance or matches it with minimal or no performance loss in terms of RMSE/MAD and variable selection for almost all scenarios. The only exception is the oracle estimators, which assume that the true support of parameter vector $\bb_0$ is known. These results suggest that the adaptive $\tau$-Lasso and $\tau$-Lasso estimators can be effectively utilized for a variety of sparse linear regression problems, particularly in high-dimensional settings and when the data is contaminated by outliers and high-leverage points. It is worth noting that none of the compared estimators has the best performance in all considered scenarios.
  

\vspace{-8.5 pt}

\bibliographystyle{IEEEtran}
\bibliography{refs}

\newpage

\vfill

\newpage


\section*{Supplementary material}

# Supplemental Material for "The Adaptive $\tau$-Lasso: Robustness and Oracle Properties"

Emadaldin Mozafari-Majd, *Graduate Student Member, IEEE,* and Visa Koivunen, *Fellow, IEEE*

This supplemental material is organized as follows. Section S.I provides a background and brief description of the $\tau$-Lasso estimator, including its computation algorithm, along with deriving an upper-bound on the $\tau$-Lasso estimates. In Sections S.II-S.VIII, we provide detailed proofs of Propositions 1 and 2, along with those of Theorems 1-5. Section S.IX demonstrates how to express the adaptive $\tau$-Lasso estimator in the standard form of two-stage regularized $M$-estimators. In Section S.X, we transform the estimating equations of the resulting two-stage regularized $M$-estimator into a population version. Sections S.XI and S.XII contain proofs of Theorems 6 and 7, establishing influence functions of $\tau$-Lasso and adaptive $\tau$-Lasso estimators.

*Remark S1*: We emphasize that our asymptotic results are derived for the global minima of the $\tau$-Lasso and adaptive $\tau$-Lasso programs. Consequently, these results do not necessarily extend to local minima, which may result from gradient-based optimization methods. This is a customary treatment when dealing with non-convex regularized least-squares estimators.

*Remark S2*: It is important to note that $\hat{\boldsymbol{\beta}}_{\text{PT}}$ and $\hat{\boldsymbol{\beta}}_{\text{AT}}$, which are used in establishing consistency and oracle properties, are actually sequences of random vectors. To avoid notational clutter, we adopt the shorthand notation

$$\hat{\boldsymbol{\beta}}_{\text{PT}} := (\hat{\boldsymbol{\beta}}_{\text{PT}})_{n=1}^{\infty} \quad \text{and} \quad \hat{\boldsymbol{\beta}}_{\text{AT}} := (\hat{\boldsymbol{\beta}}_{\text{AT}})_{n=1}^{\infty} \qquad \text{(S1)}$$

to represent these sequences, respectively.

## S.I. Boundedness and computation algorithm for $\tau$-Lasso estimates

### A. Definition

We are given a dataset of $n$ observations that consists of a vector $\mathbf{y} \in \mathbb{R}^n$ of response variables, and a matrix $\mathbf{X} \in \mathbb{R}^{n \times p}$ of predictors, where a proportion of the response variables $y_i$ are contaminated by outliers, a proportion of the predictors $\mathbf{x}_{[i]}$ are contaminated by high-leverage points, or the additive errors are heavy-tailed. We aim to estimate the unknown true coefficient vector $\boldsymbol{\beta}_0$ based on $n$ observations. Under such adverse conditions, when data is also high-dimensional $p > n$, it is necessary to use regularized robust estimators for reliable estimation of the true coefficient vector $\boldsymbol{\beta}_0$. In this way, Martinez-Camara et al. [S1, S2] originally developed the $\tau$-Lasso that robustly estimates the coefficient vector by solving the optimization problem

$$\hat{\boldsymbol{\beta}}_{\text{PT}} = \underset{\boldsymbol{\beta} \in \mathbb{R}^p}{\operatorname{argmin}} \mathcal{L}_n(\boldsymbol{\beta}) = \underset{\boldsymbol{\beta} \in \mathbb{R}^p}{\operatorname{argmin}} \left\{ \tau_n^2(\mathbf{r}(\boldsymbol{\beta})) + \underline{\lambda}_n \|\boldsymbol{\beta}\|_{\ell_1} \right\} \qquad \text{(S2)}$$

where $\underline{\lambda}_n$ is a nonnegative regularization parameter determining the sparsity level for the parameter vector $\boldsymbol{\beta}$, $\mathbf{r}(\boldsymbol{\beta}) = \mathbf{y} - \mathbf{X}\boldsymbol{\beta}$ denotes a vector of residuals and $\tau_n(\mathbf{r}(\boldsymbol{\beta}))$ is an efficient $\tau$-scale as defined by equation (5) within the main body of the paper.

### B. An upper-bound on the $\tau$-Lasso estimates

Here, we establish an upper-bound on $\tau$-Lasso estimates and highlight its implications on robustness. Recalling the definition of the $\tau$-Lasso estimator given in equation (S2), an immediate result of regularization follows, that is, $\hat{\boldsymbol{\beta}}_{\text{PT}}$ is bounded from above by $\hat{\boldsymbol{\beta}}_{\text{T}}$ as follows:

$$\|\hat{\boldsymbol{\beta}}_{\text{PT}}\|_{\ell_1} \leq \|\hat{\boldsymbol{\beta}}_{\text{T}}\|_{\ell_1}. \qquad \text{(S3)}$$

where $\hat{\boldsymbol{\beta}}_{\text{T}}$ denotes the unregularized $\tau$-estimator [S3, S4], equivalent to the $\tau$-Lasso estimator for $\underline{\lambda}_n = 0$, this also implies the finite-sample breakdown point of $\hat{\boldsymbol{\beta}}_{\text{PT}}$ is at least as high as $\hat{\boldsymbol{\beta}}_{\text{T}}$. Proving the boundedness of $\hat{\boldsymbol{\beta}}_{\text{PT}}$ is very straightforward and requires performing some simple algebra. By the definition of the $\tau$-Lasso estimator given by equation (S2), we have

$$\tau_n^2(\mathbf{r}(\hat{\boldsymbol{\beta}}_{\text{PT}})) + \underline{\lambda}_n\|\hat{\boldsymbol{\beta}}_{\text{PT}}\|_{\ell_1} \leq \tau_n^2(\mathbf{r}(\boldsymbol{\beta})) + \underline{\lambda}_n\|\boldsymbol{\beta}\|_{\ell_1} \quad \forall \boldsymbol{\beta} \in \mathbb{R}^p, \qquad \text{(S4)}$$

and hence, combining with the feasibility of $\hat{\boldsymbol{\beta}}_{\text{T}}$, we obtain

$$\tau_n^2(\mathbf{r}(\hat{\boldsymbol{\beta}}_{\text{PT}})) + \underline{\lambda}_n\|\hat{\boldsymbol{\beta}}_{\text{PT}}\|_{\ell_1} \leq \tau_n^2(\mathbf{r}(\hat{\boldsymbol{\beta}}_{\text{T}})) + \underline{\lambda}_n\|\hat{\boldsymbol{\beta}}_{\text{T}}\|_{\ell_1}. \qquad \text{(S5)}$$

Rearranging yields

$$\tau_n^2(\mathbf{r}(\hat{\boldsymbol{\beta}}_{\text{PT}})) - \tau_n^2(\mathbf{r}(\hat{\boldsymbol{\beta}}_{\text{T}})) \leq \underline{\lambda}_n\|\hat{\boldsymbol{\beta}}_{\text{T}}\|_{\ell_1} - \underline{\lambda}_n\|\hat{\boldsymbol{\beta}}_{\text{PT}}\|_{\ell_1}. \qquad \text{(S6)}$$

Now since $\tau_n^2(\mathbf{r}(\boldsymbol{\beta}))$ achieves its minimum at $\hat{\boldsymbol{\beta}}_{\text{T}}$ as given by the definition of the $\tau$-estimator, we have $\tau_n^2(\mathbf{r}(\hat{\boldsymbol{\beta}}_{\text{PT}})) - \tau_n^2(\mathbf{r}(\hat{\boldsymbol{\beta}}_{\text{T}})) \geq 0$

$$\underline{\lambda}_n\|\hat{\boldsymbol{\beta}}_{\text{T}}\|_{\ell_1} - \underline{\lambda}_n\|\hat{\boldsymbol{\beta}}_{\text{PT}}\|_{\ell_1} \geq \tau_n^2(\mathbf{r}(\hat{\boldsymbol{\beta}}_{\text{PT}})) - \tau_n^2(\mathbf{r}(\hat{\boldsymbol{\beta}}_{\text{T}})) \qquad \text{(S7)}$$
$$\geq 0, \qquad \text{(S8)}$$

from which the claim follows, verifying $\|\hat{\boldsymbol{\beta}}_{\text{PT}}\|_{\ell_1} \leq \|\hat{\boldsymbol{\beta}}_{\text{T}}\|_{\ell_1}$.

### C. Computation of the $\tau$-Lasso estimates

In order to compute the $\tau$-Lasso estimates, one needs to minimize the non-convex and non-smooth objective function, given in equation (S2), by taking its generalized gradient w.r.t. $\boldsymbol{\beta}$, denoted by $\partial_{\boldsymbol{\beta}}(\tau_n^2(\mathbf{r}(\boldsymbol{\beta})) + \underline{\lambda}_n\|\boldsymbol{\beta}\|_{\ell_1})$ [S5]. As discussed in our previous work [S6], we have that the sub-gradient of the weighted least-squares penalized by $\ell_1$-norm is equivalent to the generalized gradient of the $\tau$-Lasso objective function for a fixed regularization parameter $\underline{\lambda}_n = \lambda$. Hence, the original

optimization problem may be rewritten in a new form as follows:

$$\hat{\boldsymbol{\beta}}_{\text{PT}} = \underset{\boldsymbol{\beta}}{\operatorname{argmin}} \left\{ \|\boldsymbol{\Omega}(\mathbf{y} - \mathbf{X}\boldsymbol{\beta})\|_{\ell_2}^2 + \bar{\lambda}_n \|\boldsymbol{\beta}\|_{\ell_1} \right\} \quad (S9)$$

where $\bar{\lambda}_n = 2n\underline{\lambda}_n$, $\boldsymbol{\Omega}$ denotes a diagonal matrix with weights $\sqrt{\omega_i}$ on the diagonal. $\omega_i$ is given by

$$\omega_i = \frac{\psi_n(\tilde{r}_i(\boldsymbol{\beta}))}{\tilde{r}_i(\boldsymbol{\beta})} = \frac{\overline{W}_n \psi_0(\tilde{r}_i(\boldsymbol{\beta})) + \psi_1(\tilde{r}_i(\boldsymbol{\beta}))}{\tilde{r}_i(\boldsymbol{\beta})} \quad (S10a)$$

$$\overline{W}_n = \frac{\sum_{i=1}^n [2\rho_1(\tilde{r}_i(\boldsymbol{\beta})) - \psi_1(\tilde{r}_i(\boldsymbol{\beta}))\tilde{r}_i(\boldsymbol{\beta})]}{\sum_{i=1}^n \psi_0(\tilde{r}_i(\boldsymbol{\beta}))\tilde{r}_i(\boldsymbol{\beta})} \quad (S10b)$$

where the notation $\tilde{r}_i(\boldsymbol{\beta})$ is a shorthand for $r_i(\boldsymbol{\beta})/s_n$. Notably, the $\tau$-Lasso estimator and the weighted least-squares penalized by $\ell_1$-norm coincide when Assumption 2 holds. Fulfilling this assumption ensures $\overline{W}_n \geq 0$, which is necessary to keep $\omega_i$ values nonnegative as the real-valued square root of negative $\omega_i$ values does not exist.

Unlike the regularized weighted least-squares estimator, the weights $w_i$ are a function of the unknown $\boldsymbol{\beta}$. In order to deal with this issue, we use iteratively reweighted Lasso (IR-LASSO) by alternating between estimating the weight matrix $\boldsymbol{\Omega}$, refining $\hat{\boldsymbol{\beta}}_{\text{PT}}$, and updating the $M$-scale estimate of residuals $s_n$. We use the $S$-Lasso estimate of $\boldsymbol{\beta}_0$ as the initial estimate for solving the $\tau$-Lasso optimization problem. Interested readers may refer to the Supplementary Material of [S7] and our previous work [S6] for a detailed explanation of the computation algorithm for obtaining initial $S$-Lasso estimates and $M$-scale estimates of residuals. Furthermore, we use the function `dalsql1` of the MATLAB package `DAL` for the Lasso estimation subproblems within IR-LASSO.

## S.II. PROOF OF PROPOSITION 1

We now turn to the proof of Proposition 1, which states that the $\tau$-Lasso estimator $\hat{\boldsymbol{\beta}}_{\text{PT}}$ is a strongly consistent estimator of the true regression coefficient $\boldsymbol{\beta}_0$ when $\underline{\lambda}_n \to 0$. Before proceeding with the proof, we introduce two important notations crucial for understanding the proof. We denote by $r(\boldsymbol{\beta}) \coloneqq y - \mathbf{x}^T\boldsymbol{\beta}$ the error at $\boldsymbol{\beta} \in \mathbb{R}^p$. Secondly, we use $r(\boldsymbol{\beta}_0) \coloneqq y - \mathbf{x}^T\boldsymbol{\beta}_0$ to represent the error at the true coefficient vector $\boldsymbol{\beta}_0 \in \mathbb{R}^p$, which coincides with the random measurement noise variable $u$ that generates the data. Note that both $r(\boldsymbol{\beta})$ and $r(\boldsymbol{\beta}_0)$ are random variables ($r(\boldsymbol{\beta}_0) = u$). We now outline the proof of Proposition 1, which relies on verifying that the following conditions hold

1) the $\tau$-Lasso objective function $\underline{\mathcal{L}}_n(\boldsymbol{\beta})$ given in equation (S2) converges almost surely to the population $\tau$-scale of error at $\boldsymbol{\beta}$ squared, $\tau^2(\boldsymbol{\beta})$, uniformly over any compact set $K$
2) $\hat{\boldsymbol{\beta}}_{\text{PT}}$ is bounded with probability 1

With the above conditions satisfied, strong consistency of $\hat{\boldsymbol{\beta}}_{\text{PT}}$ for $\boldsymbol{\beta}_0$ follows immediately. Note that conditions 1 and 4 of Assumption 3 are not required for Proposition 1.

To proceed, we denote the population $\tau$-scale of error at $\boldsymbol{\beta}$ by $\tau(\boldsymbol{\beta})$ and define it as

$$\tau^2(\boldsymbol{\beta}) = s^2(\boldsymbol{\beta})\mathbb{E}_H\left[\rho_1\left(\frac{y - \mathbf{x}^T\boldsymbol{\beta}}{s(\boldsymbol{\beta})}\right)\right], \quad (S11)$$

where $s(\boldsymbol{\beta})$, the population $M$-scale of error at $\boldsymbol{\beta}$ is given by

$$\mathbb{E}_H\left[\rho_0\left(\frac{y - \mathbf{x}^T\boldsymbol{\beta}}{s(\boldsymbol{\beta})}\right)\right] = \delta. \quad (S12)$$

By Lemma 4.2 of [S3] (Fisher consistency of the $\tau$-estimates), we know that $\tau(\boldsymbol{\beta})$ has a unique minimum at $\boldsymbol{\beta} = \boldsymbol{\beta}_0$. Next, we complete the proof by showing that conditions 1) and 2), stated above, hold for the $\tau$-Lasso estimator.

### A. Verifying condition 1)

We now establish condition 1) using uniform convergence of $\underline{\mathcal{L}}_n(\boldsymbol{\beta})$, which allows us to show that for any compact set $K \subseteq \mathbb{R}^p$,

$$\sup_{\boldsymbol{\beta} \in K} |\underline{\mathcal{L}}_n(\boldsymbol{\beta}) - \tau^2(\boldsymbol{\beta})| \xrightarrow{a.s.} 0. \quad (S13)$$

Recall that the $\tau$-Lasso objective function $\underline{\mathcal{L}}_n(\boldsymbol{\beta})$ consists of $\tau_n^2(\mathbf{r}(\boldsymbol{\beta}))$ and $\underline{\lambda}_n \|\boldsymbol{\beta}\|_{\ell_1}$. By Lemma 4.5 of [S3], we have

$$\sup_{\boldsymbol{\beta} \in K} |\tau_n(\mathbf{r}(\boldsymbol{\beta})) - \tau(\boldsymbol{\beta})| \xrightarrow{a.s.} 0. \quad (S14)$$

On the other hand, $\|\boldsymbol{\beta}\|_{\ell_1}$ is not stochastic and is bounded for any $\boldsymbol{\beta} \in K$. Combined with the assumption $\underline{\lambda}_n \to 0$ for $n \to \infty$, it follows that $\underline{\lambda}_n \|\boldsymbol{\beta}\|_{\ell_1} \to 0$, that is, the second term in $\underline{\mathcal{L}}_n(\boldsymbol{\beta})$ converges uniformly to zero over compact sets. By the continuity of $\tau(\boldsymbol{\beta})$ and strong consistency of $\tau_n(\mathbf{r}(\boldsymbol{\beta}))$ as given by equation (S14), one can conclude that condition 1) holds.

### B. Verifying condition 2)

We now focus on condition 2) by which boundedness of $\hat{\boldsymbol{\beta}}_{\text{PT}}$ with probability one is required. By Theorem 4.1 of [S4], we know that $\hat{\boldsymbol{\beta}}_{\text{T}}$ converges almost surely to the true parameter vector $\boldsymbol{\beta}_0$. Recall that by equation (S3), $\|\hat{\boldsymbol{\beta}}_{\text{PT}}\|_{\ell_1}$ is upper-bounded by $\|\hat{\boldsymbol{\beta}}_{\text{T}}\|_{\ell_1}$. Thus, combining the above, we conclude that $\hat{\boldsymbol{\beta}}_{\text{PT}}$ is bounded with probability one. Hence, our claim is established then by proving the above conditions and $\hat{\boldsymbol{\beta}}_{\text{PT}} \xrightarrow{a.s.} \boldsymbol{\beta}_0$.

## S.III. PROOF OF PROPOSITION 2

Next, we focus on the proof of Proposition 2, which asserts that the adaptive $\tau$-Lasso estimator $\hat{\boldsymbol{\beta}}_{\text{AT}}$ is a strongly consistent estimator of the true regression coefficient $\boldsymbol{\beta}_0$ when $\lambda_n \to 0$ and $\underline{\lambda}_n \to 0$. To proceed, we first provide a short sketch of the proof reasoning. Loosely speaking, we will show that $\tau(\boldsymbol{\beta}_0)$, the population $\tau$-scale of true error, is bounded from above and below by the limit of sequence $\tau_n(\mathbf{r}(\boldsymbol{\beta}))$ each for specific values of $\boldsymbol{\beta}$. To be more precise, the sketch of the proof is as follows:

- *Step 1:* We first show that $\tau(\boldsymbol{\beta}_0)$ at which the true $\tau$-scale attains its minimum value remains lower-bounded almost surely by the limit superior of the sequence $\tau_n(\mathbf{r}(\hat{\boldsymbol{\beta}}_{\text{AT}}))$.
- *Step 2:* In contrast, if we take the infimum of the sequence $\tau_n(\mathbf{r}(\boldsymbol{\beta}))$ over the set of all $\boldsymbol{\beta} \in \mathbb{R}^p$ outside and including the Euclidean ball centered around the true value $\boldsymbol{\beta}_0$ of radius $\epsilon$, denoted by $\mathbb{B}(\boldsymbol{\beta}_0, \epsilon) \equiv \{\boldsymbol{\beta} \in \mathbb{R}^p : \|\boldsymbol{\beta} - \boldsymbol{\beta}_0\|_{\ell_2} \leq \epsilon\}$. We then can show that $\tau(\boldsymbol{\beta}_0)$ is smaller than the limit

inferior of the resulting sequence almost surely for any $\epsilon > 0$, that is,

$$\tau(\boldsymbol{\beta}_0) < \overbrace{\liminf_{n \to \infty} \inf_{\|\boldsymbol{\beta} - \boldsymbol{\beta}_0\|_{\ell_2} \geq \epsilon} \tau_n(\mathbf{r}(\boldsymbol{\beta}))}^{\text{term}(i)} \quad a.s. \quad \forall \epsilon > 0. \tag{S15}$$

where the term highlighted in blue provides a lower-bound for the sequence $\tau_n(\mathbf{r}(\boldsymbol{\beta}))$ within the set $\|\boldsymbol{\beta} - \boldsymbol{\beta}_0\|_{\ell_2} \geq \epsilon$. We then consider the term $(i)$, which takes a liminf of this lower-bound and yields another lower-bound of the resulting sequence as $n$ approaches infinity. The expression on the right-hand side of the inequality can be roughly viewed as the asymptotic lower-bound of the sequence over the set $\|\boldsymbol{\beta} - \boldsymbol{\beta}_0\|_{\ell_2} \geq \epsilon$.

If we can prove that the conditions in steps 1 and 2 hold for the adaptive $\tau$-Lasso estimator. We can then say $\hat{\boldsymbol{\beta}}_{\text{AT}}$ is a strongly consistent estimator of $\boldsymbol{\beta}_0$. Note that conditions 1 and 4 of Assumption 3 are not required for Proposition 2.

### A. Proof of step 1

We will first establish the condition mentioned in step 1 of the proof sketch. By the definition of the adaptive $\tau$-Lasso estimator as given by equation (4) within the main body of the paper and minimality of $\hat{\boldsymbol{\beta}}_{\text{AT}}$, we have

$$\hat{\boldsymbol{\beta}}_{\text{AT}} = \operatorname*{argmin}_{\boldsymbol{\beta} \in \mathbb{R}^p} \left\{ \tau_n^2(\mathbf{r}(\boldsymbol{\beta})) + \lambda_n \sum_{j=1}^p \frac{|\hat{\beta}_j|}{|\tilde{\beta}_j|^\gamma} \right\} \tag{S16}$$

where $\gamma$ is a positive constant. In other words, $\forall \boldsymbol{\beta} \in \mathbb{R}^p$, we have

$$\tau_n^2(\mathbf{r}(\hat{\boldsymbol{\beta}}_{\text{AT}})) + \lambda_n \sum_{j=1}^p \frac{|\hat{\beta}_{\text{AT},j}|}{|\tilde{\beta}_j|^\gamma} \leq \tau_n^2(\mathbf{r}(\boldsymbol{\beta})) + \lambda_n \sum_{j=1}^p \frac{|\beta_j|}{|\tilde{\beta}_j|^\gamma}. \tag{S17}$$

Since the above inequality holds true for $\boldsymbol{\beta} = \boldsymbol{\beta}_0$ as well, we can deduce that

$$\tau_n^2(\mathbf{r}(\hat{\boldsymbol{\beta}}_{\text{AT}})) + \lambda_n \sum_{j=1}^p \frac{|\hat{\beta}_{\text{AT},j}|}{|\tilde{\beta}_j|^\gamma} \leq \tau_n^2(\boldsymbol{\beta}_0) + \lambda_n \sum_{j=1}^p \frac{|\beta_{0,j}|}{|\tilde{\beta}_j|^\gamma}. \tag{S18}$$

On the other hand, we have $\lambda_n \sum_{j=1}^p \frac{|\hat{\beta}_{\text{AT},j}|}{|\tilde{\beta}_j|^\gamma} \geq 0$. Hence, it follows that

$$\tau_n^2(\mathbf{r}(\hat{\boldsymbol{\beta}}_{\text{AT}})) \leq \tau_n^2(\boldsymbol{\beta}_0) + \lambda_n \sum_{j=1}^p \frac{|\beta_{0,j}|}{|\tilde{\beta}_j|^\gamma}. \tag{S19}$$

By setting the initial estimator $\tilde{\boldsymbol{\beta}}$ to be $\hat{\boldsymbol{\beta}}_{\text{PT}}$, as mentioned in Proposition 2 statement, and the strong consistency of $\hat{\boldsymbol{\beta}}_{\text{PT}}$ for $\boldsymbol{\beta}_0$ under the assumptions of Proposition 2, combined with $\lambda_n \to 0$ for $n \to \infty$, we can conclude that

$$\lambda_n \sum_{j=1}^p \frac{|\beta_{0,j}|}{|\tilde{\beta}_j|^\gamma} = \lambda_n \left( \sum_{j=1}^{k_0} \frac{|\beta_{0,j}|}{|\tilde{\beta}_j|^\gamma} + \underbrace{\sum_{j=k_0+1}^p \frac{|\beta_{0,j}|}{|\tilde{\beta}_j|^\gamma}}_{=0} \right)$$

$$\stackrel{(i)}{=} \lambda_n \sum_{j=1}^{k_0} \frac{|\beta_{0,j}|}{|\tilde{\beta}_j|^\gamma} \xrightarrow{a.s.} 0 \tag{S20}$$

where $k_0$ denotes the number of non-zero coefficients of the true parameter vector $\boldsymbol{\beta}_0$ and the equality $(i)$ is a consequence of $\beta_{0,j}$ being equal to zero for $j = k_0+1, \cdots, p$, causing the second term on the right-hand side of the top equation to vanish. By the Lemma 4.5 of [S3] and the almost sure convergence of $\lambda_n \sum_{j=1}^p |\beta_{0,j}|/|\tilde{\beta}_j|^\gamma$ to zero as stated by equation (S20), we now get

$$\tau_n^2(\boldsymbol{\beta}_0) + \lambda_n \sum_{j=1}^p \frac{|\beta_{0,j}|}{|\tilde{\beta}_j|^\gamma} \xrightarrow{a.s.} \tau^2(\mathbf{r}(\boldsymbol{\beta}_0)). \tag{S21}$$

Thus, from the above statements in equations (S19) and (S21), we conclude that the limit superior of the sequence $\tau_n^2(\mathbf{r}(\hat{\boldsymbol{\beta}}_{\text{AT}}))$ is upper-bounded almost surely as follows:

$$\limsup_{n \to \infty} \tau_n^2(\mathbf{r}(\hat{\boldsymbol{\beta}}_{\text{AT}})) \leq \tau^2(\boldsymbol{\beta}_0) \quad a.s. \tag{S22}$$

### B. Proof of step 2

Moving forward with the proof, we now prove the condition mentioned in step 2 of the proof sketch. Using Lemma 4.2 and 4.5 of [S3], we can infer that

$$\tau^2(\boldsymbol{\beta}_0) < \liminf_{n \to \infty} \inf_{\|\boldsymbol{\beta} - \boldsymbol{\beta}_0\|_{\ell_2} \geq \epsilon} \tau_n^2(\mathbf{r}(\boldsymbol{\beta})) \quad a.s. \quad \text{for any } \epsilon > 0 \tag{S23}$$

which, together with the fact stated in equation (S22), completes the proof, and we have

$$\hat{\boldsymbol{\beta}}_{\text{AT}} \xrightarrow{a.s.} \boldsymbol{\beta}_0, \tag{S24}$$

that is, the adaptive $\tau$-Lasso is a strongly consistent estimator of $\boldsymbol{\beta}_0$.

## S.IV. PROOF OF THEOREM 1

To establish the root-$n$ consistency of the $\tau$-Lasso estimator, we provide a high-level sketch of the proof. Note that our objective is to ultimately obtain an inequality that states $\sqrt{n}\|\hat{\boldsymbol{\beta}}_{\text{PT}} - \boldsymbol{\beta}_0\|_{\ell_2}$ is upper-bounded with arbitrarily high probability for sufficiently large $n$. Alternatively, we can express it as $\sqrt{n}\|\hat{\boldsymbol{\beta}}_{\text{PT}} - \boldsymbol{\beta}_0\|_{\ell_2} \leq$ some positive constant with arbitrarily high probability for sufficiently large $n$. This is equivalent to saying that $\sqrt{n}(\hat{\boldsymbol{\beta}}_{\text{PT}} - \boldsymbol{\beta}_0) = O_P(1)$, implying the root-$n$ consistency of $\hat{\boldsymbol{\beta}}_{\text{PT}}$. We begin the proof by leveraging the optimality of $\hat{\boldsymbol{\beta}}_{\text{PT}}$, which ensures that the $\tau$-Lasso objective function $\underline{\mathcal{L}}_n(\hat{\boldsymbol{\beta}}_{\text{PT}})$ is smaller than or equal to $\underline{\mathcal{L}}_n(\boldsymbol{\beta})$ for any $\boldsymbol{\beta} \in \mathbb{R}^P$, including $\boldsymbol{\beta} = \boldsymbol{\beta}_0$. This leads to the inequality

$$\underline{\mathcal{L}}_n(\hat{\boldsymbol{\beta}}_{\text{PT}}) \leq \underline{\mathcal{L}}_n(\boldsymbol{\beta}_0) \tag{S25}$$

or equivalently,

$$\underline{\mathcal{L}}_n(\hat{\boldsymbol{\beta}}_{\text{PT}}) - \underline{\mathcal{L}}_n(\boldsymbol{\beta}_0) \leq 0. \tag{S26}$$

Next, we apply the first-order Taylor series expansion around $\boldsymbol{\beta}_0$ for the smooth term within the left-hand side of inequality (S26), resulting in a simplified expression in terms of $(\hat{\boldsymbol{\beta}}_{\text{PT}} - \boldsymbol{\beta}_0)$. By rearranging and analyzing the asymptotic behavior of the derived terms, we obtain an expression involving $\|\hat{\boldsymbol{\beta}}_{\text{PT}} - \boldsymbol{\beta}_0\|_{\ell_2}$. Finally, we will establish an equality via algebraic manipulations that indicates with high probability,

$\sqrt{n}\|\hat{\boldsymbol{\beta}}_{\text{PT}} - \boldsymbol{\beta}_0\|_{\ell_2}$ remains upper-bounded by a positive constant for sufficiently large $n$, thereby proving the claim. While the proof is conceptually straightforward, it is important to note that it is an extremely tedious process.

We will now construct a proof of the root-$n$ consistency for $\hat{\boldsymbol{\beta}}_{\text{PT}}$ by following the below steps.

- *Step 1:* We will begin by exploiting the optimality of $\hat{\boldsymbol{\beta}}_{\text{PT}}$, which implies the $\tau$-Lasso objective function $\underline{\mathcal{L}}_n$ attains its minimum value at $\hat{\boldsymbol{\beta}}_{\text{PT}}$. In other words, we have $\underline{\mathcal{L}}_n(\hat{\boldsymbol{\beta}}_{\text{PT}}) - \underline{\mathcal{L}}_n(\boldsymbol{\beta}_0) \leq 0$, as expressed in inequality (S26).
- *Step 2:* We shall now write the first-order Taylor series expansion for the smooth term (continuously differentiable term) $\tau_n^2(\mathbf{r}(\hat{\boldsymbol{\beta}}_{\text{PT}}))$ of $\underline{\mathcal{L}}_n(\hat{\boldsymbol{\beta}}_{\text{PT}}) - \underline{\mathcal{L}}_n(\boldsymbol{\beta}_0)$ around the true parameter vector $\boldsymbol{\beta}_0$. This will allow us to obtain a simplified expression in terms of $(\hat{\boldsymbol{\beta}}_{\text{PT}} - \boldsymbol{\beta}_0)$, which will enable us to establish the root-$n$ consistency. Opting for the Taylor series expansion around $\boldsymbol{\beta}_0$ is justified by the strong consistency of $\hat{\boldsymbol{\beta}}_{\text{PT}}$ for $\boldsymbol{\beta}_0$.
- *Step 3:* Due to the difficulty of analyzing the asymptotic behavior of certain terms, we will expand those terms around $i_{\text{th}}$ element of the true error vector $\mathbf{u}$, $u_i$, via the first-order Taylor series to derive a more manageable form. We then carefully rearrange the expression deduced from the application of Taylor series expansion.
- *Step 4:* Our next step is to examine the asymptotic behavior of the stochastic terms in the expression obtained in steps 2 and 3, along with the non-smooth term, within the left-hand side of inequality (S26). Using the Strong Law of Large Numbers, the Central Limit Theorem (CLT), and other convergence theorems and lemmas, which will be mentioned later, we will show that some of the stochastic terms converge almost surely to their expected values, implying boundedness in probability $O_P(1)$ while others converge almost surely to zero, implying $o_P(1)$ convergence.
- *Step 5:* We will then substitute $\underline{\mathcal{L}}_n(\hat{\boldsymbol{\beta}}_{\text{PT}}) - \underline{\mathcal{L}}_n(\boldsymbol{\beta}_0)$ with the resulting equation from the previous step. The remainder of our proof will isolate the terms involving $\sqrt{n}\|\hat{\boldsymbol{\beta}}_{\text{PT}} - \boldsymbol{\beta}_0\|_{\ell_2}$ on the left-hand side of the inequality and move the remaining terms to the right-hand side of the inequality. By doing so, it suffices to solve the inequality for $\sqrt{n}\|\hat{\boldsymbol{\beta}}_{\text{PT}} - \boldsymbol{\beta}_0\|_{\ell_2}$, which can be achieved by canceling out the coefficient of $\sqrt{n}\|\hat{\boldsymbol{\beta}}_{\text{PT}} - \boldsymbol{\beta}_0\|_{\ell_2}$. We then conclude that $\sqrt{n}\|\hat{\boldsymbol{\beta}}_{\text{PT}} - \boldsymbol{\beta}_0\|_{\ell_2}$ is bounded with arbitrarily high probability for sufficiently large $n$, in turn, implying that $\sqrt{n}\|\hat{\boldsymbol{\beta}}_{\text{PT}} - \boldsymbol{\beta}_0\|_{\ell_2}$ is bounded in probability or, equivalently, $\sqrt{n}(\hat{\boldsymbol{\beta}}_{\text{PT}} - \boldsymbol{\beta}_0) = O_P(1)$.

## A. Proof of step 1

We will first make use of the $\tau$-Lasso estimator definition by which $\hat{\boldsymbol{\beta}}_{\text{PT}}$ is the minimizer of the objective function $\underline{\mathcal{L}}_n(\boldsymbol{\beta})$ as follows:

$$\hat{\boldsymbol{\beta}}_{\text{PT}} = \underset{\boldsymbol{\beta}\in\mathbb{R}^p}{\operatorname{argmin}} \underline{\mathcal{L}}_n(\boldsymbol{\beta}) = \underset{\boldsymbol{\beta}\in\mathbb{R}^p}{\operatorname{argmin}} \left\{ \tau_n^2(\mathbf{r}(\boldsymbol{\beta})) + \underline{\lambda}_n\|\boldsymbol{\beta}\|_{\ell_1} \right\},$$
(S27)

which implies

$$\tau_n^2(\mathbf{r}(\hat{\boldsymbol{\beta}}_{\text{PT}})) + \underline{\lambda}_n\|\hat{\boldsymbol{\beta}}_{\text{PT}}\|_{\ell_1} \leq \tau_n^2(\mathbf{r}(\boldsymbol{\beta})) + \underline{\lambda}_n\|\boldsymbol{\beta}\|_{\ell_1}, \quad \forall \boldsymbol{\beta}\in\mathbb{R}^p.$$
(S28)

Combined with the feasibility of $\boldsymbol{\beta}_0$, we find that

$$\underbrace{\tau_n^2(\mathbf{r}(\hat{\boldsymbol{\beta}}_{\text{PT}})) + \underline{\lambda}_n\|\hat{\boldsymbol{\beta}}_{\text{PT}}\|_{\ell_1}}_{\underline{\mathcal{L}}_n(\hat{\boldsymbol{\beta}}_{\text{PT}})} \leq \underbrace{\tau_n^2(\mathbf{r}(\boldsymbol{\beta}_0)) + \underline{\lambda}_n\|\boldsymbol{\beta}_0\|_{\ell_1}}_{\underline{\mathcal{L}}_n(\boldsymbol{\beta}_0)}. \quad (\text{S29})$$

Rearranging the terms yields the inequality

$$\underbrace{\tau_n^2(\mathbf{r}(\hat{\boldsymbol{\beta}}_{\text{PT}})) - \tau_n^2(\mathbf{r}(\boldsymbol{\beta}_0)) + \underline{\lambda}_n\|\hat{\boldsymbol{\beta}}_{\text{PT}}\|_{\ell_1} - \underline{\lambda}_n\|\boldsymbol{\beta}_0\|_{\ell_1}}_{\underline{\mathcal{L}}_n(\hat{\boldsymbol{\beta}}_{\text{PT}}) - \underline{\mathcal{L}}_n(\boldsymbol{\beta}_0)} \leq 0.$$
(S30)

## B. Proof of step 2

Now let us consider $\underline{\mathcal{L}}_n(\hat{\boldsymbol{\beta}}_{\text{PT}}) - \underline{\mathcal{L}}_n(\boldsymbol{\beta}_0)$, which corresponds to the left-hand side of inequality (S30). In particular, we will focus on the smooth term $\tau_n^2(\mathbf{r}(\hat{\boldsymbol{\beta}}_{\text{PT}}))$ and proceed to apply a first-order Taylor series expansion to this term, highlighted in red in the following equation.

$$\underline{\mathcal{L}}_n(\hat{\boldsymbol{\beta}}_{\text{PT}}) - \underline{\mathcal{L}}_n(\boldsymbol{\beta}_0)$$
$$= \tau_n^2(\mathbf{r}(\hat{\boldsymbol{\beta}}_{\text{PT}})) - \tau_n^2(\mathbf{r}(\boldsymbol{\beta}_0)) + \underbrace{(\underline{\lambda}_n\|\hat{\boldsymbol{\beta}}_{\text{PT}}\|_{\ell_1} - \underline{\lambda}_n\|\boldsymbol{\beta}_0\|_{\ell_1})}_{B_n}$$
(S31)

Next, we set

$$\mathbf{v}_n = \kappa(\hat{\boldsymbol{\beta}}_{\text{PT}} - \boldsymbol{\beta}_0), \text{ with } 0 < \kappa < 1 \quad (\text{S32})$$

and

$$\boldsymbol{\beta}_n^* = \boldsymbol{\beta}_0 + \mathbf{v}_n \quad (\text{S33})$$

where $\boldsymbol{\beta}_n^*$ falls on the line segment connecting $\hat{\boldsymbol{\beta}}_{\text{PT}}$ and $\boldsymbol{\beta}_0$. Using equations (S32) and (S33), we will form a first-order Taylor series expansion of $\tau_n^2(\mathbf{r}(\hat{\boldsymbol{\beta}}_{\text{PT}}))$ around $\boldsymbol{\beta}_0$ as follows:

$$\tau_n^2(\mathbf{r}(\hat{\boldsymbol{\beta}}_{\text{PT}})) = \tau_n^2(\mathbf{r}(\boldsymbol{\beta}_0)) + (\hat{\boldsymbol{\beta}}_{\text{PT}} - \boldsymbol{\beta}_0)^T$$
$$\times \left[ \frac{\partial s_n(\mathbf{r}(\boldsymbol{\beta}))}{\partial \boldsymbol{\beta}} \frac{2 s_n(\mathbf{r}(\boldsymbol{\beta}))}{n} \sum_{i=1}^n \rho_1\left(\frac{r_i(\boldsymbol{\beta})}{s_n(\mathbf{r}(\boldsymbol{\beta}))}\right) \right.$$
$$+ \frac{1}{n} \sum_{i=1}^n \psi_1\left(\frac{r_i(\boldsymbol{\beta})}{s_n(\mathbf{r}(\boldsymbol{\beta}))}\right)$$
$$\left. \times \left[ -s_n(\mathbf{r}(\boldsymbol{\beta})) \mathbf{x}_{[i]} - r_i(\boldsymbol{\beta}) \frac{\partial s_n(\mathbf{r}(\boldsymbol{\beta}))}{\partial \boldsymbol{\beta}} \right] \right]_{\boldsymbol{\beta} = \boldsymbol{\beta}_n^*}$$
$$= \tau_n^2(\mathbf{r}(\boldsymbol{\beta}_0)) + (\hat{\boldsymbol{\beta}}_{\text{PT}} - \boldsymbol{\beta}_0)^T$$
$$\times \left[ -2 \frac{\frac{1}{n}\sum_{i=1}^n \rho_1\left(\frac{u_i - \mathbf{x}_{[i]}^T \mathbf{v}_n}{s_n(\mathbf{r}(\boldsymbol{\beta}_n^*))}\right)}{\frac{1}{n}\sum_{i=1}^n \psi_0\left(\frac{u_i - \mathbf{x}_{[i]}^T \mathbf{v}_n}{s_n(\mathbf{r}(\boldsymbol{\beta}_n^*))}\right) [u_i - \mathbf{x}_{[i]}^T \mathbf{v}_n]} \right.$$
$$\times \frac{s_n^2(\mathbf{r}(\boldsymbol{\beta}_n^*))}{n} \sum_{i=1}^n \psi_0\left(\frac{u_i - \mathbf{x}_{[i]}^T \mathbf{v}_n}{s_n(\mathbf{r}(\boldsymbol{\beta}_n^*))}\right) \mathbf{x}_{[i]}$$
$$+ \frac{-s_n(\mathbf{r}(\boldsymbol{\beta}_n^*))}{n} \sum_{i=1}^n \psi_1\left(\frac{u_i - \mathbf{x}_{[i]}^T \mathbf{v}_n}{s_n(\mathbf{r}(\boldsymbol{\beta}_n^*))}\right) \mathbf{x}_{[i]}$$
$$+ \frac{\left(\frac{-1}{n}\sum_{i=1}^n \psi_1\left(\frac{u_i - \mathbf{x}_{[i]}^T \mathbf{v}_n}{s_n(\mathbf{r}(\boldsymbol{\beta}_n^*))}\right)[u_i - \mathbf{x}_{[i]}^T \mathbf{v}_n]\right)}{\left(\frac{1}{n}\sum_{i=1}^n \psi_0\left(\frac{u_i - \mathbf{x}_{[i]}^T \mathbf{v}_n}{s_n(\mathbf{r}(\boldsymbol{\beta}_n^*))}\right)[u_i - \mathbf{x}_{[i]}^T \mathbf{v}_n]\right)}$$
$$\left. \times \frac{-s_n(\mathbf{r}(\boldsymbol{\beta}_n^*))}{n} \sum_{i=1}^n \psi_0\left(\frac{u_i - \mathbf{x}_{[i]}^T \mathbf{v}_n}{s_n(\mathbf{r}(\boldsymbol{\beta}_n^*))}\right) \mathbf{x}_{[i]} \right]. \quad \text{(S34)}$$

Rearranging and a simple algebraic manipulation yields

$$\tau_n^2(\mathbf{r}(\hat{\boldsymbol{\beta}}_{\text{PT}})) = \tau_n^2(\mathbf{r}(\boldsymbol{\beta}_0)) + (\hat{\boldsymbol{\beta}}_{\text{PT}} - \boldsymbol{\beta}_0)^T$$
$$\times \left[ -2 \underbrace{\frac{\overbrace{\frac{1}{n}\sum_{i=1}^n \rho_1\left(\frac{u_i - \mathbf{x}_{[i]}^T \mathbf{v}_n}{s_n(\mathbf{r}(\boldsymbol{\beta}_n^*))}\right)}^{Z_n^1}}{\frac{1}{n}\sum_{i=1}^n \psi_0\left(\frac{u_i - \mathbf{x}_{[i]}^T \mathbf{v}_n}{s_n(\mathbf{r}(\boldsymbol{\beta}_n^*))}\right) \frac{u_i - \mathbf{x}_{[i]}^T \mathbf{v}_n}{s_n(\mathbf{r}(\boldsymbol{\beta}_n^*))}}}_{A_n^0} \right.$$
$$\times \frac{s_n(\mathbf{r}(\boldsymbol{\beta}_n^*))}{n} \sum_{i=1}^n \psi_0\left(\frac{u_i - \mathbf{x}_{[i]}^T \mathbf{v}_n}{s_n(\mathbf{r}(\boldsymbol{\beta}_n^*))}\right) \mathbf{x}_{[i]}$$
$$+ \frac{-s_n(\mathbf{r}(\boldsymbol{\beta}_n^*))}{n} \sum_{i=1}^n \psi_1\left(\frac{u_i - \mathbf{x}_{[i]}^T \mathbf{v}_n}{s_n(\mathbf{r}(\boldsymbol{\beta}_n^*))}\right) \mathbf{x}_{[i]}$$
$$+ \overbrace{\frac{\left(\frac{-1}{n}\sum_{i=1}^n \psi_1\left(\frac{u_i - \mathbf{x}_{[i]}^T \mathbf{v}_n}{s_n(\mathbf{r}(\boldsymbol{\beta}_n^*))}\right) \frac{u_i - \mathbf{x}_{[i]}^T \mathbf{v}_n}{s_n(\mathbf{r}(\boldsymbol{\beta}_n^*))}\right)}{\underbrace{\left(\frac{1}{n}\sum_{i=1}^n \psi_0\left(\frac{u_i - \mathbf{x}_{[i]}^T \mathbf{v}_n}{s_n(\mathbf{r}(\boldsymbol{\beta}_n^*))}\right) \frac{u_i - \mathbf{x}_{[i]}^T \mathbf{v}_n}{s_n(\mathbf{r}(\boldsymbol{\beta}_n^*))}\right)}_{A_n^0}}}^{-A_n^1}$$
$$\left. \times \frac{-s_n(\mathbf{r}(\boldsymbol{\beta}_n^*))}{n} \sum_{i=1}^n \psi_0\left(\frac{u_i - \mathbf{x}_{[i]}^T \mathbf{v}_n}{s_n(\mathbf{r}(\boldsymbol{\beta}_n^*))}\right) \mathbf{x}_{[i]} \right] \quad \text{(S35)}$$

In order to further simplify the above Taylor series expansion of $\tau_n^2(\mathbf{r}(\hat{\boldsymbol{\beta}}_{\text{PT}}))$, we will use the shorthand notation and substitute the underlined terms with $A_n^0$ and overlined terms with $Z_n^1$, $-A_n^1$, respectively, as follows:

$$\tau_n^2(\mathbf{r}(\hat{\boldsymbol{\beta}}_{\text{PT}})) = \tau_n^2(\mathbf{r}(\boldsymbol{\beta}_0)) + \left[ \underbrace{\left(\frac{2Z_n^1 - A_n^1}{A_n^0}\right)}_{W_n^0} \right.$$
$$\times \underbrace{\frac{-s_n(\mathbf{r}(\boldsymbol{\beta}_n^*))}{n\kappa} \sum_{i=1}^n \psi_0\left(\frac{u_i - \mathbf{x}_{[i]}^T \mathbf{v}_n}{s_n(\mathbf{r}(\boldsymbol{\beta}_n^*))}\right) \mathbf{x}_{[i]}^T \mathbf{v}_n}_{-R_n^0}$$
$$\left. + \underbrace{\frac{-s_n(\mathbf{r}(\boldsymbol{\beta}_n^*))}{n\kappa} \sum_{i=1}^n \psi_1\left(\frac{u_i - \mathbf{x}_{[i]}^T \mathbf{v}_n}{s_n(\mathbf{r}(\boldsymbol{\beta}_n^*))}\right) \mathbf{x}_{[i]}^T \mathbf{v}_n}_{-R_n^1} \right] \quad \text{(S36)}$$

which allows us to re-express $\tau_n^2(\mathbf{r}(\hat{\boldsymbol{\beta}}_{\text{PT}}))$ in terms of $(\hat{\boldsymbol{\beta}}_{\text{PT}} - \boldsymbol{\beta}_0)$ using the relation $\mathbf{v}_n = \kappa(\hat{\boldsymbol{\beta}}_{\text{PT}} - \boldsymbol{\beta}_0)$. The constant $\kappa$ lies within the interval $(0, 1)$ as defined in equation (S32). Analogously, we will denote the underlined terms by $W_n^0$, $-R_n^0$, and $-R_n^1$, respectively.

### C. Proof of step 3

We now proceed to the remaining terms $R_n^0$ and $R_n^1$ as given by equation (S36) and establish their asymptotic behavior. In order to deal with the term $R_n^0$, $\psi_0((u_i - \mathbf{x}_{[i]}^T \mathbf{v}_n)/s_n(\mathbf{r}(\boldsymbol{\beta}_n^*)))$ is expanded by using Taylor series around $u_i$. We then get

$$R_n^0 = s_n(\mathbf{r}(\boldsymbol{\beta}_n^*)) \left( \frac{1}{n\kappa} \sum_{i=1}^n \psi_0\left(\frac{u_i}{s_n(\mathbf{r}(\boldsymbol{\beta}_n^*))}\right) \mathbf{x}_{[i]}^T \mathbf{v}_n \right.$$
$$\left. + \frac{-1}{s_n(\mathbf{r}(\boldsymbol{\beta}_n^*))n\kappa} \sum_{i=1}^n \psi_0'\left(\frac{u_i - \underline{\kappa}\mathbf{x}_{[i]}^T \mathbf{v}_n}{s_n(\mathbf{r}(\boldsymbol{\beta}_n^*))}\right) \mathbf{v}_n^T \mathbf{x}_{[i]} \mathbf{x}_{[i]}^T \mathbf{v}_n \right)$$
$$\stackrel{(i)}{=} \overbrace{\frac{(\hat{\boldsymbol{\beta}}_{\text{PT}} - \boldsymbol{\beta}_0)^T}{\sqrt{n}} \left[ \frac{s_n(\mathbf{r}(\boldsymbol{\beta}_n^*))}{\sqrt{n}} \sum_{i=1}^n \psi_0\left(\frac{u_i}{s_n(\mathbf{r}(\boldsymbol{\beta}_n^*))}\right) \mathbf{x}_{[i]} \right]}^{E_n^0}$$
$$- \underbrace{\kappa(\hat{\boldsymbol{\beta}}_{\text{PT}} - \boldsymbol{\beta}_0)^T \left[ \frac{1}{n} \sum_{i=1}^n \psi_0'\left(\frac{u_i - \underline{\kappa}\mathbf{x}_{[i]}^T \mathbf{v}_n}{s_n(\mathbf{r}(\boldsymbol{\beta}_n^*))}\right) \mathbf{x}_{[i]} \mathbf{x}_{[i]}^T \right] (\hat{\boldsymbol{\beta}}_{\text{PT}} - \boldsymbol{\beta}_0)}_{D_n^0}$$
$$\quad \text{(S37)}$$

where $u_i - \underline{\kappa}\mathbf{x}_{[i]}^T \mathbf{v}_n$ lies on the line segment connecting $u_i$ and $u_i - \mathbf{x}_{[i]}^T \mathbf{v}_n$ with $0 < \underline{\kappa} < 1$ and the equality $(i)$ follows from substituting $\mathbf{v}_n = \kappa(\hat{\boldsymbol{\beta}}_{\text{PT}} - \boldsymbol{\beta}_0)$ into the right-hand side of the first equality.

### D. Proof of step 4

In what follows, we will deal with the asymptotic behavior of the underlined term $W_n^0$, given by equation (S36), as well as the underlined term $D_n^0$ and the overlined term $E_n^0$ in the above

expression. We will now analyze the asymptotic behavior of each term separately, as follows:

- *Asymptotic behavior of $W_n^0$:* By Proposition 1, we have that $\hat{\boldsymbol{\beta}}_{\text{PT}}$ is a strongly consistent estimator of $\boldsymbol{\beta}_0$. Thus, the random sequence $\mathbf{v}_n$ converges almost surely to zero, i.e., $\mathbf{v}_n \xrightarrow{a.s.} 0$. Using part (a) of Lemma 3 of [S8], we have that $s_n(\mathbf{r}(\boldsymbol{\beta}_n^*)) \xrightarrow{a.s.} s(\boldsymbol{\beta}_0)$. By the Strong Law of Large Numbers, the Continuous Mapping Theorem, and the continuity and boundedness of $\rho_1(.)$, the term $Z_n^1$ converges almost surely to $\mathbb{E}_F\big[\rho_1(u/s(\boldsymbol{\beta}_0))\big]$. Combined with part (b) of Lemma 3 of [S8], we then obtain $W_n^0 \xrightarrow{a.s.} W^0$. We define $W^0$, that is, the population analog of $W_n^0$ evaluated at true regression vector $\boldsymbol{\beta}_0$ as follows:

$$W^0 = \frac{2\mathbb{E}_F\big[\rho_1\big(\frac{u}{s(\boldsymbol{\beta}_0)}\big)\big] - \mathbb{E}_F\big[\psi_1\big(\frac{u}{s(\boldsymbol{\beta}_0)}\big)\frac{u}{s(\boldsymbol{\beta}_0)}\big]}{\mathbb{E}_F\big[\psi_0\big(\frac{u}{s(\boldsymbol{\beta}_0)}\big)\frac{u}{s(\boldsymbol{\beta}_0)}\big]} \quad \text{(S38)}$$

where $u$ denotes the random measurement noise variable.
*Remark S3:* Note that both $W_n^0$ and $W^0$ are closely related to $\overline{W}_n$ and $\overline{W}$ appeared in equations (S10) and (8) within the main body of the paper, respectively.

- *Asymptotic behavior of $D_n^0$:* (defined in equation (S37)) By the Lemma 4.2 of [S9], Proposition 1, strong consistency of $s_n(\mathbf{r}(\boldsymbol{\beta}_n^*)) \xrightarrow{a.s.} s(\boldsymbol{\beta}_0)$, condition 4 of assumption 3, and the Strong Law of Large Numbers, we can show that

$$\frac{1}{n}\sum_{i=1}^{n} \psi_0'\bigg(\frac{u_i - \underline{\kappa}\mathbf{x}_{[i]}^T\mathbf{v}_n}{s_n(\mathbf{r}(\boldsymbol{\beta}_n^*))}\bigg)\mathbf{x}_{[i]}\mathbf{x}_{[i]}^T$$
$$\xrightarrow{a.s.} \mathbb{E}_H\big[\psi_0'\big(\frac{u}{s(\boldsymbol{\beta}_0)}\big)\mathbf{x}\mathbf{x}^T\big] \quad \text{(S39)}$$

The left-hand and right-hand sides are the robust analog of the empirical (or sample) second-moment matrix and its population version, respectively. Using the results in the proof of Proposition 2 in [S10], we derive a lower bound of $D_n^0$ as follows:

$$D_n^0 \geq \overline{D}_n^0 \|\hat{\boldsymbol{\beta}}_{\text{PT}} - \boldsymbol{\beta}_0\|_{\ell_2}^2 \quad \text{(S40)}$$

that holds for some sequence $\overline{D}_n^0$ that almost surely converges to its expected value $\overline{D}^0 > 0$ or $\overline{D}_n^0 \xrightarrow{a.s.} \overline{D}^0 > 0$.

- *Asymptotic behavior of $E_n^0$:* (defined in equation (S37)) On the other hand, it follows from Lemma 5.1 of [S9], the results in the proof of Theorem 3 in [S8] and strong consistency of $s_n(\mathbf{r}(\boldsymbol{\beta}_n^*)) \xrightarrow{a.s.} s(\boldsymbol{\beta}_0)$ that $E_n^0 := O_P(1)$. Alternatively, one can say $|E_n^0|$ is bounded from above by some $E^0 > 0$ with arbitrarily high probability for sufficiently large $n$. Hence, by applying the Cauchy-Schwarz inequality to the first term in equation (S37) and taking into account the boundedness of $E_n^0$ in probability, we can conclude that

$$\frac{(\hat{\boldsymbol{\beta}}_{\text{PT}} - \boldsymbol{\beta}_0)^T}{\sqrt{n}}E_n^0 \leq \frac{1}{\sqrt{n}}\|\hat{\boldsymbol{\beta}}_{\text{PT}} - \boldsymbol{\beta}_0\|_{\ell_2}\|E_n^0\|_{\ell_2}$$
$$\leq \frac{E^0}{\sqrt{n}}\|\hat{\boldsymbol{\beta}}_{\text{PT}} - \boldsymbol{\beta}_0\|_{\ell_2} \quad \text{(S41)}$$

with arbitrarily high probability for large enough $n$.

Similar results can be derived for $R_n^1$, $E_n^1$ and $D_n^1$ in equation (S36). Note that one can readily obtain $R_n^1$ from equation (S37) by replacing each occurrence of $\psi_0(.)$ and $\psi_0'(.)$ with $\psi_1(.)$ and $\psi_1'(.)$, respectively. Consequently, $E_n^1$ and $D_n^1$ correspond to the terms resulting from these replacements within $E_n^0$ and $D_n^0$, respectively.

Next, we establish a lower bound of the difference-of-norms term $B_n = \underline{\lambda}_n(\|\hat{\boldsymbol{\beta}}_{\text{PT}}\|_{\ell_1} - \|\boldsymbol{\beta}_0\|_{\ell_1})$ in equation (S31) and characterize its asymptotic behavior. By definition, we have

$$B_n = \underline{\lambda}_n \sum_{j=1}^{p} |\hat{\beta}_{\text{PT},j}| - |\beta_{0,j}|$$
$$\overset{(i)}{\geq} \underline{\lambda}_n \sum_{j=1}^{k_0} |\hat{\beta}_{\text{PT},j}| - |\beta_{0,j}| \quad \text{(S42)}$$

where the inequality $(i)$ uses the assumption that $\beta_{0,j} = 0$ for $j = k_0 + 1, \cdots, p$. We now turn our attention to simplifying the summation term on the right-hand side of inequality (S42). To do so, we exploit the strong consistency of $\hat{\boldsymbol{\beta}}_{\text{PT}}$ for $\boldsymbol{\beta}_0$, as demonstrated by Proposition 1, and subsequently apply the mean-value theorem to each of the terms in this summation. We must first ensure that every term in the summation satisfies the differentiability condition of the mean-value theorem within the given interval. More specifically, we need to show that the absolute function $|\beta_j|$ is differentiable along the entire interval $(\beta_{0,j}, \hat{\beta}_{\text{PT},j})$.

To gain a deeper understanding of strong consistency, we elaborate on this concept. We define an estimator $\hat{\boldsymbol{\beta}}_{\text{PT}}$ to be strongly consistent for $\boldsymbol{\beta}_0$ if, given any $\epsilon > 0$, no matter how small, there exists a sufficiently large sample size $n = n_0$ such that

$$\|\hat{\boldsymbol{\beta}}_{\text{PT}} - \boldsymbol{\beta}_0\|_{\ell_2} < \epsilon \quad \text{(S43)}$$

with probability one for all elements of the sequence $\hat{\boldsymbol{\beta}}_{\text{PT}}$, including sequence indices $n_0, n_0 + 1, n_0 + 2, \cdots$ (almost everywhere in the sample space).

Considering that $\beta_{0,j}$ is non-zero for all $j = 1, \cdots, k_0$ within the summation term, the strong consistency of $\hat{\boldsymbol{\beta}}_{\text{PT}}$ for $\boldsymbol{\beta}_0$ ensures two important outcomes. First, the first $k_0$ non-zero coefficients of $\hat{\boldsymbol{\beta}}_{\text{PT}}$ stay away from zero with probability one for sufficiently large $n$ (almost everywhere in the sample space). Simultaneously, $\hat{\beta}_{\text{PT},j}$ gets arbitrarily close to $\beta_{0,j}$ for sufficiently large $n$ with probability one, guaranteeing that both coefficients have the same sign for all $j = 1, \cdots, k_0$. As a result of this behavior, the absolute value function $|\beta_j|$ remains differentiable along the entire line segment connecting $\beta_{0,j}$ and $\hat{\beta}_{\text{PT},j}$ for all $j = 1, \cdots, k_0$ since it does not cross zero with probability one for sufficiently large $n$. Therefore, the application of the mean-value theorem to the right-hand side of inequality (S42) is completely valid with probability one for sufficiently large $n$.

By individually applying the mean-value theorem to each term within the summation on the right-hand side of inequality (S42) and subsequently combining it with the strong consistency of $\hat{\boldsymbol{\beta}}_{\text{PT}}$, we find that

$$\underline{\lambda}_n \sum_{j=1}^{k_0} |\hat{\beta}_{\text{PT},j}| - |\beta_{0,j}| = \underline{\lambda}_n \sum_{j=1}^{k_0} \text{sgn}(\bar{\beta}_j)(\hat{\beta}_{\text{PT},j} - \beta_{0,j}) \quad \text{(S44)}$$

that holds with probability one for sufficiently large $n$ (almost everywhere in the sample space). Here, the term $\text{sgn}(\bar{\beta}_j)$ corresponds to the derivative of $|\beta_j|$ evaluated at $\bar{\beta}_j$ when $\beta_j \neq 0$ and $\bar{\beta}_j$ lies on the line segment connecting $\hat{\beta}_{\text{PT},j}$ and $\beta_{0,j}$. Of particular significance is that a similar result has been obtained for an elastic net penalty. This can be found in the proof of Proposition 2 within the Supplementary Material of [S11].

Combining expression (S44) with inequality (S42) yields

$$B_n \geq \underline{\lambda}_n \sum_{j=1}^{k_0} \text{sgn}(\bar{\beta}_j)(\hat{\beta}_{\text{PT},j} - \beta_{0,j}) \tag{S45}$$

with probability one for sufficiently large $n$. Note that

$$\sum_{j=1}^{k_0} \text{sgn}(\bar{\beta}_j)(\hat{\beta}_{\text{PT},j} - \beta_{0,j}) \overset{(i)}{\geq} -\|\hat{\beta}_{\text{PT}} - \beta_0\|_{\ell_1}$$

$$\overset{(ii)}{\geq} -\sqrt{p}\|\hat{\beta}_{\text{PT}} - \beta_0\|_{\ell_2}. \tag{S46}$$

where the inequality $(i)$ follows from lower-bounding each of the terms in the summation, and the inequality $(ii)$ follows from the $\ell_1$-$\ell_2$ norm inequality. Combining with the theorem's assumption $\underline{\lambda}_n = O(1/\sqrt{n})$ and strong consistency of $\hat{\beta}_{\text{PT}}$ for $\beta_0$, we can show that $B_n$ is lower-bounded with arbitrarily high probability for sufficiently large $n$ as follows:

$$B_n \geq \frac{-M}{\sqrt{n}} \|\hat{\beta}_{\text{PT}} - \beta_0\|_{\ell_2} \tag{S47}$$

where $M$ is some positive constant. We use the resulting lower bound on $B_n$ in terms of $\|\hat{\beta}_{\text{PT}} - \beta_0\|_{\ell_2}$ to establish root-$n$ consistency in the subsequent step.

### E. Proof of step 5

In order to conclude the proof of Theorem 1, we shall first exploit inequality (S30), arising from the optimality of $\hat{\beta}_{\text{PT}}$. We then exploit the bounds derived in the preceding lines for $B_n$, $W_n^1$, $R_n^1$ and $R_n^0$, and prove $\sqrt{n}\|\hat{\beta}_{\text{PT}} - \beta_0\|_{\ell_2}$ is bounded in probability [S12].

Returning to our earlier inequality (S30), we have

$$\tau_n^2(\mathbf{r}(\hat{\beta}_{\text{PT}})) - \tau_n^2(\mathbf{r}(\beta_0)) + \underline{\lambda}_n\|\hat{\beta}_{\text{PT}}\|_{\ell_1} - \underline{\lambda}_n\|\beta_0\|_{\ell_1} \leq 0. \tag{S48}$$

Recalling that $B_n = \underline{\lambda}_n(\|\hat{\beta}_{\text{PT}}\|_{\ell_1} - \|\beta_0\|_{\ell_1})$ and substituting in the relation $\tau_n^2(\mathbf{r}(\hat{\beta}_{\text{PT}})) = \tau_n^2(\mathbf{r}(\beta_0)) - W_n^0 R_n^0 - R_n^1$ given in equation (S36) into the left-hand side of the above inequality (S48) readily yields

$$B_n - W_n^0 R_n^0 - R_n^1 \leq 0 \tag{S49}$$

Setting $R_n^0 = ((\hat{\beta}_{\text{PT}} - \beta_0)^T/\sqrt{n})E_n^0 - D_n^0$ and $R_n^1 = ((\hat{\beta}_{\text{PT}} - \beta_0)^T/\sqrt{n})E_n^1 - D_n^1$ as given by equation (S37) results in

$$B_n - \frac{(\hat{\beta}_{\text{PT}} - \beta_0)^T}{\sqrt{n}} W_n^0 E_n^0 + W_n^0 D_n^0 - \frac{(\hat{\beta}_{\text{PT}} - \beta_0)^T}{\sqrt{n}} E_n^1 + D_n^1 \leq 0 \tag{S50}$$

We then derive a lower bound for the left-hand side of the preceding inequality, given by equation (S50), by using the earlier bounds in equations (S40), (S41), (S47), and Cauchy-Schwarz (C-S) inequality. To avoid confusion, we will clarify the declarations of the following variables that were introduced earlier: $E_n^0$ and $E_n^1$ are vectors in $\mathbb{R}^p$; $E^0$ and $E^1$ are positive scalars; $D_n^0$ and $D_n^1$ are non-negative scalars; $\overline{D}^0$ and $\overline{D}^1$ are positive scalars. We now provide a summary of the results we have obtained thus far:

$$B_n \geq \frac{-M}{\sqrt{n}}\|\hat{\beta}_{\text{PT}} - \beta_0\|_{\ell_2}, \tag{S51a}$$

$$-\frac{(\hat{\beta}_{\text{PT}} - \beta_0)^T}{\sqrt{n}} E_n^0 \geq -\frac{E^0}{\sqrt{n}}\|\hat{\beta}_{\text{PT}} - \beta_0\|_{\ell_2}, \tag{S51b}$$

$$-\frac{(\hat{\beta}_{\text{PT}} - \beta_0)^T}{\sqrt{n}} E_n^1 \geq -\frac{E^1}{\sqrt{n}}\|\hat{\beta}_{\text{PT}} - \beta_0\|_{\ell_2}, \tag{S51c}$$

$$D_n^0 \geq \overline{D}_n^0\|\hat{\beta}_{\text{PT}} - \beta_0\|_{\ell_2}^2, \tag{S51d}$$

$$D_n^1 \geq \overline{D}_n^1\|\hat{\beta}_{\text{PT}} - \beta_0\|_{\ell_2}^2, \tag{S51e}$$

$$W_n^0 \xrightarrow{a.s.} W^0. \tag{S51f}$$

where the inequalities (S51a), (S51b) and (S51c) hold with high probability for large enough $n$. Combining the above bounds with our inequality (S50) yields

$$0 \geq B_n - \frac{(\hat{\beta}_{\text{PT}} - \beta_0)^T}{\sqrt{n}} W_n^0 E_n^0 + W_n^0 D_n^0$$
$$- \frac{(\hat{\beta}_{\text{PT}} - \beta_0)^T}{\sqrt{n}} E_n^1 + D_n^1$$

$$\geq \overbrace{\frac{-M}{\sqrt{n}}\|\hat{\beta}_{\text{PT}} - \beta_0\|_{\ell_2}}^{\text{By the bound (S51a)}} - \overbrace{W^0 \frac{E^0}{\sqrt{n}}\|\hat{\beta}_{\text{PT}} - \beta_0\|_{\ell_2}}^{\text{By the bounds (S51b) and (S51f)}}$$
$$+ \underbrace{W^0\overline{D}^0\|\hat{\beta}_{\text{PT}} - \beta_0\|_{\ell_2}^2}_{\text{By the bounds (S51d) and (S51f)}}$$
$$+ \underbrace{\overline{D}^1\|\hat{\beta}_{\text{PT}} - \beta_0\|_{\ell_2}^2}_{\text{By the bound (S51e)}} - \underbrace{\frac{E^1}{\sqrt{n}}\|\hat{\beta}_{\text{PT}} - \beta_0\|_{\ell_2}}_{\text{By the bound (S51c)}}$$

$$\overset{(ii)}{=} \frac{\|\hat{\beta}_{\text{PT}} - \beta_0\|_{\ell_2}}{\sqrt{n}}\Big[-M - W^0 E^0 - E^1$$
$$+ (W^0\overline{D}^0 + \overline{D}^1)\sqrt{n}\|\hat{\beta}_{\text{PT}} - \beta_0\|_{\ell_2}\Big] \tag{S52}$$

where the equality $(ii)$ is derived by factoring out $\|\hat{\beta}_{\text{PT}} - \beta_0\|_{\ell_2}/\sqrt{n}$ from the right-hand side of the second inequality. The expression (S52) holds with arbitrarily high probability for sufficiently large $n$. Canceling out a factor of $\|\hat{\beta}_{\text{PT}} - \beta_0\|_{\ell_2}$ from each side of the above inequality, isolating the terms involving $\sqrt{n}\|\hat{\beta}_{\text{PT}} - \beta_0\|_{\ell_2}$, and moving the remaining terms to the other side of the inequality, we find

$$(W^0\overline{D}^0 + \overline{D}^1)\sqrt{n}\|\hat{\beta}_{\text{PT}} - \beta_0\|_{\ell_2} \leq M + W^0 E^0 + E^1. \tag{S53}$$

Then performing some algebra yields

$$\sqrt{n}\|\hat{\beta}_{\text{PT}} - \beta_0\|_{\ell_2} \leq \frac{M + W^0 E^0 + E^1}{W^0\overline{D}^0 + \overline{D}^1} \tag{S54}$$

with arbitrarily high probability for sufficiently large $n$. Recalling that $M$, $\overline{D}^0$, $\overline{D}^1$, $E^0$, and $E^1$ are positive scalars, and $W^0$ is a non-negative scalar, we have that the right-hand side of the above inequality is also a positive scalar. In other

words, the left-hand side of the above inequality is bounded by some positive scalar with an arbitrarily high probability for sufficiently large $n$. Thus, we have $\sqrt{n}(\hat{\boldsymbol{\beta}}_{\text{PT}} - \boldsymbol{\beta}_0) = O_P(1)$; this implies the root-$n$ consistency of the $\tau$-Lasso estimator, $\hat{\boldsymbol{\beta}}_{\text{PT}} - \boldsymbol{\beta}_0 = O_P(1/\sqrt{n})$.

*Remark S4*: Note that we may reuse the symbols $\boldsymbol{\beta}_n^*$, $\bar{\beta}_j$, $\mathbf{v}_n$, $\kappa$, $\underline{\kappa}$, $Z_n^1$, $A_n^0$, $A_n^1$, $W_n^0$, $W^0$, $R_n^0$, $R_n^1$, $E_n^0$, $E^0$, $E_n^1$, $E^1$, $D_n^0$, $\overline{D}_n^0$, $\overline{D}^0$, $D_n^1$, $\overline{D}_n^1$, $\overline{D}^1$, $B_n$ and $M$ for similar or different purposes, meaning they are merely local to this theorem.

## S.V. Proof of Theorem 2

The root-$n$ consistency of $\hat{\boldsymbol{\beta}}_{\text{AT}}$ can be easily established by following the same line of arguments as that of Theorem 1. We only highlight some modifications applied to Theorem 1. First, each occurrence of $\hat{\boldsymbol{\beta}}_{\text{PT}}$ shall be replaced by $\hat{\boldsymbol{\beta}}_{\text{AT}}$. Furthermore, each occurrence of $\lambda_n \|\hat{\boldsymbol{\beta}}_{\text{PT}}\|_{\ell_1}$ and $\lambda_n \|\boldsymbol{\beta}_0\|_{\ell_1}$ shall be replaced by $\lambda_n \sum_{j=1}^{p} |\hat{\beta}_{\text{AT},j}|/|\tilde{\beta}_j|^{\gamma}$ and $\lambda_n \sum_{j=1}^{p} |\beta_{0,j}|/|\tilde{\beta}_j|^{\gamma}$, respectively. To avoid repetition, we only focus on characterizing the modified difference term $B_n = \lambda_n \sum_{j=1}^{p}(|\hat{\beta}_{\text{AT},j}| - |\beta_{0,j}|)/|\tilde{\beta}_j|^{\gamma}$ in the asymptotic regime and show that $B_n$ is lower-bounded by $-M(\|\hat{\boldsymbol{\beta}}_{\text{AT}} - \boldsymbol{\beta}_0\|_{\ell_2}/\sqrt{n})$ with arbitrarily high probability for sufficiently large $n$, similar to the expression (S47).

By Proposition 2, it follows that $\hat{\boldsymbol{\beta}}_{\text{AT}}$ is a strongly consistent estimator of $\boldsymbol{\beta}_0$. By leveraging this property along with the assumption $\beta_{0,j} = 0$ for $j = k_0 + 1, \cdots, p$, we can establish that $k_0$ non-zero coefficients of $\hat{\boldsymbol{\beta}}_{\text{AT}}$ stay away from zero with probability one for sufficiently large $n$ (almost everywhere in the sample space), while simultaneously getting arbitrarily close to non-zero coefficients of $\boldsymbol{\beta}_0$ with probability one for sufficiently large $n$. Building upon this result, we can then apply the Mean Value Theorem to the modified difference term $B_n$, leading to

$$B_n = \lambda_n \sum_{j=1}^{p} \frac{|\hat{\beta}_{\text{AT},j}| - |\beta_{0,j}|}{|\tilde{\beta}_j|^{\gamma}} \geq \lambda_n \sum_{j=1}^{k_0} \text{sgn}(\bar{\beta}_j) \frac{(\hat{\beta}_{\text{AT},j} - \beta_{0,j})}{|\tilde{\beta}_j|^{\gamma}}. \tag{S55}$$

for some $\bar{\beta}_j \in (\beta_{0,j}, \hat{\beta}_{\text{AT},j})$. The above inequality (S55) holds with probability one for sufficiently large $n$ (almost everywhere in the sample space). The term $\text{sgn}(\bar{\beta}_j)$ denotes the derivative of $|\beta_j|$ evaluated at $\bar{\beta}_j$ when $\beta_j \neq 0$. By the theorem's assumption $\lambda_n = O(1/\sqrt{n})$ and strong consistency of $\tilde{\boldsymbol{\beta}}$ for $\boldsymbol{\beta}_0$, we can show that with arbitrarily high probability for sufficiently large $n$

$$B_n \geq \frac{-M}{\sqrt{n}} \|\hat{\boldsymbol{\beta}}_{\text{AT}} - \boldsymbol{\beta}_0\|_{\ell_2} \tag{S56}$$

where $M$ is some positive constant.

The remaining of the proof closely follows that of Theorem 1, and we can conclude $\hat{\boldsymbol{\beta}}_{\text{AT}}$ is a root-$n$ consistent estimator of $\boldsymbol{\beta}_0$, i.e.,

$$\hat{\boldsymbol{\beta}}_{\text{AT}} - \boldsymbol{\beta}_0 = O_P(1/\sqrt{n}). \tag{S57}$$

*Remark S5*: Note that we may reuse the symbols $B_n$, $\bar{\beta}_j$, and $M$ for similar or different purposes, meaning they are merely local to this theorem.

## S.VI. Proof of Theorem 3

To prove the variable selection consistency of the adaptive $\tau$-Lasso estimator, we follow the steps outlined below.

- *Step 1:* We begin by using an intuitive interpretation of the root-$n$ consistency of the adaptive $\tau$-Lasso estimator, which states that $\hat{\boldsymbol{\beta}}_{\text{AT}}$ should converge to $\boldsymbol{\beta}_0 \in \mathbb{R}^p$ at a rate of $1/\sqrt{n}$. By applying this definition, we have that $\|\hat{\boldsymbol{\beta}}_{\text{AT}} - \boldsymbol{\beta}_0\|_{\ell_2} \leq L/\sqrt{n}$ with some constant $L > 0$ for sufficiently large $n$ with arbitrarily high probability. This implies that $\hat{\boldsymbol{\beta}}_{\text{AT}}$ lies within a ball centered around the true value $\boldsymbol{\beta}_0$ of radius $L/\sqrt{n}$, denoted by $\mathbb{B}(\boldsymbol{\beta}_0, L/\sqrt{n}) := \{\boldsymbol{\beta} : \|\hat{\boldsymbol{\beta}}_{\text{AT}} - \boldsymbol{\beta}_0\|_{\ell_2} \leq L/\sqrt{n}\}$ for sufficiently large $n$ with arbitrarily high probability.

- *Step 2:* Next, we let

$$\begin{aligned} \boldsymbol{\beta} &= (\boldsymbol{\beta}_{\mathcal{A}}, \boldsymbol{\beta}_{\mathcal{A}^c}), \quad \text{where } \boldsymbol{\beta} \in \mathbb{R}^p. \\ \boldsymbol{\beta}_{\mathcal{A}} &= [\boldsymbol{\beta}_0]_{\mathcal{A}} + \mathbf{v}_1/\sqrt{n}, \quad \text{where } \mathcal{A} := \{1, 2, \cdots, k_0\}. \\ \boldsymbol{\beta}_{\mathcal{A}^c} &= [\boldsymbol{\beta}_0]_{\mathcal{A}^c} + \mathbf{v}_2/\sqrt{n}, \quad \text{where } \mathcal{A}^c := \{k_0 + 1, \cdots, p\}. \end{aligned} \tag{S58}$$

Here, $\mathbf{v}_1 \in \mathbb{R}^{k_0}$ and $\mathbf{v}_2 \in \mathbb{R}^{p-k_0}$ serve as optimization variables. $\boldsymbol{\beta}_0 = ([\boldsymbol{\beta}_0]_{\mathcal{A}}, [\boldsymbol{\beta}_0]_{\mathcal{A}^c})$ denotes the true regression vector, with $[\boldsymbol{\beta}_0]_{\mathcal{A}} \in \mathbb{R}^{k_0}$ representing the non-zero entries in $\boldsymbol{\beta}_0$ and $[\boldsymbol{\beta}_0]_{\mathcal{A}^c} \in \mathbb{R}^{p-k_0}$ corresponding to the zero entries in $\boldsymbol{\beta}_0$. We then construct a function $U_n(\mathbf{v}_1, \mathbf{v}_2)$ as follows:

$$U_n(\mathbf{v}_1, \mathbf{v}_2) := \mathcal{L}_n(\boldsymbol{\beta}). \tag{S59}$$

Here, $U_n(\mathbf{v}_1, \mathbf{v}_2)$ describes the adaptive $\tau$-Lasso objective function $\mathcal{L}_n(\boldsymbol{\beta})$ given in equation (4) within the main body of the paper. By substituting the relations given in equation (S58) into the objective function, we obtain an objective function $U_n(\mathbf{v}_1, \mathbf{v}_2)$ which is non-smooth and non-convex in terms of $\mathbf{v}_1$ and $\mathbf{v}_2$.

- *Step 3:* To establish the variable selection consistency of $\boldsymbol{\beta}_0$, we will show that

$$\underbrace{U_n(\mathbf{v}_1, \mathbf{0}_{p-k_0})}_{\text{for } \mathbf{v}_2 = \mathbf{0}_{p-k_0}} < \underbrace{U_n(\mathbf{v}_1, \mathbf{v}_2)}_{\text{for } \|\mathbf{v}_2\|_{\ell_2} > 0} \tag{S60}$$

holds over the ball $\mathbb{B}(\mathbf{0}, L) := \{(\mathbf{v}_1, \mathbf{v}_2) : \|\mathbf{v}_1\|_{\ell_2}^2 + \|\mathbf{v}_2\|_{\ell_2}^2 \leq L^2\}$ for sufficiently large $n$ with arbitrarily high probability. In turn, this would suggest that when $U_n(\mathbf{v}_1, \mathbf{v}_2)$ attains its minimum, the probability of $\mathbf{v}_2$ being set to $\mathbf{0}_{p-k_0}$ becomes arbitrarily high for sufficiently large $n$.

Using equation (S58) and the assumption $[\boldsymbol{\beta}_0]_{\mathcal{A}^c} = \mathbf{0}_{p-k_0}$, we can deduce that $\boldsymbol{\beta}_{\mathcal{A}^c} = \mathbf{v}_2/\sqrt{n}$. Combining this result with the outcome of *Step 3*, we will conclude that $[\hat{\boldsymbol{\beta}}_{\text{AT}}]_{\mathcal{A}^c} = \mathbf{0}_{p-k_0}$ for sufficiently large $n$ with arbitrarily high probability.

### A. Proof of step 1

We now proceed with the proof of Theorem 3. From the root-$n$ consistency of the adaptive $\tau$-Lasso estimator and the arguments stated in step 1, it follows that the minimum value of the adaptive $\tau$-Lasso objective function occurs in the ball

$\mathbb{B}(\boldsymbol{\beta}_0, L/\sqrt{n})$ for sufficiently large $n$ with arbitrarily high probability.

## B. Proof of step 2

Next, we rewrite the adaptive $\tau$-Lasso objective function by introducing the function $U_n(\mathbf{v}_1, \mathbf{v}_2)$ expressed in $\mathbf{v}_1$ and $\mathbf{v}_2$. To do so, we replace $\boldsymbol{\beta}_\mathcal{A}$, the first $k_0$ elements of the coefficient vector, with $[\boldsymbol{\beta}_0]_\mathcal{A} + \mathbf{v}_1/\sqrt{n}$ and $\boldsymbol{\beta}_{\mathcal{A}^c}$, the remaining $p - k_0$ elements of the coefficient vector with $\mathbf{v}_2/\sqrt{n}$ as follows:

$$U_n(\mathbf{v}_1, \mathbf{v}_2) = \mathcal{L}_n\left(\overbrace{(\boldsymbol{\beta}_\mathcal{A}, \boldsymbol{\beta}_{\mathcal{A}^c})}^{=\boldsymbol{\beta}}\right)$$

$$= \tau_n^2\left(\mathbf{r}\left(\underbrace{([\boldsymbol{\beta}_0]_\mathcal{A} + \frac{\mathbf{v}_1}{\sqrt{n}}, [\boldsymbol{\beta}_0]_{\mathcal{A}^c} + \frac{\mathbf{v}_2}{\sqrt{n}})}_{=\boldsymbol{\beta}, \text{ by equation (S58)}}\right)\right)$$

$$+ \lambda_n\left(\sum_{j=1}^{k_0} \frac{|\beta_{0,j} + \frac{v_{1,j}}{\sqrt{n}}|}{|\tilde{\beta}_j|^\gamma} + \sum_{j=k_0+1}^{p} \frac{|\beta_{0,j} + \frac{v_{2,j-k_0}}{\sqrt{n}}|}{|\tilde{\beta}_j|^\gamma}\right) \quad \text{(S61)}$$

where $[\boldsymbol{\beta}_0]_{\mathcal{A}^c} = \mathbf{0}_{p-k_0}$ and $\gamma$ is some positive constant. Alternatively, for sufficiently large $n$ with arbitrarily high probability one can estimate $\hat{\boldsymbol{\beta}}_{\text{AT}}$ by minimizing $U_n(\mathbf{v}_1, \mathbf{v}_2)$ over $\mathbf{v}_1$ and $\mathbf{v}_2$ belonging to the ball $\|\mathbf{v}_1\|_{\ell_2}^2 + \|\mathbf{v}_2\|_{\ell_2}^2 \leq L^2$.

Therefore, the variable selection consistency of the adaptive $\tau$-Lasso estimator can be established by showing that

$$\underbrace{U_n(\mathbf{v}_1, \mathbf{0}_{p-k_0})}_{\text{for } \mathbf{v}_2 = \mathbf{0}_{p-k_0}} < \underbrace{U_n(\mathbf{v}_1, \mathbf{v}_2)}_{\text{for } \|\mathbf{v}_2\|_{\ell_2} > 0} \quad \text{(S62)}$$

for any $(\mathbf{v}_1, \mathbf{v}_2)$ within the ball $\mathbb{B}(\mathbf{0}, L)$ for sufficiently large $n$ with arbitrarily high probability. In order to prove the above statement, we start by decomposing $U_n(\mathbf{v}_1, \mathbf{v}_2) - U_n(\mathbf{v}_1, \mathbf{0}_{p-k_0})$ into its smooth and non-smooth terms

$$U_n(\mathbf{v}_1, \mathbf{v}_2) - U_n(\mathbf{v}_1, \mathbf{0}_{p-k_0}) = \tau_n^2\left(\mathbf{r}\left([\boldsymbol{\beta}_0]_\mathcal{A} + \frac{\mathbf{v}_1}{\sqrt{n}}, \frac{\mathbf{v}_2}{\sqrt{n}})\right)\right)$$

$$- \tau_n^2\left(\mathbf{r}\left(([\boldsymbol{\beta}_0]_\mathcal{A} + \frac{\mathbf{v}_1}{\sqrt{n}}, \mathbf{0}_{p-k_0})\right)\right) + \frac{\lambda_n}{\sqrt{n}} \sum_{j=k_0+1}^{p} \frac{|v_{2,j-k_0}|}{|\tilde{\beta}_j|^\gamma}. \quad \text{(S63)}$$

To move forward with the proof, we define $\boldsymbol{\beta}_n^* = \boldsymbol{\beta}_0 + \mathbf{v}_n$, where

$$\mathbf{v}_n = \begin{bmatrix} \frac{\mathbf{v}_1}{\sqrt{n}} \\ \frac{\kappa \mathbf{v}_2}{\sqrt{n}} \end{bmatrix} \quad \text{(S64)}$$

with $0 < \kappa < 1$. This choice of $\boldsymbol{\beta}_n^*$ ensures that it lies within the interval between

$$\begin{bmatrix} [\boldsymbol{\beta}_0]_\mathcal{A} + \frac{\mathbf{v}_1}{\sqrt{n}} \\ \frac{\mathbf{v}_2}{\sqrt{n}} \end{bmatrix} \quad \text{(S65)}$$

and

$$\begin{bmatrix} [\boldsymbol{\beta}_0]_\mathcal{A} + \frac{\mathbf{v}_1}{\sqrt{n}} \\ \mathbf{0}_{p-k_0} \end{bmatrix}. \quad \text{(S66)}$$

We then apply the Mean-Value Theorem to the smooth terms of the equation (S63) as follows:

$$\tau_n^2\left(\mathbf{r}\left(([\boldsymbol{\beta}_0]_\mathcal{A} + \frac{\mathbf{v}_1}{\sqrt{n}}, \frac{\mathbf{v}_2}{\sqrt{n}})\right)\right) - \tau_n^2\left(\mathbf{r}\left(([\boldsymbol{\beta}_0]_\mathcal{A} + \frac{\mathbf{v}_1}{\sqrt{n}}, \mathbf{0}_{p-k_0})\right)\right)$$

$$= [\mathbf{0}_{k_0}^T, \frac{\mathbf{v}_2^T}{\sqrt{n}}] \times \overbrace{\frac{1}{n}\sum_{i=1}^{n} \psi_0\left(\frac{u_i - \mathbf{x}_{[i]}^T \mathbf{v}_n}{s_n(\mathbf{r}(\boldsymbol{\beta}_n^*))}\right) \mathbf{x}_{[i]}}^{Z_n^1}$$

$$\times \left[-2 \times \overbrace{\frac{1}{n}\sum_{i=1}^{n} \rho_1\left(\frac{u_i - \mathbf{x}_{[i]}^T \mathbf{v}_n}{s_n(\mathbf{r}(\boldsymbol{\beta}_n^*))}\right)}^{A_n^1}\right.$$

$$\times \underbrace{\frac{s_n(\mathbf{r}(\boldsymbol{\beta}_n^*))}{\frac{1}{n}\sum_{i=1}^{n} \psi_0\left(\frac{u_i - \mathbf{x}_{[i]}^T \mathbf{v}_n}{s_n(\mathbf{r}(\boldsymbol{\beta}_n^*))}\right) \frac{u_i - \mathbf{x}_{[i]}^T \mathbf{v}_n}{s_n(\mathbf{r}(\boldsymbol{\beta}_n^*))}}}_{B_n^0}$$

$$+ \left(\frac{1}{n}\sum_{i=1}^{n} \psi_1\left(\frac{u_i - \mathbf{x}_{[i]}^T \mathbf{v}_n}{s_n(\mathbf{r}(\boldsymbol{\beta}_n^*))}\right) \frac{u_i - \mathbf{x}_{[i]}^T \mathbf{v}_n}{s_n(\mathbf{r}(\boldsymbol{\beta}_n^*))}\right)$$

$$\times \underbrace{\frac{s_n(\mathbf{r}(\boldsymbol{\beta}_n^*))}{\left(\frac{1}{n}\sum_{i=1}^{n} \psi_0\left(\frac{u_i - \mathbf{x}_{[i]}^T \mathbf{v}_n}{s_n(\mathbf{r}(\boldsymbol{\beta}_n^*))}\right) \frac{u_i - \mathbf{x}_{[i]}^T \mathbf{v}_n}{s_n(\mathbf{r}(\boldsymbol{\beta}_n^*))}\right)}}_{B_n^0} \right]$$

$$- [\mathbf{0}_{k_0}^T, \frac{\mathbf{v}_2^T}{\sqrt{n}}] \times \frac{1}{n}\sum_{i=1}^{n} \psi_1\left(\frac{u_i - \mathbf{x}_{[i]}^T \mathbf{v}_n}{s_n(\mathbf{r}(\boldsymbol{\beta}_n^*))}\right) \mathbf{x}_{[i]} \times s_n(\mathbf{r}(\boldsymbol{\beta}_n^*)). \quad \text{(S67)}$$

This allows us to examine the asymptotic behavior of the smooth component of $U_n(\mathbf{v}_1, \mathbf{v}_2) - U_n(\mathbf{v}_1, \mathbf{0}_{p-k_0})$ over the ball $\mathbb{B}(\mathbf{0}, L)$. Through this analysis, we gain valuable insights into the asymptotic behavior of $U_n(\mathbf{v}_1, \mathbf{v}_2) - U_n(\mathbf{v}_1, \mathbf{0}_{p-k_0})$ over that region, which, in turn, can be used to establish the variable selection consistency of the adaptive $\tau$-Lasso estimator. To simplify notation, we will substitute the overlined terms with $Z_n^1$ and $A_n^1$ and the underlined terms with $B_n^0$ in the above expression, resulting in

$$-\frac{1}{\sqrt{n}}\left(\overbrace{[\mathbf{0}_{k_0}^T, \mathbf{v}_2^T] \times \frac{1}{n}\sum_{i=1}^{n} \psi_0\left(\frac{u_i - \mathbf{x}_{[i]}^T \mathbf{v}_n}{s_n(\mathbf{r}(\boldsymbol{\beta}_n^*))}\right) \mathbf{x}_{[i]}}^{C_n^0}\right.$$

$$\times \left[(2Z_n^1 - A_n^1)B_n^0\right]$$

$$+ \underbrace{[\mathbf{0}_{k_0}^T, \mathbf{v}_2^T] \times \frac{1}{n}\sum_{i=1}^{n} \psi_1\left(\frac{u_i - \mathbf{x}_{[i]}^T \mathbf{v}_n}{s_n(\mathbf{r}(\boldsymbol{\beta}_n^*))}\right) \mathbf{x}_{[i]} \times s_n(\mathbf{r}(\boldsymbol{\beta}_n^*))}_{C_n^1}\right). \quad \text{(S68)}$$

To streamline the notation further, we will define $C_n^0$, the overlined term, and $C_n^1$, the underlined term, as illustrated in the above expression.

## C. Proof of step 3

We shall now exploit the results derived in Lemma 4 and Theorem 4 of [S8] and show that $U_n(\mathbf{v}_1, \mathbf{v}_2) - U_n(\mathbf{v}_1, \mathbf{0}_{p-k_0})$ is uniformly bounded by 0 from below in probability over the ball

$\mathbb{B}(0, L)$ with $\|\mathbf{v}_2\|_{\ell_2} > 0$. Alternatively, for sufficiently large $n$ with arbitrarily high probability $U_n(\mathbf{v}_1, \mathbf{v}_2) - U_n(\mathbf{v}_1, \mathbf{0}_{p-k_0})$ remains strictly positive.

To do so, we begin by analyzing the asymptotic behavior of all the components $B_n^0$, $2Z_n^1 - A_n^1$, $s_n(\mathbf{r}(\boldsymbol{\beta}_n^*))$, $C_n^0$, and $C_n^1$, respectively.

- *Asymptotic behavior of $B_n^0$:* From Lemma 4 of [S8], we know that the term $B_n^0$ is uniformly bounded in probability over the ball $\mathbb{B}(\mathbf{0}, L) := \{(\mathbf{v}_1, \mathbf{v}_2) : \|\mathbf{v}_1\|_{\ell_2}^2 + \|\mathbf{v}_2\|_{\ell_2}^2 \le L^2\}$. By the definition of boundedness in probability [S12], we can state that for sufficiently large values of $n$, $|B_n^0|$ is upper-bounded by some $B > 0$ with arbitrarily high probability over the entire ball.

- *Asymptotic behavior of $2Z_n^1 - A_n^1$:* By the results derived in Lemma 4 of [S8], the term $|2Z_n^1 - A_n^1|$ is upper-bounded by some $D = 2Z^1 - A^1 + \epsilon$ for every $\epsilon > 0$ with arbitrarily high probability for sufficiently large $n$ where $A^1$ and $Z^1$ are defined as follows:

$$A^1 = \mathbb{E}_F\left[\psi_1\left(\frac{u}{s(\boldsymbol{\beta}_0)}\right)\frac{u}{s(\boldsymbol{\beta}_0)}\right] \quad \text{and}$$
$$Z^1 = \mathbb{E}_F\left[\rho_1\left(\frac{u}{s(\boldsymbol{\beta}_0)}\right)\right]. \tag{S69}$$

*Remark S6*: Note that $2Z^1 - A^1$ and $2Z_n^1 - A_n^1$ are nonnegative by assumption 2. $2Z^1 - A^1$ and $2Z_n^1 - A_n^1$ are closely related to the numerators of $\overline{W}$ and $\overline{W}_n$ that appeared in equations (S10) and (8) within the main body of the paper, respectively.

- *Asymptotic behavior of $s_n(\mathbf{r}(\boldsymbol{\beta}_n^*))$:* Furthermore, it follows from Lemma 4 of [S8] that for every $\epsilon > 0$ and sufficiently large $n$

$$s_n(\mathbf{r}(\boldsymbol{\beta}_n^*)) < \overbrace{s(\boldsymbol{\beta}_0)}^{>0} + \epsilon. \tag{S70}$$

In other words, for sufficiently large $n$

$$s_n(\mathbf{r}(\boldsymbol{\beta}_n^*)) < S \tag{S71}$$

holds for some $S > 0$ or $s_n(\mathbf{r}(\boldsymbol{\beta}_n^*))$ is upper-bounded by some $S > 0$ for large enough $n$.

- *Asymptotic behavior of $C_n^0$ and $C_n^1$:* By the same argument given in the proof of Theorem 4 of [S8], we have that the overlined term $C_n^0$ and the underlined term $C_n^1$, as given by equation (S68), are upper bounded as follows:

$$|C_n^0| < \|\mathbf{v}_2\|_{\ell_2} C^0 \quad \text{and} \quad |C_n^1| < \|\mathbf{v}_2\|_{\ell_2} C^1 \tag{S72}$$

for some positive constants $C^0$ and $C1$ with arbitrarily high probability for sufficiently large $n$.

Writing down the results we have obtained so far, the following expressions hold with arbitrarily high probability for sufficiently large $n$.

$$|2Z_n^1 - A_n^1| < \overbrace{2Z^1 - A^1 + \epsilon}^{D}, \tag{S73a}$$
$$|B_n^0| < B, \tag{S73b}$$
$$s_n(\mathbf{r}(\boldsymbol{\beta}_n^*)) < S, \tag{S73c}$$
$$\overline{C_n^0} \le |C_n^0|$$
$$< \|\mathbf{v}_2\|_{\ell_2} C^0 \quad \text{and} \tag{S73d}$$
$$\underline{C_n^1} \le |C_n^1|$$
$$< \|\mathbf{v}_2\|_{\ell_2} C^1. \tag{S73e}$$

All terms on the right-hand side of the inequality are positive, as justified in the preceding statements. In conjunction with equation (S68), we obtain

$$\overline{C_n^0}\left[(2Z_n^1 - A_n^1)B_n^0\right] + \underline{C_n^1} s_n(\mathbf{r}(\boldsymbol{\beta}_n^*))$$
$$< (C^0 D B \|\mathbf{v}_2\|_{\ell_2} + C^1 S \|\mathbf{v}_2\|_{\ell_2}) \tag{S74}$$

with arbitrarily high probability for sufficiently large $n$, implying that

$$-\frac{1}{\sqrt{n}}\left(\overline{C_n^0}\left[(2Z_n^1 - A_n^1)B_n^0\right] + \underline{C_n^1} s_n(\mathbf{r}(\boldsymbol{\beta}_n^*))\right)$$
$$> -\frac{\|\mathbf{v}_2\|_{\ell_2}}{\sqrt{n}}(C^0 D B + C^1 S) \tag{S75}$$

with arbitrarily high probability for sufficiently large $n$. Hence, we have the lower bound

$$\overbrace{\tau_n^2\left(\mathbf{r}\left(([\boldsymbol{\beta}_0]_{\mathcal{A}} + \frac{\mathbf{v}_1}{\sqrt{n}}, \frac{\mathbf{v}_2}{\sqrt{n}})\right)\right) - \tau_n^2\left(\mathbf{r}\left(([\boldsymbol{\beta}_0]_{\mathcal{A}} + \frac{\mathbf{v}_1}{\sqrt{n}}, \mathbf{0}_{p-k_0})\right)\right)}^{\text{the smooth term of } U_n(\mathbf{v}_1,\mathbf{v}_2)-U_n(\mathbf{v}_1,\mathbf{0}_{p-k_0})}$$
$$> -\frac{\|\mathbf{v}_2\|_{\ell_2}}{\sqrt{n}}\underbrace{\left(DBC^0 + SC^1\right)}_{M_1 > 0} \tag{S76}$$

for the smooth term of $U_n(\mathbf{v}_1, \mathbf{v}_2) - U_n(\mathbf{v}_1, \mathbf{0}_{p-k_0})$, which holds with arbitrarily high probability for sufficiently large $n$. We now focus on the asymptotic behavior of the penalty term $(\lambda_n/\sqrt{n})\sum_{j=k_0+1}^p |v_{2,j-k_0}|/|\tilde{\beta}_j|^\gamma$. By the results provided in the proof of Theorem 4 of [S8], we have

$$\frac{\lambda_n}{\sqrt{n}}\sum_{j=k_0+1}^p \frac{|v_{2,j-k_0}|}{|\tilde{\beta}_j|^\gamma} > \frac{\|\mathbf{v}_2\|_{\ell_2}}{\sqrt{n}} \times \frac{\lambda_n n^{\gamma/2}}{M_2^\gamma} \tag{S77}$$

with arbitrarily high probability for sufficiently large $n$ for some $M_2 > 0$.

Recalling step 3, we achieve the variable selection consistency by showing that the difference term $U_n(\mathbf{v}_1, \mathbf{v}_2) - U_n(\mathbf{v}_1, \mathbf{0}_{p-k_0})$ is strictly positive for sufficiently large $n$ with arbitrarily high probability. Thus, the penalty term $(\lambda_n/\sqrt{n})\sum_{j=k_0+1}^p |v_{2,j-k_0}|/|\tilde{\beta}_j|^\gamma$ is required to grow large enough to ensure that the difference term mentioned above remains strictly positive. Putting together all of the above, we get

$$U_n(\mathbf{v}_1, \mathbf{v}_2) - U_n(\mathbf{v}_1, \mathbf{0}_{p-k_0}) > \frac{\|\mathbf{v}_2\|_{\ell_2}}{\sqrt{n}}\left(-M_1 + \frac{\lambda_n n^{\gamma/2}}{M_2^\gamma}\right). \tag{S78}$$

Therefore, the above difference term is uniformly lower-bounded over the ball $\{(\mathbf{v}_1, \mathbf{v}_2) : \|\mathbf{v}_1\|_{\ell_2}^2 + \|\mathbf{v}_2\|_{\ell_2}^2 \leq L^2\}$ with arbitrarily high probability for sufficiently large $n$. Moreover, by the assumption of the theorem $\lambda_n n^{\gamma/2} \to \infty$ as $n \to \infty$, the right-hand side of equation (S78) becomes strictly positive, which implies that the adaptive $\tau$-Lasso estimator is variable selection consistent as claimed.

*Remark S7:* Note that we may reuse the symbols $\beta_n^*$, $\mathbf{v}_n$, $\kappa$, $Z_n^1$, $Z^1$, $A_n^1$, $A^1$, $B_n^0$, $B$, $C_n^0$, $C^0$, $C_n^1$, $C^1$, $D$, $L$, $S$, $M_1$, and $M_2$ for similar or different purposes, meaning they are merely local to this theorem.

## S.VII. Proof of Theorem 4

To begin, we sketch out the proof establishing the asymptotic normality of the adaptive $\tau$-Lasso for the true non-zero coefficients of the parameter vector $\boldsymbol{\beta}_0$. We then provide a detailed proof. The key steps are as follows:

- *Step 1:* In order to establish the asymptotic normality of the adaptive $\tau$-Lasso estimator for the regression coefficients corresponding to truly active predictors, we shall first take the (partial) generalized gradient of the objective function $\mathcal{L}_n(\boldsymbol{\beta})$ with respect to $\boldsymbol{\beta}_{\mathcal{A}}$ and evaluate it at its minimum point $\hat{\boldsymbol{\beta}}_{\mathrm{AT}}$. Combining this with Proposition 2 and Theorem 3, we will conclude that for sufficiently large $n$ with arbitrarily high probability, the generalized gradient $\partial_{\boldsymbol{\beta}_{\mathcal{A}}} \mathcal{L}_n(\hat{\boldsymbol{\beta}}_{\mathrm{AT}})$ exists and contains $\mathbf{0}_{k_0}$, i.e., $\mathbf{0}_{k_0} \in \partial_{\boldsymbol{\beta}_{\mathcal{A}}} \mathcal{L}_n(\hat{\boldsymbol{\beta}}_{\mathrm{AT}})$. Recalling the smoothness of the penalty term for the non-zero entries of $\boldsymbol{\beta}$, the zero-generalized gradient condition $\mathbf{0}_{k_0} \in \partial_{\boldsymbol{\beta}_{\mathcal{A}}} \mathcal{L}_n(\hat{\boldsymbol{\beta}}_{\mathrm{AT}})$ simplifies to a zero-gradient condition, $\nabla_{\boldsymbol{\beta}_{\mathcal{A}}} \mathcal{L}_n(\hat{\boldsymbol{\beta}}_{\mathrm{AT}}) = \mathbf{0}_{k_0}$.
- *Step 2:* For the second part of the proof, we shall simplify the gradient term $\nabla_{\boldsymbol{\beta}_{\mathcal{A}}} \mathcal{L}_n(\hat{\boldsymbol{\beta}}_{\mathrm{AT}})$ by algebraic manipulations, apply the Mean-Value Theorem to the smooth terms of the resulting equation, and factor out the common term $\sqrt{n}([\hat{\boldsymbol{\beta}}_{\mathrm{AT}}]_{\mathcal{A}} - [\boldsymbol{\beta}_0]_{\mathcal{A}})$. We will subsequently solve the obtained expression for $\sqrt{n}([\hat{\boldsymbol{\beta}}_{\mathrm{AT}}]_{\mathcal{A}} - [\boldsymbol{\beta}_0]_{\mathcal{A}})$
- *Step 3:* By the Continuous Mapping Theorem [S13], the Strong Law of Large Numbers, the Central Limit Theorem and the results derived in the preceding sections of this paper, we will show that $\sqrt{n}([\hat{\boldsymbol{\beta}}_{\mathrm{AT}}]_{\mathcal{A}} - [\boldsymbol{\beta}_0]_{\mathcal{A}})$ converges in distribution to a multivariate Gaussian. Hence, the asymptotic normality of the adaptive $\tau$-Lasso estimator is established, which completes our proof.

The detailed proof proceeds as follows:

### A. Proof of step 1

To begin, we have that $\hat{\boldsymbol{\beta}}_{\mathrm{AT}}$ is a minimum of the adaptive $\tau$-Lasso objective function $\mathcal{L}_n(\boldsymbol{\beta})$, as verified by the definition of the adaptive $\tau$-Lasso estimator given in equation (4) within the main body of the paper. Combining the minimality condition with the local Lipschitzity of $\mathcal{L}_n(\boldsymbol{\beta})$ and Proposition 2.3.2 of [S5], we conclude that $\mathbf{0}_p$ belongs to the generalized gradient of the objective function $\mathcal{L}_n(\boldsymbol{\beta})$ with respect to $\boldsymbol{\beta}$ evaluated at $\boldsymbol{\beta} = \hat{\boldsymbol{\beta}}_{\mathrm{AT}}$, i.e., $\mathbf{0}_p \in \partial_{\boldsymbol{\beta}} \mathcal{L}_n(\hat{\boldsymbol{\beta}}_{\mathrm{AT}})$. On the other hand, $\hat{\boldsymbol{\beta}}_{\mathrm{AT}} \xrightarrow{a.s.}$ $\boldsymbol{\beta}_0$ implies the estimated coefficients associated with the truly active predictors are bounded away from zero for sufficiently large $n$ with probability one. By Theorem 3, $[\hat{\boldsymbol{\beta}}_{\mathrm{AT}}]_{\mathcal{A}^c} = \mathbf{0}_{p-k_0}$ (a column vector of $p - k_0$ zeros) for sufficiently large $n$ with arbitrarily high probability. Putting together all of the above, with arbitrarily high probability for large enough $n$ the partial generalized derivative of the objective function with respect to $\boldsymbol{\beta}_{\mathcal{A}}$ at $\hat{\boldsymbol{\beta}}_{\mathrm{AT}}$ exists and contains $\mathbf{0}_{k_0}$ (a column vector of $k_0$ zeros), or mathematically speaking $\mathbf{0}_{k_0} \in \partial_{\boldsymbol{\beta}_{\mathcal{A}}} \mathcal{L}_n(\boldsymbol{\beta})|_{\boldsymbol{\beta} = \hat{\boldsymbol{\beta}}_{\mathrm{AT}}}$. Given that the penalty term is smooth for the non-zero entries of $\boldsymbol{\beta}$, the zero-generalized gradient condition reduces to a zero-gradient condition, specifically $\nabla_{\boldsymbol{\beta}_{\mathcal{A}}} \mathcal{L}_n(\hat{\boldsymbol{\beta}}_{\mathrm{AT}}) = \mathbf{0}_{k_0}$. In simpler terms, this means that the gradient of the objective function $\mathcal{L}_n(\hat{\boldsymbol{\beta}}_{\mathrm{AT}})$ with respect to $\boldsymbol{\beta}_{\mathcal{A}}$ at its minimum point $\hat{\boldsymbol{\beta}}_{\mathrm{AT}}$ is equal to $\mathbf{0}_{k_0}$ as follows:

$$\nabla_{\boldsymbol{\beta}_{\mathcal{A}}} \mathcal{L}_n(\hat{\boldsymbol{\beta}}_{\mathrm{AT}}) = \mathbf{0}_{k_0}. \tag{S79}$$

By expanding out the gradient term and rearranging the resulting equation, we have

$$\mathbf{0}_{k_0} = \Bigg[\Bigg( -2 \underbrace{\frac{\frac{1}{n}\sum_{i=1}^n \rho_1\Big(\frac{y_i - \mathbf{x}_{[i]}^T \hat{\boldsymbol{\beta}}_{\mathrm{AT}}}{s_n(\mathbf{r}(\hat{\boldsymbol{\beta}}_{\mathrm{AT}}))}\Big)}{\frac{1}{n}\sum_{i=1}^n \psi_0\Big(\frac{y_i - \mathbf{x}_{[i]}^T \hat{\boldsymbol{\beta}}_{\mathrm{AT}}}{s_n(\mathbf{r}(\hat{\boldsymbol{\beta}}_{\mathrm{AT}}))}\Big)\frac{y_i - \mathbf{x}_{[i]}^T \hat{\boldsymbol{\beta}}_{\mathrm{AT}}}{s_n(\mathbf{r}(\hat{\boldsymbol{\beta}}_{\mathrm{AT}}))}}}_{A_n^0}$$
$$\times \frac{s_n(\mathbf{r}(\hat{\boldsymbol{\beta}}_{\mathrm{AT}}))}{n}\sum_{i=1}^n \psi_0\Big(\frac{y_i - \mathbf{x}_{[i]}^T \hat{\boldsymbol{\beta}}_{\mathrm{AT}}}{s_n(\mathbf{r}(\hat{\boldsymbol{\beta}}_{\mathrm{AT}}))}\Big)\mathbf{x}_{[i],\mathcal{A}}\Bigg)$$
$$+ \underbrace{\Bigg(\frac{-s_n(\mathbf{r}(\hat{\boldsymbol{\beta}}_{\mathrm{AT}}))}{n}\sum_{i=1}^n \psi_1\Big(\frac{y_i - \mathbf{x}_{[i]}^T \hat{\boldsymbol{\beta}}_{\mathrm{AT}}}{s_n(\mathbf{r}(\hat{\boldsymbol{\beta}}_{\mathrm{AT}}))}\Big)\mathbf{x}_{[i],\mathcal{A}}\Bigg)}_{-A_n^1}$$
$$+ \Bigg(\underbrace{\frac{\Big(\frac{-1}{n}\sum_{i=1}^n \psi_1\Big(\frac{y_i - \mathbf{x}_{[i]}^T \hat{\boldsymbol{\beta}}_{\mathrm{AT}}}{s_n(\mathbf{r}(\hat{\boldsymbol{\beta}}_{\mathrm{AT}}))}\Big)\frac{y_i - \mathbf{x}_{[i]}^T \hat{\boldsymbol{\beta}}_{\mathrm{AT}}}{s_n(\mathbf{r}(\hat{\boldsymbol{\beta}}_{\mathrm{AT}}))}\Big)}{\Big(\frac{1}{n}\sum_{i=1}^n \psi_0\Big(\frac{y_i - \mathbf{x}_{[i]}^T \hat{\boldsymbol{\beta}}_{\mathrm{AT}}}{s_n(\mathbf{r}(\hat{\boldsymbol{\beta}}_{\mathrm{AT}}))}\Big)\frac{y_i - \mathbf{x}_{[i]}^T \hat{\boldsymbol{\beta}}_{\mathrm{AT}}}{s_n(\mathbf{r}(\hat{\boldsymbol{\beta}}_{\mathrm{AT}}))}\Big)}}_{A_n^0}$$
$$\times \frac{-s_n(\mathbf{r}(\hat{\boldsymbol{\beta}}_{\mathrm{AT}}))}{n}\sum_{i=1}^n \psi_0\Big(\frac{y_i - \mathbf{x}_{[i]}^T \hat{\boldsymbol{\beta}}_{\mathrm{AT}}}{s_n(\mathbf{r}(\hat{\boldsymbol{\beta}}_{\mathrm{AT}}))}\Big)\mathbf{x}_{[i],\mathcal{A}}\Bigg)\Bigg]$$
$$+ \lambda_n \nabla_{\boldsymbol{\beta}_{\mathcal{A}}}\Big(\sum_{j=1}^p \frac{|\beta_j|}{|\tilde{\beta}_j|^\gamma}\Big)\Bigg|_{\boldsymbol{\beta} = \hat{\boldsymbol{\beta}}_{\mathrm{AT}}}. \tag{S80}$$

To simplify the notation, we will use the shorthand $A_n^0$ for the underlined terms and $Z_n^1$ and $A_n^1$ for the overlined terms, respectively, as follows:

$$\mathbf{0}_{k_0} = \Bigg[\Bigg(\underbrace{\bigg(\frac{2Z_n^1 - A_n^1}{A_n^0}\bigg)}_{W_n^0}$$

$$\times \bigg(-s_n(\mathbf{r}(\hat{\boldsymbol{\beta}}_{\text{AT}}))\frac{1}{n}\sum_{i=1}^{n}\psi_0\bigg(\frac{y_i - \mathbf{x}_{[i]}^T\hat{\boldsymbol{\beta}}_{\text{AT}}}{s_n(\mathbf{r}(\hat{\boldsymbol{\beta}}_{\text{AT}}))}\bigg)\mathbf{x}_{[i],\mathcal{A}}\bigg)\Bigg)$$

$$+ \bigg(-s_n(\mathbf{r}(\hat{\boldsymbol{\beta}}_{\text{AT}}))\frac{1}{n}\sum_{i=1}^{n}\psi_1\bigg(\frac{y_i - \mathbf{x}_{[i]}^T\hat{\boldsymbol{\beta}}_{\text{AT}}}{s_n(\mathbf{r}(\hat{\boldsymbol{\beta}}_{\text{AT}}))}\bigg)\mathbf{x}_{[i],\mathcal{A}}\bigg)\Bigg]$$

$$+\lambda_n \nabla_{\boldsymbol{\beta}_{\mathcal{A}}}\bigg(\sum_{j=1}^{p}\frac{|\beta_j|}{|\tilde{\beta}_j|^\gamma}\bigg)\bigg|_{\boldsymbol{\beta}=\hat{\boldsymbol{\beta}}_{\text{AT}}}. \quad (S81)$$

and in what follows the term $(2Z_n^1 - A_n^1)/A_n^0$ is replaced by $W_n^0$ for the sake of proof clarity.

### B. Proof of step 2

Turning to step 2 of the proof, we carry on by further simplification of the gradient expression $\nabla_{\boldsymbol{\beta}_{\mathcal{A}}}\mathcal{L}_n(\hat{\boldsymbol{\beta}}_{\text{AT}})$ by decomposing $\psi_0\bigg(\frac{y_i - \mathbf{x}_{[i]}^T\hat{\boldsymbol{\beta}}_{\text{AT}}}{s_n(\mathbf{r}(\hat{\boldsymbol{\beta}}_{\text{AT}}))}\bigg)$ and $\psi_1\bigg(\frac{y_i - \mathbf{x}_{[i]}^T\hat{\boldsymbol{\beta}}_{\text{AT}}}{s_n(\mathbf{r}(\hat{\boldsymbol{\beta}}_{\text{AT}}))}\bigg)$ into two components, each as follows:

$$\psi_0\bigg(\frac{y_i - \mathbf{x}_{[i]}^T\hat{\boldsymbol{\beta}}_{\text{AT}}}{s_n(\mathbf{r}(\hat{\boldsymbol{\beta}}_{\text{AT}}))}\bigg) = \psi_0\bigg(\frac{y_i - \mathbf{x}_{[i],\mathcal{A}}^T[\hat{\boldsymbol{\beta}}_{\text{AT}}]_{\mathcal{A}}}{s_n(\mathbf{r}(\hat{\boldsymbol{\beta}}_{\text{AT}}))}\bigg) + o_i^0,$$

$$\psi_1\bigg(\frac{y_i - \mathbf{x}_{[i]}^T\hat{\boldsymbol{\beta}}_{\text{AT}}}{s_n(\mathbf{r}(\hat{\boldsymbol{\beta}}_{\text{AT}}))}\bigg) = \psi_1\bigg(\frac{y_i - \mathbf{x}_{[i],\mathcal{A}}^T[\hat{\boldsymbol{\beta}}_{\text{AT}}]_{\mathcal{A}}}{s_n(\mathbf{r}(\hat{\boldsymbol{\beta}}_{\text{AT}}))}\bigg) + o_i^1. \quad (S82)$$

where we introduce the notations $o_i^0$ and $o_i^1$ to account for the approximation errors arising from the use of $\mathbf{x}_{[i],\mathcal{A}}^T[\hat{\boldsymbol{\beta}}_{\text{AT}}]_{\mathcal{A}}$ instead of $\mathbf{x}_{[i]}^T\hat{\boldsymbol{\beta}}_{\text{AT}}$.

Moving forward in our analysis, we shift our focus to the asymptotic behavior of $o_i^0$ and $o_i^1$. Using Theorem 3 and continuity of $\psi_0(.)$ and $\psi_1(.)$, we can conclude that $\mathbb{P}(o_i^0 = 0) \to 1$ and $\mathbb{P}(o_i^1 = 0) \to 1$ for some $o_i^0$ and $o_i^1$ as $n$ grows to infinity. We can then rewrite the zero-gradient equation (S81) as follows:

$$\Bigg[-W_n^0 s_n(\mathbf{r}(\hat{\boldsymbol{\beta}}_{\text{AT}}))\frac{1}{\sqrt{n}}\sum_{i=1}^{n}\psi_0\bigg(\frac{y_i - \mathbf{x}_{[i],\mathcal{A}}^T[\hat{\boldsymbol{\beta}}_{\text{AT}}]_{\mathcal{A}}}{s_n(\mathbf{r}(\hat{\boldsymbol{\beta}}_{\text{AT}}))}\bigg)\mathbf{x}_{[i],\mathcal{A}}$$

$$- s_n(\mathbf{r}(\hat{\boldsymbol{\beta}}_{\text{AT}}))\frac{1}{\sqrt{n}}\sum_{i=1}^{n}\psi_1\bigg(\frac{y_i - \mathbf{x}_{[i],\mathcal{A}}^T[\hat{\boldsymbol{\beta}}_{\text{AT}}]_{\mathcal{A}}}{s_n(\mathbf{r}(\hat{\boldsymbol{\beta}}_{\text{AT}}))}\bigg)\mathbf{x}_{[i],\mathcal{A}}$$

$$- s_n(\mathbf{r}(\hat{\boldsymbol{\beta}}_{\text{AT}}))W_n^0\underbrace{\frac{1}{\sqrt{n}}\sum_{i=1}^{n}o_i^0\mathbf{x}_{[i],\mathcal{A}}}_{R_n^0}$$

$$- s_n(\mathbf{r}(\hat{\boldsymbol{\beta}}_{\text{AT}}))\underbrace{\frac{1}{\sqrt{n}}\sum_{i=1}^{n}o_i^1\mathbf{x}_{[i],\mathcal{A}}}_{R_n^1}\Bigg]$$

$$+\sqrt{n}\lambda_n \nabla_{\boldsymbol{\beta}_{\mathcal{A}}}\bigg(\sum_{j=1}^{p}\frac{|\beta_j|}{|\tilde{\beta}_j|^\gamma}\bigg)\bigg|_{\boldsymbol{\beta}=\hat{\boldsymbol{\beta}}_{\text{AT}}} = \mathbf{0}_{k_0} \quad (S83)$$

where the underlined terms in the above equation possess the asymptotic properties of $\mathbb{P}(R_n^0 = \mathbf{0}_{k_0}) \to 1$ and $\mathbb{P}(R_n^1 = \mathbf{0}_{k_0}) \to 1$ as $n \to \infty$. Applying the mean-value theorem to the smooth terms of the above zero-gradient equation associated with the adaptive $\tau$-Lasso objective function evaluated at its minimum $\boldsymbol{\beta} = \hat{\boldsymbol{\beta}}_{\text{AT}}$ yields

$$\Bigg[-s_n(\mathbf{r}(\hat{\boldsymbol{\beta}}_{\text{AT}}))W_n^0\frac{1}{\sqrt{n}}\sum_{i=1}^{n}\psi_0\bigg(\frac{y_i - \mathbf{x}_{[i],\mathcal{A}}^T[\boldsymbol{\beta}_0]_{\mathcal{A}}}{s_n(\mathbf{r}(\hat{\boldsymbol{\beta}}_{\text{AT}}))}\bigg)\mathbf{x}_{[i],\mathcal{A}}$$

$$- s_n(\mathbf{r}(\hat{\boldsymbol{\beta}}_{\text{AT}}))\frac{1}{\sqrt{n}}\sum_{i=1}^{n}\psi_1\bigg(\frac{y_i - \mathbf{x}_{[i],\mathcal{A}}^T[\boldsymbol{\beta}_0]_{\mathcal{A}}}{s_n(\mathbf{r}(\hat{\boldsymbol{\beta}}_{\text{AT}}))}\bigg)\mathbf{x}_{[i],\mathcal{A}}$$

$$- s_n(\mathbf{r}(\hat{\boldsymbol{\beta}}_{\text{AT}}))W_n^0 R_n^0 - s_n(\mathbf{r}(\hat{\boldsymbol{\beta}}_{\text{AT}}))R_n^1\Bigg]$$

$$+ \bigg(s_n(\mathbf{r}(\hat{\boldsymbol{\beta}}_{\text{AT}}))W_n^0 \times \frac{1}{s_n(\mathbf{r}(\hat{\boldsymbol{\beta}}_{\text{AT}}))}$$

$$\times \bigg[\frac{1}{\sqrt{n}}\sum_{i=1}^{n}\psi_0'\bigg(\frac{u_i - \zeta_i\mathbf{x}_{[i],\mathcal{A}}^T([\hat{\boldsymbol{\beta}}_{\text{AT}}]_{\mathcal{A}} - [\boldsymbol{\beta}_0]_{\mathcal{A}})}{s_n(\mathbf{r}(\hat{\boldsymbol{\beta}}_{\text{AT}}))}\bigg)\mathbf{x}_{[i],\mathcal{A}}\mathbf{x}_{[i],\mathcal{A}}^T\bigg]$$

$$\times ([\hat{\boldsymbol{\beta}}_{\text{AT}}]_{\mathcal{A}} - [\boldsymbol{\beta}_0]_{\mathcal{A}})\bigg)$$

$$+ \bigg(s_n(\mathbf{r}(\hat{\boldsymbol{\beta}}_{\text{AT}})) \times \frac{1}{s_n(\mathbf{r}(\hat{\boldsymbol{\beta}}_{\text{AT}}))}$$

$$\times \bigg[\frac{1}{\sqrt{n}}\sum_{i=1}^{n}\psi_1'\bigg(\frac{u_i - \zeta_i\mathbf{x}_{[i],\mathcal{A}}^T([\hat{\boldsymbol{\beta}}_{\text{AT}}]_{\mathcal{A}} - [\boldsymbol{\beta}_0]_{\mathcal{A}})}{s_n(\mathbf{r}(\hat{\boldsymbol{\beta}}_{\text{AT}}))}\bigg)\mathbf{x}_{[i],\mathcal{A}}\mathbf{x}_{[i],\mathcal{A}}^T\bigg]$$

$$\times ([\hat{\boldsymbol{\beta}}_{\text{AT}}]_{\mathcal{A}} - [\boldsymbol{\beta}_0]_{\mathcal{A}})\bigg) +\sqrt{n}\lambda_n \nabla_{\boldsymbol{\beta}_{\mathcal{A}}}\bigg(\sum_{j=1}^{p}\frac{|\beta_j|}{|\tilde{\beta}_j|^\gamma}\bigg)\bigg|_{\boldsymbol{\beta}=\hat{\boldsymbol{\beta}}_{\text{AT}}} = \mathbf{0}_{k_0} \quad (S84)$$

where $0 < \zeta_i < 1$ for $i = 1, \cdots, n$. We shall now establish the asymptotic normality of $[\hat{\boldsymbol{\beta}}_{\text{AT}}]_{\mathcal{A}}$, that is, showing that $\sqrt{n}([\hat{\boldsymbol{\beta}}_{\text{AT}}]_{\mathcal{A}} - [\boldsymbol{\beta}_0]_{\mathcal{A}})$ converges in distribution to a zero-mean multivariate Gaussian with a covariance matrix, whose specific form will be derived in the subsequent lines. Continuing our analysis, we will isolate the terms involving $\sqrt{n}([\hat{\boldsymbol{\beta}}_{\text{AT}}]_{\mathcal{A}} - [\boldsymbol{\beta}_0]_{\mathcal{A}})$, move the remaining terms to the right-hand side of the equality and factor out the term $\sqrt{n}([\hat{\boldsymbol{\beta}}_{\text{AT}}]_{\mathcal{A}} - [\boldsymbol{\beta}_0]_{\mathcal{A}})$ from the left-hand side, resulting in

$$\Bigg[\frac{1}{n}\sum_{i=1}^{n}W_n^0\psi_0'\bigg(\frac{u_i - \zeta_i\mathbf{x}_{[i],\mathcal{A}}^T([\hat{\boldsymbol{\beta}}_{\text{AT}}]_{\mathcal{A}} - [\boldsymbol{\beta}_0]_{\mathcal{A}})}{s_n(\mathbf{r}(\hat{\boldsymbol{\beta}}_{\text{AT}}))}\bigg)\mathbf{x}_{[i],\mathcal{A}}\mathbf{x}_{[i],\mathcal{A}}^T$$

$$+ \frac{1}{n}\sum_{i=1}^{n}\psi_1'\bigg(\frac{u_i - \zeta_i\mathbf{x}_{[i],\mathcal{A}}^T([\hat{\boldsymbol{\beta}}_{\text{AT}}]_{\mathcal{A}} - [\boldsymbol{\beta}_0]_{\mathcal{A}})}{s_n(\mathbf{r}(\hat{\boldsymbol{\beta}}_{\text{AT}}))}\bigg)\mathbf{x}_{[i],\mathcal{A}}\mathbf{x}_{[i],\mathcal{A}}^T\Bigg]$$

$$\times \sqrt{n}([\hat{\boldsymbol{\beta}}_{\text{AT}}]_{\mathcal{A}} - [\boldsymbol{\beta}_0]_{\mathcal{A}})$$

$$= \Bigg[s_n(\mathbf{r}(\hat{\boldsymbol{\beta}}_{\text{AT}}))W_n^0\frac{1}{\sqrt{n}}\sum_{i=1}^{n}\psi_0\bigg(\frac{y_i - \mathbf{x}_{[i],\mathcal{A}}^T[\boldsymbol{\beta}_0]_{\mathcal{A}}}{s_n(\mathbf{r}(\hat{\boldsymbol{\beta}}_{\text{AT}}))}\bigg)\mathbf{x}_{[i],\mathcal{A}}$$

$$+ s_n(\mathbf{r}(\hat{\boldsymbol{\beta}}_{\text{AT}}))\frac{1}{\sqrt{n}}\sum_{i=1}^{n}\psi_1\bigg(\frac{y_i - \mathbf{x}_{[i],\mathcal{A}}^T[\boldsymbol{\beta}_0]_{\mathcal{A}}}{s_n(\mathbf{r}(\hat{\boldsymbol{\beta}}_{\text{AT}}))}\bigg)\mathbf{x}_{[i],\mathcal{A}}$$


$$+ s_n(\mathbf{r}(\hat{\boldsymbol{\beta}}_{\text{AT}}))W_n^0 R_n^0 + s_n(\mathbf{r}(\hat{\boldsymbol{\beta}}_{\text{AT}}))R_n^1 \Big]$$
$$- \sqrt{n}\lambda_n \nabla_{\boldsymbol{\beta}_{\mathcal{A}}}\Big(\sum_{j=1}^{p}\frac{|\beta_j|}{|\tilde{\beta}_j|^\gamma}\Big)\Bigg|_{\boldsymbol{\beta}=\hat{\boldsymbol{\beta}}_{\text{AT}}}. \quad (S85)$$

For the simplicity of exposition, we let

$$Q_n = \frac{1}{n}\sum_{i=1}^{n}W_n^0\psi_0'\Big(\frac{u_i - \zeta_i\mathbf{x}_{[i],\mathcal{A}}^T([\hat{\boldsymbol{\beta}}_{\text{AT}}]_{\mathcal{A}} - [\boldsymbol{\beta}_0]_{\mathcal{A}})}{s_n(\mathbf{r}(\hat{\boldsymbol{\beta}}_{\text{AT}}))}\Big)\mathbf{x}_{[i],\mathcal{A}}\mathbf{x}_{[i],\mathcal{A}}^T$$
$$+ \frac{1}{n}\sum_{i=1}^{n}\psi_1'\Big(\frac{u_i - \zeta_i\mathbf{x}_{[i],\mathcal{A}}^T([\hat{\boldsymbol{\beta}}_{\text{AT}}]_{\mathcal{A}} - [\boldsymbol{\beta}_0]_{\mathcal{A}})}{s_n(\mathbf{r}(\hat{\boldsymbol{\beta}}_{\text{AT}}))}\Big)\mathbf{x}_{[i],\mathcal{A}}\mathbf{x}_{[i],\mathcal{A}}^T \quad (S86)$$

that is the multiplier of $\sqrt{n}([\hat{\boldsymbol{\beta}}_{\text{AT}}]_{\mathcal{A}} - [\boldsymbol{\beta}_0]_{\mathcal{A}})$ in the equation (S85). To establish the convergence in distribution of $\sqrt{n}([\hat{\boldsymbol{\beta}}_{\text{AT}}]_{\mathcal{A}} - [\boldsymbol{\beta}_0]_{\mathcal{A}})$ to a zero-mean multivariate Gaussian, we shall now solve equation (S85) for $\sqrt{n}([\hat{\boldsymbol{\beta}}_{\text{AT}}]_{\mathcal{A}} - [\boldsymbol{\beta}_0]_{\mathcal{A}})$ as follows:

$$\sqrt{n}([\hat{\boldsymbol{\beta}}_{\text{AT}}]_{\mathcal{A}} - [\boldsymbol{\beta}_0]_{\mathcal{A}})$$
$$= \Big[s_n(\mathbf{r}(\hat{\boldsymbol{\beta}}_{\text{AT}}))Q_n^{-1}\Big(\underbrace{W_n^0\frac{1}{\sqrt{n}}\sum_{i=1}^{n}\psi_0\Big(\frac{y_i - \mathbf{x}_{[i],\mathcal{A}}^T[\boldsymbol{\beta}_0]_{\mathcal{A}}}{s_n(\mathbf{r}(\hat{\boldsymbol{\beta}}_{\text{AT}}))}\Big)\mathbf{x}_{[i],\mathcal{A}}}_{\text{first term}}$$
$$\underbrace{+ \frac{1}{\sqrt{n}}\sum_{i=1}^{n}\psi_1\Big(\frac{y_i - \mathbf{x}_{[i],\mathcal{A}}^T[\boldsymbol{\beta}_0]_{\mathcal{A}}}{s_n(\mathbf{r}(\hat{\boldsymbol{\beta}}_{\text{AT}}))}\Big)\mathbf{x}_{[i],\mathcal{A}}\Big)}_{\text{the remaining of first term}}$$
$$+ \underbrace{s_n(\mathbf{r}(\hat{\boldsymbol{\beta}}_{\text{AT}}))W_n^0 Q_n^{-1}R_n^0}_{\text{second term}} + \underbrace{s_n(\mathbf{r}(\hat{\boldsymbol{\beta}}_{\text{AT}}))Q_n^{-1}R_n^1}_{\text{third term}}\Big]$$
$$- \underbrace{\sqrt{n}\lambda_n Q_n^{-1}\nabla_{\boldsymbol{\beta}_{\mathcal{A}}}\Big(\sum_{j=1}^{p}\frac{|\beta_j|}{|\tilde{\beta}_j|^\gamma}\Big)\Bigg|_{\boldsymbol{\beta}=\hat{\boldsymbol{\beta}}_{\text{AT}}}}_{\text{fourth term}}. \quad (S87)$$

### C. Proof of step 3

In the final step, we shall prove that the term on the right-hand side of the above equation converges in distribution to a zero-mean multivariate Gaussian with a covariance matrix. The specific form of the covariance matrix will be presented here in equation (S91). To do so, we will first examine the asymptotic behavior of its constituting components each, separately, as follows:

- *Asymptotic behavior of $s_n(\mathbf{r}(\hat{\boldsymbol{\beta}}_{\text{AT}}))$:* By Lemma 3 of [S8] and strong consistency of $\hat{\boldsymbol{\beta}}_{\text{AT}}$ for $\boldsymbol{\beta}_0$ given by Proposition 2, we get $s_n(\mathbf{r}(\hat{\boldsymbol{\beta}}_{\text{AT}})) \xrightarrow{a.s.} s(\boldsymbol{\beta}_0)$.
- *Asymptotic behavior of $W_n$:* By Lemma 4.2 of [S9], the Continuous Mapping Theorem, Proposition 2, strong consistency of $s_n(\mathbf{r}(\hat{\boldsymbol{\beta}}_{\text{AT}}))$ for $s(\boldsymbol{\beta}_0)$, Lemmas 2 and 3 of [S8], and similar results in the proof of step 4 of Theorem 1, we have that $W_n^0 \xrightarrow{a.s.} W^0$, i.e., $W_n^0$ converges almost surely to $W^0$ defined as

$$W^0 = \frac{2\mathbb{E}_F\big[\rho_1\big(\frac{u}{s(\boldsymbol{\beta}_0)}\big)\big] - \mathbb{E}_F\big[\psi_1\big(\frac{u}{s(\boldsymbol{\beta}_0)}\big)\frac{u}{s(\boldsymbol{\beta}_0)}\big]}{\mathbb{E}_F\big[\psi_0\big(\frac{u}{s(\boldsymbol{\beta}_0)}\big)\frac{u}{s(\boldsymbol{\beta}_0)}\big]}. \quad (S88)$$

- *Asymptotic behavior of $Q_n$:* Strong consistency of $Q_n$ for $Q$ follows by Lemma 4.2 of [S9], Proposition 2, the Strong Law of Large Numbers, condition 4 of Assumption 3, strong consistency of $s_n(\hat{\boldsymbol{\beta}}_{\text{AT}})$ for $s(\boldsymbol{\beta}_0)$, and $W_n^0 \xrightarrow{a.s.} W^0$. Hence, we obtain

$$Q_n \xrightarrow{a.s.} \underbrace{\mathbb{E}_F\Big[W^0\psi_0'\big(\frac{u}{s(\boldsymbol{\beta}_0)}\big) + \psi_1'\big(\frac{u}{s(\boldsymbol{\beta}_0)}\big)\Big]\mathbb{E}_G\big[\mathbf{x}_{\mathcal{A}}\mathbf{x}_{\mathcal{A}}^T\big]}_{Q}. \quad (S89)$$

- *Asymptotic behavior of the **first term**:* On the other hand, by Lemma 5.1 of [S9], the Central Limit Theorem, $W_n^0 \xrightarrow{a.s.} W^0$, Proposition 2, and $s_n(\mathbf{r}(\hat{\boldsymbol{\beta}}_{\text{AT}})) \xrightarrow{a.s.} s(\boldsymbol{\beta}_0)$, it follows that

$$\frac{1}{\sqrt{n}}\sum_{i=1}^{n}\Big[W_n^0\psi_0\Big(\frac{y_i - \mathbf{x}_{[i],\mathcal{A}}^T[\boldsymbol{\beta}_0]_{\mathcal{A}}}{s_n(\mathbf{r}(\hat{\boldsymbol{\beta}}_{\text{AT}}))}\Big)$$
$$+ \psi_1\Big(\frac{y_i - \mathbf{x}_{[i],\mathcal{A}}^T[\boldsymbol{\beta}_0]_{\mathcal{A}}}{s_n(\mathbf{r}(\hat{\boldsymbol{\beta}}_{\text{AT}}))}\Big)\Big]\mathbf{x}_{[i],\mathcal{A}}$$
$$\xrightarrow{d} \mathcal{N}\Big(\mathbf{0}_{k_0}, \mathbb{E}_F\Big[\Big(W^0\psi_0\big(\frac{u}{s(\boldsymbol{\beta}_0)}\big) + \psi_1\big(\frac{u}{s(\boldsymbol{\beta}_0)}\big)\Big)^2\Big]\mathbf{V}_{\mathcal{A}}\Big). \quad (S90)$$

To simplify matters, it is convenient to define $\mathbf{V}_{\mathcal{A}} = \mathbb{E}_G\big[\mathbf{x}_{\mathcal{A}}\mathbf{x}_{\mathcal{A}}^T\big]$.

- *Asymptotic behavior of the **second** and **third terms**:* Using $\mathbb{P}(R_n^0 = \mathbf{0}_{k_0}) \to 1$ and $\mathbb{P}(R_n^1 = \mathbf{0}_{k_0}) \to 1$ as $n$ grows to infinity, we conclude that the terms $s_n(\mathbf{r}(\hat{\boldsymbol{\beta}}_{\text{AT}}))W_n^0 Q_n^{-1}R_n^0$ and $s_n(\mathbf{r}(\hat{\boldsymbol{\beta}}_{\text{AT}}))Q_n^{-1}R_n^1$ converge to $\mathbf{0}_{k_0}$ in probability.
- *Asymptotic behavior of the **fourth term**:* By the assumption $\sqrt{n}\lambda_n \to 0$ and strong consistency of $Q_n$ for $Q$, we have that $\sqrt{n}\lambda_n Q_n^{-1}\nabla_{\boldsymbol{\beta}_{\mathcal{A}}}(\sum_{j=1}^{p}|\beta_j|/|\tilde{\beta}_j|^\gamma)|_{\boldsymbol{\beta}=\hat{\boldsymbol{\beta}}_{\text{AT}}}$ converges almost surely to $\mathbf{0}_{k_0}$.

We are now in a position to establish the asymptotic normality of $[\hat{\boldsymbol{\beta}}_{\text{AT}}]_{\mathcal{A}}$. By Slutsky's theorem and the results derived in the preceding lines of the proof, it follows that

$$\sqrt{n}([\hat{\boldsymbol{\beta}}_{\text{AT}}]_{\mathcal{A}} - [\boldsymbol{\beta}_0]_{\mathcal{A}}) \xrightarrow{d} \mathcal{N}\Big(\mathbf{0}_{k_0}, s(\boldsymbol{\beta}_0)^2 \frac{a(\psi_0, \psi_1, F_0)}{\big(b(\psi_0', \psi_1', F_0)\big)^2}\mathbf{V}_{\mathcal{A}}^{-1}\Big) \quad (S91)$$

where $a(\psi_0, \psi_1, F_0)$ and $b(\psi_0', \psi_1', F_0)$ are defined as

$$a(\psi_0, \psi_1, F) = \mathbb{E}_F\Big[\Big(W^0\psi_0\big(\frac{u}{s(\boldsymbol{\beta}_0)}\big) + \psi_1\big(\frac{u}{s(\boldsymbol{\beta}_0)}\big)\Big)^2\Big],$$
$$b(\psi_0', \psi_1', F) = \mathbb{E}_F\Big[W^0\psi_0'\big(\frac{u}{s(\boldsymbol{\beta}_0)}\big) + \psi_1'\big(\frac{u}{s(\boldsymbol{\beta}_0)}\big)\Big]. \quad (S92)$$

The above results imply that the adaptive $\tau$-Lasso estimator of regression coefficients corresponding to truly active predictors is asymptotically normal. Combining Theorem 3 and Theorem 4 establishes the Oracle property of the adaptive $\tau$-Lasso estimator.

*Remark S8:* Note that we may reuse the symbols $Z_n^1$, $A_n^0$, $A_n^1$, $W_n^0$, $W^0$, $R_n^0$, $R_n^1$, $Q_n$, $Q$, $o_i^0$, and $o_i^1$ for similar or different purposes, meaning they are merely local to this theorem.

## S.VIII. Proof of Theorem 5

The finite-sample breakdown point of the adaptive $\tau$-Lasso estimator, denoted by $\varepsilon^*(\hat{\boldsymbol{\beta}}_{\mathrm{AT}};\mathbf{Z})$, can be derived by following the same line of reasoning as Theorem 1 of [S6], so we will only highlight the necessary modifications. Analogous to the proof given for the finite-sample breakdown point of the $\tau$-Lasso estimator, the proof comprises two primary stages.

- *Stage 1:* First, we establish that $\varepsilon^*(\hat{\boldsymbol{\beta}}_{\mathrm{AT}};\mathbf{Z})$ is bounded from below by $m(\delta)/n$.
- *Stage 2:* Second, we show that $\varepsilon^*(\hat{\boldsymbol{\beta}}_{\mathrm{AT}};\mathbf{Z})$ is bounded from above by $\delta$.

In order to present a more detailed proof, we first provide some necessary notation as follows:

- We use $(\mathbf{Z}_m^k)_{k\in\mathbb{N}} = ((\mathbf{y}_m^k)_{k\in\mathbb{N}}, (\mathbf{X}_m^k)_{k\in\mathbb{N}})$ to denote a sequence of samples obtained by replacing $m$ observations of $\mathbf{Z}$ with arbitrary values.
- $(\hat{\boldsymbol{\beta}}^k)_{k\in\mathbb{N}}$ represents the sequence of the adaptive $\tau$-Lasso estimates that are computed based on the contaminated sequence $(\mathbf{Z}_m^k)_{k\in\mathbb{N}}$. To simplify notation, here, the shorthand $(\hat{\boldsymbol{\beta}}^k)_{k\in\mathbb{N}}$ refers to $(\hat{\boldsymbol{\beta}}_{\mathrm{AT}}^k)_{k\in\mathbb{N}}$.
- $(\tilde{\boldsymbol{\beta}}^k)_{k\in\mathbb{N}}$ is the sequence of the pilot estimates computed based on the contaminated sequence $(\mathbf{Z}_m^k)_{k\in\mathbb{N}}$.
- We define $\breve{\boldsymbol{\beta}}$ to be a vector in $\mathbb{R}^p$ that is bounded in $\ell_1$-norm. The choice of boundedness in $\ell_1$-norm is specifically tailored to the adaptive $\tau$-Lasso problem and offers advantages in terms of derivations. In general, a bounded $\breve{\boldsymbol{\beta}}$ would suffice.

### A. Proof of stage 1: Bounding $\varepsilon^*(\hat{\boldsymbol{\beta}}_{\mathrm{AT}};\mathbf{Z})$ from below

We begin the proof of **Stage 1** by noting that the boundedness of $\varepsilon^*(\hat{\boldsymbol{\beta}}_{\mathrm{AT}};\mathbf{Z})$ from below by $m \leq m(\delta)$ implies $\hat{\boldsymbol{\beta}}_{\mathrm{AT}}$ remains bounded for $m \leq m(\delta)$. In order to establish the boundedness of $\hat{\boldsymbol{\beta}}_{\mathrm{AT}}$ from below, we shall use the proof by contradiction. That is, we assume the original statement to be false and the sequence $(\hat{\boldsymbol{\beta}}^k)_{k\in\mathbb{N}}$ to be unbounded. Using the definition of the adaptive $\tau$-Lasso estimator given by equation (4) within the main body of the paper, we will then show that the $\tau$-Lasso objective function at $\hat{\boldsymbol{\beta}}^k$ does not attain the minimum value. Hence, this leads to a contradiction, and $\hat{\boldsymbol{\beta}}^k$ shall remain bounded for all $m \leq m(\delta)$. The proof of boundedness from below proceeds in four steps as follows:

- *Step 1:* We first evaluate the adaptive $\tau$-Lasso objective function $\mathcal{L}_n(\boldsymbol{\beta};\mathbf{Z}_m^k)$ at some $\breve{\boldsymbol{\beta}} \in \mathbb{R}^p$ with a bounded $\ell_1$-norm.
- *Step 2:* For the second step, we evaluate the adaptive $\tau$-Lasso objective function $\mathcal{L}_n(\boldsymbol{\beta};\mathbf{Z}_m^k)$ at $\hat{\boldsymbol{\beta}}^k$.
- *Step 3:* Comparing $\mathcal{L}_n(\breve{\boldsymbol{\beta}};\mathbf{Z}_m^k)$ with $\mathcal{L}_n(\hat{\boldsymbol{\beta}}^k;\mathbf{Z}_m^k)$, we see that there exists a $k^\star$ such that for every $k \geq k^\star$

$$\mathcal{L}_n(\breve{\boldsymbol{\beta}};\mathbf{Z}_m^k) < \mathcal{L}_n(\hat{\boldsymbol{\beta}}^k;\mathbf{Z}_m^k) \tag{S93}$$

- *Step 4:* By combining the argument derived in the preceding step with $\hat{\boldsymbol{\beta}}^k = \mathrm{argmin}_{\boldsymbol{\beta}\in\mathbb{R}^p} \mathcal{L}_n(\boldsymbol{\beta};\mathbf{Z}_m^k)$, we arrive at a contradiction. Hence, we conclude $\hat{\boldsymbol{\beta}}^k$ must be bounded in order to attain the minimum condition $\hat{\boldsymbol{\beta}}^k = \mathrm{argmin}_{\boldsymbol{\beta}\in\mathbb{R}^p} \mathcal{L}_n(\boldsymbol{\beta};\mathbf{Z}_m^k)$.

In the subsequent paragraphs, we provide a rigorous proof by following the above steps to establish the boundedness of $\hat{\boldsymbol{\beta}}_{\mathrm{AT}}$ from below for $m \leq m(\delta)$.

*1) Step 1:* Let us begin by evaluating $\mathcal{L}_n(\breve{\boldsymbol{\beta}};\mathbf{Z}_m^k)$. By Theorem 1 of [S6], we have that $\tilde{\boldsymbol{\beta}}^k$ remains bounded for $m \leq m(\delta)$. Combining this with the $\ell_1$-norm boundedness of $\breve{\boldsymbol{\beta}}$, we obtain

$$\sum_{j=1}^p \frac{|\breve{\beta}_j|}{|\tilde{\beta}_j^k|^\gamma} < \infty \tag{S94}$$

where $\gamma$ is a positive constant. For the sake of convenience, we adopt the convention that $|\breve{\beta}_j|/|\tilde{\beta}_j^k|^\gamma = 0$ if $\tilde{\beta}_j^k = 0$. Recall that by Theorem 1 of [S6], $\tau_n(\mathbf{r}^k(\breve{\boldsymbol{\beta}},\mathbf{Z}_m^k))$ is bounded for $m \leq m(\delta)$ and some $\breve{\boldsymbol{\beta}}$ with a bounded $\ell_1$-norm. Hence, we get

$$\sup_{k'\in\mathbb{N}} \left(\tau_n^2(\mathbf{r}^{k'}(\breve{\boldsymbol{\beta}},\mathbf{Z}_m^{k'})) + \lambda_n \sum_{j=1}^p \frac{|\breve{\beta}_j|}{|\tilde{\beta}_j^{k'}|^\gamma}\right) < \infty, \tag{S95}$$

implying that

$$\mathcal{L}_n(\breve{\boldsymbol{\beta}};\mathbf{Z}_m^k) < \infty. \tag{S96}$$

Here, $\lambda_n$ determines the amount of regularization induced by $\ell_1$-norm at the adaptive $\tau$-Lasso estimation problem.

*2) Step 2:* Recall from the Theorem 1 of [S6] that $\tilde{\boldsymbol{\beta}}^k$ remains bounded for $m \leq m(\delta)$, while $\hat{\boldsymbol{\beta}}^k$ is unbounded by assumption for $m \leq m(\delta)$. Using these facts, we can conclude that the penalty term of $\mathcal{L}_n(\hat{\boldsymbol{\beta}}^k;\mathbf{Z}_m^k)$, denoted by

$$\lambda_n \sum_{j=1}^p \frac{|\hat{\beta}_j|}{|\tilde{\beta}_j^k|^\gamma}, \tag{S97}$$

should also be unbounded. Consequently, it follows that $\mathcal{L}_n(\hat{\boldsymbol{\beta}}^k;\mathbf{Z}_m^k)$ itself is unbounded. Combining the unboundedness of the penalty term in $\mathcal{L}_n(\hat{\boldsymbol{\beta}}^k;\mathbf{Z}_m^k)$ as presented in equation (S97) with the boundedness of the expression in equation (S95), we can conclude that there exists a sequence index $k^\star$ such that

$$\lambda_n \sum_{j=1}^p \frac{|\hat{\beta}_j^{k^\star}|}{|\tilde{\beta}_j^{k^\star}|^\gamma} > \sup_{k'\in\mathbb{N}} \left(\tau_n^2(\mathbf{r}^{k'}(\breve{\boldsymbol{\beta}},\mathbf{Z}_m^{k'})) + \lambda_n \sum_{j=1}^p \frac{|\breve{\beta}_j|}{|\tilde{\beta}_j^{k'}|^\gamma}\right). \tag{S98}$$

*3) Step 3:* Putting together the pieces, we find that for every $k \geq k^\star$,

$$\overbrace{\tau_n^2(\mathbf{r}^k(\hat{\boldsymbol{\beta}}^k, \mathbf{Z}_m^k)) + \lambda_n \sum_{j=1}^{p} \frac{|\hat{\beta}_j^k|}{|\tilde{\beta}_j^k|^\gamma}}^{\mathcal{L}_n(\hat{\boldsymbol{\beta}}^k \,; \mathbf{Z}_m^k)}$$

$$\stackrel{(i)}{>} \tau_n^2(\mathbf{r}^k(\hat{\boldsymbol{\beta}}^k, \mathbf{Z}_m^k))$$
$$+ \sup_{k' \in \mathbb{N}} \left( \tau_n^2(\mathbf{r}^{k'}(\breve{\boldsymbol{\beta}}, \mathbf{Z}_m^{k'})) + \lambda_n \sum_{j=1}^{p} \frac{|\breve{\beta}_j|}{|\tilde{\beta}_j^{k'}|^\gamma} \right)$$
$$\stackrel{(ii)}{>} \tau_n^2(\mathbf{r}^k(\breve{\boldsymbol{\beta}}, \mathbf{Z}_m^k)) + \lambda_n \sum_{j=1}^{p} \frac{|\breve{\beta}_j|}{|\tilde{\beta}_j^k|^\gamma}$$
$$= \mathcal{L}_n(\breve{\boldsymbol{\beta}} \,; \mathbf{Z}_m^k) \quad \text{(S99)}$$

where the inequality $(i)$ is derived by adding $\tau_n^2(\mathbf{r}^k(\hat{\boldsymbol{\beta}}^k, \mathbf{Z}_m^k))$ to both sides of the inequality in equation (S98). The inequality (ii) follows from the definition of supremum and the fact that $\tau_n^2(\mathbf{r}^k(\hat{\boldsymbol{\beta}}^k, \mathbf{Z}_m^k))$ is positive. Therefore, for every $k \geq k^*$

$$\mathcal{L}_n(\hat{\boldsymbol{\beta}}^k \,; \mathbf{Z}_m^k) > \mathcal{L}_n(\breve{\boldsymbol{\beta}} \,; \mathbf{Z}_m^k). \quad \text{(S100)}$$

Please note that, for clarity, we introduce the notation $k'$ to distinguish the sequence index inside the supremum from the sequence index $k$ for terms outside it.

*4) Step 4:* The loss associated with $\hat{\boldsymbol{\beta}}^k$ is larger, implying that the sequence $\hat{\boldsymbol{\beta}}^k$ fails to attain the minimum condition. Hence, $\hat{\boldsymbol{\beta}}^k$ shall remain bounded for $m \leq m(\delta)$.

## B. Proof of stage 2: Bounding $\varepsilon^*(\hat{\boldsymbol{\beta}}_{\mathrm{AT}}; \mathbf{Z})$ from above

We establish the boundedness of $\varepsilon^*(\hat{\boldsymbol{\beta}}_{\mathrm{AT}}; \mathbf{Z})$ from above by showing that $\hat{\boldsymbol{\beta}}_{\mathrm{AT}}$ breaks down for $m > n\delta$. Analogous to the proof of Theorem 1 of [S6], our strategy relies on *proof by contradiction*. More precisely, we assume the sequence of estimates $(\hat{\boldsymbol{\beta}}^k)_{k \in \mathbb{N}}$ to be bounded for $m > n\delta$. Under the stated assumption, we show that $\hat{\boldsymbol{\beta}}^k$ is not a minimizer of the adaptive $\tau$-Lasso given by equation (4) in the main body of the paper. As a result, the minimality assumption of $\hat{\boldsymbol{\beta}}^k$ is violated, and $\hat{\boldsymbol{\beta}}^k$ shall be unbounded. Our route to establish boundedness from above is outlined as follows:

- *Step 1:* We first evaluate the adaptive $\tau$-Lasso objective function $\mathcal{L}_n(\boldsymbol{\beta} \,; \mathbf{Z}_m^k)$ at the bounded sequence of estimates $\hat{\boldsymbol{\beta}}^k$.
- *Step 2:* Given an unbounded sequence $\breve{\boldsymbol{\beta}}^k$, we then turn to evaluate the adaptive $\tau$-Lasso objective function $\mathcal{L}_n(\boldsymbol{\beta} \,; \mathbf{Z}_m^k)$ at $\breve{\boldsymbol{\beta}}^k$.
- *Step 3:* By comparing $\mathcal{L}_n(\breve{\boldsymbol{\beta}}^k \,; \mathbf{Z}_m^k)$ with $\mathcal{L}_n(\hat{\boldsymbol{\beta}}^k \,; \mathbf{Z}_m^k)$, it follows that there exists a $k^*$ such that for all $k \geq k^*$

$$\mathcal{L}_n(\hat{\boldsymbol{\beta}}^k \,; \mathbf{Z}_m^k) > \mathcal{L}_n(\breve{\boldsymbol{\beta}}^k \,; \mathbf{Z}_m^k) \quad \text{(S101)}$$

- *Step 4:* The above statement implies $\hat{\boldsymbol{\beta}}^k$ is not a global minimizer of the adaptive $\tau$-Lasso objective function. Hence, the sequence of estimates $\hat{\boldsymbol{\beta}}^k$ shall be unbounded for $m > n\delta$, and our claim is established.

In the subsequent paragraphs, we provide proof of boundedness from above by following the above steps.

*1) Step 1: Evaluation of adaptive $\tau$-Lasso objective function for $\hat{\boldsymbol{\beta}}^{(k)}$:* In order to evaluate the adaptive $\tau$-Lasso objective function $\mathcal{L}_n(\boldsymbol{\beta} \,; \mathbf{Z}_m^k)$ at $\hat{\boldsymbol{\beta}}^k$, it is convenient to split $\mathcal{L}_n(\boldsymbol{\beta} \,; \mathbf{Z}_m^k)$ into three primary components and work with each separately. The objective function $\mathcal{L}_n(\boldsymbol{\beta} \,; \mathbf{Z}_m^k)$ comprises three primary components as follows:

- Component 1: M-scale of residuals, $s_n(\mathbf{r}^k(\boldsymbol{\beta}, \mathbf{Z}_m^k))$
- Component 2: $\frac{1}{n} \sum_{i=1}^{n} \rho_1\left( \frac{r_i^k(\boldsymbol{\beta}, \mathbf{Z}_m^k)}{s_n(\mathbf{r}^k(\boldsymbol{\beta}, \mathbf{Z}_m^k))} \right)$,
- Component 3: $\lambda_n \sum_{j=1}^{p} |\beta_j| / |\tilde{\beta}_j^k|^\gamma$

Let us begin by constructing a sequence of contaminated samples with $m$ outliers as follows:

$$(y_{m,i}^k, \mathbf{x}_{m,[i]}^k) = \begin{cases} (k^{\nu+1}, \mathbf{x}_{[0]} k) & i \in C \\ (y_i, \mathbf{x}_{[i]}) & i \notin C \end{cases}, \quad \text{(S102)}$$

where $C \subset \{1, \cdots, n\}$ represents the indices of observations replaced by outliers, $0 < \nu \leq 1$ and $(k^{\nu+1}, \mathbf{x}_{[0]} k)$ corresponds to outlying observations. For small values of $k$, the outlying impact of contaminated observations is relatively minimal. However, as the sequence index $k$ grows, the influence of outlying observations becomes more pronounced. The sequence for outlying observations diverges as $k \to \infty$. Without loss of generality, $\mathbf{x}_{[0]} \in \mathbb{R}^p$ is assumed to have a unit $\ell_2$-norm, $\|\mathbf{x}_{[0]}\|_{\ell_2} = 1$.

Following the same line of reasoning as the proof of Theorem 1 of [S6], we obtain the same expressions for components 1 and 2. We now turn to analyze component 3 at $\hat{\boldsymbol{\beta}}^k$. By Theorem 1 of [S6], we know that $\tilde{\boldsymbol{\beta}}^k$ is unbounded for $m > n\delta$. Combining this with the boundedness assumption of the sequence $\hat{\boldsymbol{\beta}}^k$, we get

$$\lambda_n \sum_{j=1}^{p} \frac{|\hat{\beta}_j^k|}{|\tilde{\beta}_j^k|^\gamma} < \infty, \quad \text{(S103)}$$

i.e., component 3 remains bounded for $m > n\delta$ under the stated assumption.

Following the same procedure as in the proof of Theorem 1 of [S6], we can evaluate the normalized adaptive $\tau$-Lasso objective function when $k$ goes to infinity, as follows:

$$\lim_{k \to \infty} \frac{\mathcal{L}_n(\hat{\boldsymbol{\beta}}^k \,; \mathbf{Z}_m^k)}{k^{2\nu+2}}$$
$$= \lim_{k \to \infty} \left( \frac{\tau_n^2(\mathbf{r}^k(\hat{\boldsymbol{\beta}}^k, \mathbf{Z}_m^k))}{k^{2\nu+2}} + \frac{\lambda_n}{k^{2\nu+2}} \times \sum_{j=1}^{p} \frac{|\hat{\beta}_j^k|}{|\tilde{\beta}_j^k|^\gamma} \right). \quad \text{(S104)}$$

Using the boundedness of the regularization term, we have

$$\lim_{k \to \infty} \frac{1}{k^{2\nu+2}} \times \sum_{j=1}^{p} \frac{|\hat{\beta}_j^k|}{|\tilde{\beta}_j^k|^\gamma} = 0. \quad \text{(S105)}$$

Letting $\vartheta = \lim_{k \to \infty} s_n(\mathbf{r}^k(\hat{\boldsymbol{\beta}}^k, \mathbf{Z}_m^k))/k^{\nu+1}$ and combining the derived expressions for components 1 and 2 as outlined in the proof of Theorem 1 in [S6], specifically in step 1 of **stage 2**, we obtain

$$\lim_{k \to \infty} \frac{\tau_n^2(\mathbf{r}^k(\hat{\boldsymbol{\beta}}^k, \mathbf{Z}_m^k))}{k^{2\nu+2}} = \frac{m\vartheta^2}{n}\rho_1\left(\frac{1}{\vartheta}\right). \quad \text{(S106)}$$

Putting together the pieces yields that

$$\lim_{k \to \infty} \frac{\mathcal{L}_n(\hat{\boldsymbol{\beta}}^k; \mathbf{Z}_m^k)}{k^{2\nu+2}} = \frac{m\vartheta^2}{n}\rho_1\left(\frac{1}{\vartheta}\right). \quad \text{(S107)}$$

Note that $k$ represents the sequence index, and for each value of $k$, we obtain an adaptive $\tau$-Lasso estimate based on the contaminated sample $\mathbf{Z}_m^k$, which corresponds to a different set of outliers. As we increase $k$ to infinity, the contaminated datasets may contain arbitrarily large outliers.

*2) Step 2: Evaluation of adaptive $\tau$-Lasso Objective Function for $\breve{\boldsymbol{\beta}}^k$:* We now evaluate the adaptive $\tau$-Lasso objective function $\mathcal{L}_n(\boldsymbol{\beta}; \mathbf{Z}_m^k)$ at $\breve{\boldsymbol{\beta}}^k$ by following the strategy outlined in the proof of Theorem 1 in [S6], as part of step 2 in **stage 2**. To begin, let $\breve{\boldsymbol{\beta}}^k = k^\nu \mathbf{x}_{[0]}/2$ be the unbounded sequence where $\|\mathbf{x}_{[0]}\|_{\ell_2}^2 = 1$. We obtain the same expressions as the proof of Theorem 1 of [S6] for components 1 and 2 evaluated at $\breve{\boldsymbol{\beta}}^k$, so that it remains to calculate component 3 at $\breve{\boldsymbol{\beta}}^k$. Recall that the sequence of estimates $\tilde{\boldsymbol{\beta}}^k$ remains unbounded for $m > n\delta$. Combining this with unboundedness assumption of the sequence $\breve{\boldsymbol{\beta}}^k$, we get

$$\lim_{k \to \infty} \frac{1}{k^{2\nu+2}} \times \sum_{j=1}^p \frac{|\breve{\beta}_j^k|}{|\tilde{\beta}_j^k|^\gamma} = \lim_{k \to \infty} \frac{1}{k^{2\nu+2}} \times \sum_{j=1}^p \frac{|k^\nu x_{0,j}|}{|\tilde{\beta}_j^k|^\gamma}$$

$$= \lim_{k \to \infty} \frac{1}{k^{\nu+2}} \times \underbrace{\sum_{j=1}^p \frac{|x_{0,j}|}{|\tilde{\beta}_j^k|^\gamma}}_{bounded}$$

$$= 0. \quad \text{(S108)}$$

By the proof of Theorem 1 in [S6], specifically step 2 of **stage 2**, we have that $\lim_{k \to \infty} s_n(\mathbf{r}^k(\breve{\boldsymbol{\beta}}^k, \mathbf{Z}_m^k))/(k^{\nu+1}/2)$ is also equal to $\vartheta$. Using the relation $\vartheta = \lim_{k \to \infty} s_n(\mathbf{r}^k(\breve{\boldsymbol{\beta}}^k, \mathbf{Z}_m^k))/(k^{\nu+1}/2)$ and combining the derived expressions for components 1 and 2 presented in the proof of Theorem 1 in [S6], as outlined in step 1 of stage 2, we conclude that

$$\lim_{k \to \infty} \frac{\tau_n^2(\mathbf{r}^k(\breve{\boldsymbol{\beta}}^k, \mathbf{Z}_m^k))}{k^{2\nu+2}} = \frac{m\vartheta^2}{4n}\rho_1\left(\frac{1}{\vartheta}\right). \quad \text{(S109)}$$

Putting together the pieces yields

$$\lim_{k \to \infty} \frac{\mathcal{L}_n(\breve{\boldsymbol{\beta}}^k, \mathbf{Z}_m^k)}{k^{2\nu+2}}$$

$$= \lim_{k \to \infty} \left( \frac{\tau_n^2(\mathbf{r}^k(\breve{\boldsymbol{\beta}}^k, \mathbf{Z}_m^k))}{k^{2\nu+2}} + \frac{\lambda_n}{k^{2\nu+2}} \times \sum_{j=1}^p \frac{|\breve{\beta}_j^k|}{|\tilde{\beta}_j^k|^\gamma} \right)$$

$$= \frac{m\vartheta^2}{4n}\rho_1\left(\frac{1}{\vartheta}\right). \quad \text{(S110)}$$

*3) Steps 3-4: Comparison:* By comparing the expression of equation (S107) with that of equation (S110), we conclude that for large enough $k^\star$,

$$\frac{\mathcal{L}_n(\breve{\boldsymbol{\beta}}^k, \mathbf{Z}_m^k)}{k^{2\nu+2}} < \frac{\mathcal{L}_n(\hat{\boldsymbol{\beta}}^k, \mathbf{Z}_m^k)}{k^{2\nu+2}}, \quad \forall k \geq k^\star. \quad \text{(S111)}$$

Canceling out a factor of $1/k^{2\nu+2}$ from both sides, we have

$$\mathcal{L}_n(\breve{\boldsymbol{\beta}}^k, \mathbf{Z}_m^k) < \mathcal{L}_n(\hat{\boldsymbol{\beta}}^k, \mathbf{Z}_m^k), \quad \forall k \geq k^\star. \quad \text{(S112)}$$

The above expression contradicts the fact that the adaptive $\tau$-Lasso objective function attains its minimum value at $\hat{\boldsymbol{\beta}}^k$, and implies that the sequence of estimates $\hat{\boldsymbol{\beta}}^k$ must be unbounded. Thus, the adaptive $\tau$-Lasso estimator breaks down for $m > n\delta$ and retains the finite-sample breakdown point of $\tau$-Lasso estimator.

*Remark S9*: Note that the symbols $C$ and $\vartheta$ are merely local to this theorem and may be reused for other purposes.

## S.IX. From adaptive $\tau$-Lasso to two-stage regularized $M$-estimators

Interestingly, one may express the adaptive $\tau$-Lasso estimator in the standard form of regularized $M$-estimators as shown in equation (22) in the main body of the paper. It follows from the first-order condition, commonly referred to as the zero-gradient condition, that the adaptive $\tau$-Lasso estimates of regression parameter vector $\hat{\boldsymbol{\beta}}_{\text{AT}}$ and scale $s_n$ shall satisfy the following system of equations:

$$-\frac{1}{n}\sum_{i=1}^n [W_n \psi_0\left(\frac{r_i(\hat{\boldsymbol{\beta}}_{\text{AT}})}{s_n}\right) + \psi_1\left(\frac{r_i(\hat{\boldsymbol{\beta}}_{\text{AT}})}{s_n}\right)]\mathbf{x}_{[i]}s_n$$

$$+\underline{\lambda}_n \sum_{j=1}^p \frac{\text{sgn}(\hat{\beta}_{\text{AT},j})}{|\hat{\beta}_{\text{PT},j}|} = \mathbf{0}_p, \quad \text{(S113a)}$$

$$\frac{1}{n}\sum_{i=1}^n \rho_0\left(\frac{r_i(\hat{\boldsymbol{\beta}}_{\text{AT}})}{s_n}\right) - \delta = 0, \quad \text{(S113b)}$$

where $r_i(\hat{\boldsymbol{\beta}}_{\text{AT}}) = y_i - \mathbf{x}_{[i]}^T \hat{\boldsymbol{\beta}}_{\text{AT}}$, and $W_n$ is defined as

$$W_n = \frac{\sum_{i=1}^n [2\rho_1\left(\frac{r_i(\hat{\boldsymbol{\beta}}_{\text{AT}})}{s_n}\right) - \psi_1\left(\frac{r_i(\hat{\boldsymbol{\beta}}_{\text{AT}})}{s_n}\right)\frac{r_i(\hat{\boldsymbol{\beta}}_{\text{AT}})}{s_n}]}{\sum_{i=1}^n \psi_0\left(\frac{r_i(\hat{\boldsymbol{\beta}}_{\text{AT}})}{s_n}\right)\frac{r_i(\hat{\boldsymbol{\beta}}_{\text{AT}})}{s_n}}. \quad \text{(S114)}$$

In order to represent the adaptive $\tau$-Lasso estimator in the standard form of a regularized $M$-estimator, we first need to express $\hat{\boldsymbol{\theta}}$ in terms of $s_n$ and $\hat{\boldsymbol{\beta}}_{\text{AT}}$. Before proceeding, we highly recommend the reader to review the section VI-B2 within the main body of the paper. This will serve as a helpful reminder and provide familiarity with the notation and the standard form of regularized $M$-estimators. We recall that any estimator $T(H_n)$, which can be characterized as the solution of

$$\left[\mathbb{E}_{H_n}[\Psi(\mathbf{z}, \boldsymbol{\theta})] + \frac{\partial q(\boldsymbol{\theta}; \lambda)}{\partial \boldsymbol{\theta}}\right]_{\boldsymbol{\theta}=T(H_n)} = \mathbf{0}_{p+1} \quad \text{(S115)}$$

is a regularized $M$-estimator. Here, $\mathbb{E}_{H_n}[\Psi(\mathbf{z}, \boldsymbol{\theta})]$ and $\partial q(\boldsymbol{\theta}; \lambda)/\partial \boldsymbol{\theta}$ represent the gradients of the data discrepancy term and the penalty term with respect to $\boldsymbol{\theta}$, respectively. We shall now determine the corresponding equivalents for each of the constituent terms, namely $\mathbb{E}_{H_n}[\Psi(\mathbf{z}, \boldsymbol{\theta})]$ and $\partial q(\boldsymbol{\theta}; \lambda)/\partial \boldsymbol{\theta}$ evaluated at $\boldsymbol{\theta} = T(H_n)$, within the estimating equations for the standard form of regularized $M$-estimators.

- *Expressing $\hat{\boldsymbol{\theta}}$ in terms of $s_n$ and $\hat{\boldsymbol{\beta}}_{\text{AT}}$*: We define the regularized *M*-estimator $\hat{\boldsymbol{\theta}}$ as

$$\hat{\boldsymbol{\theta}} = \begin{bmatrix} s_n \\ \hat{\boldsymbol{\beta}}_{\text{AT}} \end{bmatrix}, \quad (S116)$$

where $s_n$ and $\hat{\boldsymbol{\beta}}_{\text{AT}}$ are obtained from the estimating equations (S113). To ease mathematical derivation of the influence function, we express the regularized *M*-estimator $\hat{\boldsymbol{\theta}}$ as a functional $T(H_n)$ of the empirical distribution $H_n$, where $\hat{\boldsymbol{\theta}}$ consists of two components: $s_n$ and $\hat{\boldsymbol{\beta}}_{\text{AT}}$. To represent these components, we also define the *M*-scale estimator of the residual vector $s_n$ as a functional of $H_n$ and the adaptive $\tau$-Lasso estimates of regression parameter vector $\hat{\boldsymbol{\beta}}_{\text{AT}}$ as $T_{\boldsymbol{\beta}}(H_n)$ functional of $H_n$ as follows:

$$\hat{\boldsymbol{\theta}} = \begin{bmatrix} S(H_n) \\ T_{\boldsymbol{\beta}}(H_n) \end{bmatrix} = T(H_n). \quad (S117)$$

- *Equivalent term for* $\mathbb{E}_{H_n}[\Psi(\mathbf{z}, \hat{\boldsymbol{\theta}})]$: To proceed, let us recall that the expected value of $\Psi(\mathbf{z}, \hat{\boldsymbol{\theta}})$ with respect to the empirical distribution $H_n$, which is basically the sample average of the terms $\Psi(\mathbf{z}_i, \hat{\boldsymbol{\theta}})$, can be written as

$$\mathbb{E}_{H_n}[\Psi(\mathbf{z}, \hat{\boldsymbol{\theta}})] = \frac{1}{n} \sum_{i=1}^{n} \Psi(\mathbf{z}_i, \hat{\boldsymbol{\theta}}). \quad (S118)$$

Hence, we only need to determine the corresponding term for $\Psi(\mathbf{z}_i, \hat{\boldsymbol{\theta}})$, which can be obtained as

$$\Psi(\mathbf{z}_i, \hat{\boldsymbol{\theta}}) = \begin{bmatrix} \rho_0(\frac{r_i(\hat{\boldsymbol{\beta}}_{\text{AT}})}{s_n}) - \delta \\ -[W_n \psi_0(\frac{r_i(\hat{\boldsymbol{\beta}}_{\text{AT}})}{s_n}) + \psi_1(\frac{r_i(\hat{\boldsymbol{\beta}}_{\text{AT}})}{s_n})] \mathbf{x}_{[i]} s_n \end{bmatrix}. \quad (S119)$$

- *Equivalent term for* $\partial q(\boldsymbol{\theta}\ ;\lambda)/\partial\boldsymbol{\theta}$ *evaluated at* $\boldsymbol{\theta}=\hat{\boldsymbol{\theta}}$: We still need to find the corresponding term for $\partial q(\boldsymbol{\theta}\ ;\lambda)/\partial\boldsymbol{\theta}$ in the adaptive $\tau$-Lasso when evaluated at $\boldsymbol{\theta} = \hat{\boldsymbol{\theta}}$. It is evident that $\partial q(\boldsymbol{\theta}\ ;\lambda)/\partial\boldsymbol{\theta}$ should be linked with the regularization term and is dependent on the initial estimator of the regression parameter vector $\hat{\boldsymbol{\beta}}_{\text{PT}}$. To accommodate for the two-stage nature of adaptive $\tau$-Lasso and its dependence on $\hat{\boldsymbol{\beta}}_{\text{PT}}$, we will adopt a slightly different notation for $\partial q(\boldsymbol{\theta}\ ;\lambda)/\partial\boldsymbol{\theta}$ and denote it as $\partial q(\boldsymbol{\theta}, \underline{\hat{\boldsymbol{\theta}}}\ ; \lambda_n)/\partial\boldsymbol{\theta}$. We define the initial estimator $\underline{\hat{\boldsymbol{\theta}}}$ with functional representation of $\underline{T}(H_n)$ as follows:

$$\underline{\hat{\boldsymbol{\theta}}} = \begin{bmatrix} \underline{s}_n \\ \hat{\boldsymbol{\beta}}_{\text{PT}} \end{bmatrix} = \underline{T}(H_n), \quad (S120)$$

where $\underline{s}_n$ denotes the *M*-scale estimator of the residual scale, which is obtained using the $\tau$-Lasso estimator. Alternatively, the initial estimator $\underline{T}(H_n)$ may be expressed in terms of $\underline{S}(H_n)$, a functional representation of $\underline{s}_n$ and $\underline{T}_{\boldsymbol{\beta}}(H_n)$, a functional representation of $\hat{\boldsymbol{\beta}}_{\text{PT}}$, as follows:

$$\underline{T}(H_n) = \begin{bmatrix} \underline{S}(H_n) \\ \underline{T}_{\boldsymbol{\beta}}(H_n) \end{bmatrix}. \quad (S121)$$

By performing some basic calculus, we now have

$$\frac{\partial q(\boldsymbol{\theta}, \underline{\hat{\boldsymbol{\theta}}}\ ; \lambda_n)}{\partial \boldsymbol{\theta}}\Big|_{\boldsymbol{\theta}=\hat{\boldsymbol{\theta}}} = \begin{bmatrix} 0 \\ \lambda_n \frac{\text{sgn}(\hat{\beta}_{\text{AT},1})}{|\hat{\beta}_{\text{PT},1}|} \\ \vdots \\ \lambda_n \frac{\text{sgn}(\hat{\beta}_{\text{AT},p})}{|\hat{\beta}_{\text{PT},p}|} \end{bmatrix}. \quad (S122)$$

Building on the previous definitions, the adaptive $\tau$-Lasso estimating equations can be rewritten in the standard form of regularized *M*-estimators as follows:

$$\frac{1}{n}\sum_{i=1}^{n}\Psi(\mathbf{z}_i,\hat{\boldsymbol{\theta}}) + \frac{\partial q(\boldsymbol{\theta}, \underline{\hat{\boldsymbol{\theta}}}\ ; \lambda_n)}{\partial \boldsymbol{\theta}}\Big|_{\boldsymbol{\theta}=\hat{\boldsymbol{\theta}}} = \mathbf{0}_{p+1}. \quad (S123)$$

It is important to note that the two estimating equations (S113) have now been combined into a compact form of one estimating equation as given above.

## S.X. ESTIMATING EQUATIONS FOR THE TWO-STAGE REGULARIZED *M*-ESTIMATOR: POPULATION VERSION

Let us now return to the original problem at hand, studying the influence function of the adaptive $\tau$-Lasso estimator. We shall investigate the influence of an infinitesimal amount of contamination on the functional $\boldsymbol{\theta}_\infty = T(H)$. To do so, we shall modify the estimation equations of the adaptive $\tau$-Lasso expressed in the standard form of regularized *M*-estimators and rewrite them in terms of population quantities. We then adapt the tools developed for the class of two-stage regularized *M*-estimators [S14] to the setting mentioned above.

Herein, we outline the estimating equations for the adaptive $\tau$-Lasso in the standard form of regularized *M*-estimators, which are presented in two stages, as follows:

- *Stage 1:* We note that the adaptive $\tau$-Lasso functional $T(H)$ is obtained via a two-stage procedure that improves upon a preliminary $\tau$-Lasso functional $\underline{T}(H)$. It can be defined as a solution to

$$\left[\mathbb{E}_H[\Psi(\mathbf{z},\boldsymbol{\theta})] + \frac{\partial q(\boldsymbol{\theta}, \underline{T}(H)\ ; \lambda_n)}{\partial \boldsymbol{\theta}}\right]_{\boldsymbol{\theta}=T(H)} = \mathbf{0}_{p+1}, \quad (S124)$$

with

$$\Psi(\mathbf{z},\boldsymbol{\theta}) = \begin{bmatrix} \rho_0(\tilde{r}(\boldsymbol{\theta})) - \delta \\ -\psi(\tilde{r}(\boldsymbol{\theta}))\mathbf{x}s_\sigma \end{bmatrix} \quad \text{and} \quad (S125a)$$

$$\frac{\partial q(\boldsymbol{\theta}, \underline{T}(H)\ ; \lambda_n)}{\partial \boldsymbol{\theta}} = \begin{bmatrix} 0 \\ \lambda_n \frac{\text{sgn}(\beta_1)}{|\underline{\beta}_{1,\infty}|} \\ \vdots \\ \lambda_n \frac{\text{sgn}(\beta_p)}{|\underline{\beta}_{p,\infty}|} \end{bmatrix}, \quad (S125b)$$

where $\tilde{r}(\boldsymbol{\theta}) = (y - \mathbf{x}^T\boldsymbol{\beta})/s_\sigma$ and $\psi(\tilde{r}(\boldsymbol{\theta}))$ is given by

$$\psi(\tilde{r}(\boldsymbol{\theta})) = W(\tilde{r}(\boldsymbol{\theta}))\psi_0(\tilde{r}(\boldsymbol{\theta})) + \psi_1(\tilde{r}(\boldsymbol{\theta})). \quad (S126)$$

Moreover, $W(\tilde{r}(\boldsymbol{\theta}))$ is defined by

$$W(\tilde{r}(\boldsymbol{\theta})) = \frac{\left(2\mathbb{E}_H[\rho_1(\tilde{r}(\boldsymbol{\theta}))] - \mathbb{E}_H[\psi_1(\tilde{r}(\boldsymbol{\theta}))\tilde{r}(\boldsymbol{\theta})]\right)}{\mathbb{E}_H[\psi_0(\tilde{r}(\boldsymbol{\theta}))\tilde{r}(\boldsymbol{\theta})]}. \quad (S127)$$

Throughout this article, we use $q^{'}(\boldsymbol{\theta}, \underline{T}(H)\ ;\lambda_n)$ as a shorthand for $\partial q(\boldsymbol{\theta}, \underline{T}(H)\ ;\lambda_n)/\partial\boldsymbol{\theta}$.

- *Stage 2:* Furthermore, the functional $\underline{T}(H)$, which represents the asymptotic value of the $\tau$-Lasso estimator denoted as $\underline{\boldsymbol{\theta}}_\infty$ shall satisfy the estimating equations of the $\tau$-Lasso estimator as given by

$$\left[\mathbb{E}_H[\Psi(\mathbf{z},\underline{\boldsymbol{\theta}})] + \frac{\partial \underline{q}(\underline{\boldsymbol{\theta}}\ ;\lambda_n)}{\partial \underline{\boldsymbol{\theta}}}\right]_{\underline{\boldsymbol{\theta}}=\underline{T}(H)} = \mathbf{0}_{p+1}. \quad (S128)$$

where

$$\Psi(\mathbf{z}, \underline{\boldsymbol{\theta}}) = \begin{bmatrix} \rho_0(\tilde{r}(\underline{\boldsymbol{\theta}})) - \delta \\ -\psi(\tilde{r}(\underline{\boldsymbol{\theta}}))\mathbf{x}_{\underline{s}_\sigma} \end{bmatrix} \quad \text{and}$$

$$\frac{\partial \underline{q}(\underline{\boldsymbol{\theta}}\,;\underline{\lambda}_n)}{\partial \underline{\boldsymbol{\theta}}} = \begin{bmatrix} 0 \\ \underline{\lambda}_n \operatorname{sgn}(\underline{\boldsymbol{\beta}}) \end{bmatrix}. \quad (S129a)$$

Note that, we obtain $\tilde{r}(\underline{\boldsymbol{\theta}})$ and $\psi(\tilde{r}(\underline{\boldsymbol{\theta}}))$ by replacing each occurrence of $\boldsymbol{\theta}$ with $\underline{\boldsymbol{\theta}}$ in $\tilde{r}(\boldsymbol{\theta})$ and $\psi(\tilde{r}(\boldsymbol{\theta}))$, respectively. To simplify notation, we adopt the convenient shorthand $\underline{q}'(\underline{\boldsymbol{\theta}}\,;\underline{\lambda}_n)$ for $\partial \underline{q}(\underline{\boldsymbol{\theta}}\,;\underline{\lambda}_n)/\partial \underline{\boldsymbol{\theta}}$.

## S.XI. PROOF OF THEOREM 6

We first provide a sketch of obtaining the influence function of the regularized *M*-estimators with a non-smooth penalty. In the remainder of the proof, we will derive the influence function of the $\tau$-Lasso estimator by extending the analytical tools developed for studying the influence function of the regularized *M*-estimators with the non-smooth (non-differentiable) penalty to the current setting.

Before we proceed with proofs, we will establish all the necessary notations required to have a clear understanding of the proof and subsequent derivation of the influence function.

- We use $\underline{q}(\underline{\boldsymbol{\theta}}\,;\underline{\lambda}_n)$ and $\underline{q}_t(\underline{\boldsymbol{\theta}}\,;\underline{\lambda}_n)$ to represent the regularization term of the $\tau$-Lasso estimator expressed in the standard form of regularized *M*-estimators and its smooth approximation, respectively.
- $\underline{T}(H\,;\underline{q}_t) = [\underline{T}_1(H\,;\underline{q}_t), \cdots, \underline{T}_{p+1}(H\,;\underline{q}_t)]^T$ denotes a sequence of approximating regularized *M*-estimators that is obtained by replacing the non-smooth $\underline{q}(\underline{\boldsymbol{\theta}}\,;\underline{\lambda}_n)$ with a smooth $\underline{q}_t(\underline{\boldsymbol{\theta}}\,;\underline{\lambda}_n)$ when $\mathbf{z}$ follows a distribution $H$.
- Likewise, we define $\underline{T}(H_\epsilon\,;\underline{q}_t)$ as the sequence of approximating regularized *M*-estimators resulting from the $\epsilon$-contamination of the underlying distribution $H$. To lighten notation, we use the shorthand $\underline{T}_t(\epsilon) = [\underline{T}_{t,1}(\epsilon), \cdots, \underline{T}_{t,p+1}(\epsilon)]^T$ for $\underline{T}(H_\epsilon\,;\underline{q}_t)$.
- We denote by $\mathrm{IF}_{\underline{q}_t}(\mathbf{z}_0\,;H,\underline{T})$ a sequence of influence functions derived for $\tau$-Lasso estimator when the non-smooth $\underline{q}(\underline{\boldsymbol{\theta}}\,;\underline{\lambda}_n)$ is replaced with a smooth $\underline{q}_t(\underline{\boldsymbol{\theta}}\,;\underline{\lambda}_n)$.
- $\underline{\Gamma}$ represents the set of non-zero elements in $\underline{\boldsymbol{\beta}}_\infty$.

Here, we outline the key steps taken in calculating the influence function of regularized estimators with non-smooth penalty.

- *Step 1:* We will approximate the non-smooth penalty term $\underline{q}$ with a sequence of smooth penalty functions $\underline{q}_t$ such that $\lim_{t\to\infty} \underline{q}_t = \underline{q}$.
- *Step 2:* We will then derive the influence function of the sequence of approximating estimators.
- *Step 3:* Finally, we recall the framework developed in [S14] under which the influence function of the regularized *M*-estimators with the non-smooth (non-differentiable) penalty $\underline{q}$ may be derived by limiting that of the regularized *M*-estimators with a smooth (differentiable) penalty $\underline{q}_t$ when $\lim_{t\to\infty} \underline{q}_t = \underline{q}$.

In the following, we will derive the influence function of $\tau$-Lasso estimator as sketched above.

### A. Proof of step 1

To begin, we shall first make use of the $\tau$-Lasso estimating equations expressed in terms of population quantities. As stated in step 1, we will substitute the $\ell_1$-norm penalty of equation (S2) with a sequence of smooth penalty functions. Recalling the proof of Proposition 2 of [S14], we can find a sequence of smooth functions converging to the $\ell_1$-norm penalty $\|\boldsymbol{\beta}\|_{\ell_1} = \sum_{j=1}^{p} |\underline{\beta}_j|$ by approximating the absolute value terms $|\underline{\beta}_j|$ with

$$l_t(\underline{\beta}_j) = \frac{2}{t}\log(e^{t\underline{\beta}_j} + 1) - \underline{\beta}_j \xrightarrow{t\to\infty} l(\underline{\beta}_j) = |\underline{\beta}_j|. \quad (S130)$$

whose first two derivatives w.r.t $\underline{\beta}_j$ are

$$l'_t(\underline{\beta}_j) = \frac{2e^{t\underline{\beta}_j}}{e^{t\underline{\beta}_j} + 1} - 1$$

$$\downarrow t \to \infty$$

$$l'(\underline{\beta}_j) = \operatorname{sgn}(\underline{\beta}_j) \quad (S131)$$

$$= \begin{cases} 1 & \text{if } \underline{\beta}_j > 0 \\ -1 & \text{if } \underline{\beta}_j < 0 \\ 0 & \text{otherwise} \end{cases},$$

$$l''_t(\underline{\beta}_j) = \frac{2t e^{t\underline{\beta}_j}}{(e^{t\underline{\beta}_j} + 1)^2}$$

$$\downarrow t \to \infty$$

$$l''(\underline{\beta}_j) = \begin{cases} 0 & \text{if } \underline{\beta}_j \neq 0 \\ +\infty & \text{otherwise} \end{cases}. \quad (S132)$$

By doing so, we get the following sequence of approximating estimators $\underline{T}(H;\underline{q}_t)$, defined as the root of

$$\mathbb{E}_H[\Psi(\mathbf{z},\underline{\boldsymbol{\theta}})] + \frac{\partial \underline{q}_t(\underline{\boldsymbol{\theta}}\,;\underline{\lambda}_n)}{\partial \underline{\boldsymbol{\theta}}} = \mathbf{0}_{p+1} \quad (S133)$$

with

$$\frac{\partial \underline{q}_t(\underline{\boldsymbol{\theta}}\,;\underline{\lambda}_n)}{\partial \underline{\boldsymbol{\theta}}} = \underline{\lambda}_n \begin{bmatrix} 0 \\ l'_t(\underline{\beta}_1) \\ \vdots \\ l'_t(\underline{\beta}_p) \end{bmatrix} \quad (S134)$$

such that $\lim_{t\to\infty} \underline{q}_t(\underline{\boldsymbol{\theta}}\,;\underline{\lambda}_n) = \underline{q}(\underline{\boldsymbol{\theta}}\,;\underline{\lambda}_n)$. The uniqueness of the root $\underline{T}(H;\underline{q}_t)$ in the estimating equation (S133) is guaranteed by the assumptions stated in our theorem, which are originally assumptions (A1)-(A3) for the one-stage regularized *M*-estimators [S14].

### B. Proof of step 2

We now derive the influence function of the sequence of approximating estimators $\underline{T}(H;\underline{q}_t)$. Returning to the definition of influence function given by equation (21) in the main body of the paper, we have

$$\mathrm{IF}_{\underline{q}_t}(\mathbf{z}_0\,;H,\underline{T}) = \frac{\partial \underline{T}(H_\epsilon\,;\underline{q}_t)}{\partial \epsilon}\bigg|_{\epsilon=0} \quad (S135)$$

where $H_\epsilon = (1-\epsilon)H + \epsilon\Delta_{\mathbf{z}_0}$ and the regularized *M*-functional $\underline{T}(H_\epsilon; \underline{q}_t)$ associated with the sequence of approximating estimators is defined as the root of

$$Q(\epsilon, \underline{\boldsymbol{\theta}}) := \mathbb{E}_{H_\epsilon}[\Psi(\mathbf{z}, \underline{\boldsymbol{\theta}})] + \frac{\partial \underline{q}_t(\underline{\boldsymbol{\theta}}; \underline{\lambda}_n)}{\partial \underline{\boldsymbol{\theta}}} = \mathbf{0}_{p+1} \quad (S136)$$

under the $\epsilon$-contamination of the underlying distribution $H$ by $\Delta_{\mathbf{z}_0}$ (a point mass with probability one at $\mathbf{z}_0$ and zero elsewhere). For the sake of convenience, we reiterate that the notation $\underline{T}_t(\epsilon)$ is a shorthand for $\underline{T}(H_\epsilon; \underline{q}_t)$. Substituting $H_\epsilon$ with $(1-\epsilon)H + \epsilon\Delta_{\mathbf{z}_0}$ in the above equation yields

$$\underbrace{Q(\epsilon, \underline{\boldsymbol{\theta}})}_{\in \mathbb{R}^{p+1}} := (1-\epsilon)\mathbb{E}_H[\Psi(\mathbf{z}, \underline{\boldsymbol{\theta}})] + \epsilon \mathbb{E}_{\Delta_{\mathbf{z}_0}}[\Psi(\mathbf{z}, \underline{\boldsymbol{\theta}})] + \frac{\partial \underline{q}_t(\underline{\boldsymbol{\theta}}; \underline{\lambda}_n)}{\partial \underline{\boldsymbol{\theta}}} \quad (S137)$$

To proceed, it suffices to take the derivative of the estimating equations given above w.r.t. $\epsilon$ and set $\epsilon$ to zero. Applying the implicit function theorem to the above equation, we obtain

$$\frac{\partial \underline{T}_t(\epsilon)}{\partial \epsilon}\Big|_{\epsilon=0} = -\Big[\frac{\partial Q(\epsilon, \underline{\boldsymbol{\theta}})}{\partial \underline{\boldsymbol{\theta}}}\Big|_{(0, \underline{T}_t(0))}\Big]^{-1} \times \frac{\partial Q(\epsilon, \underline{\boldsymbol{\theta}})}{\partial \epsilon}\Big|_{(0, \underline{T}_t(0))} \quad (S138)$$

where

$$\frac{\partial Q(\epsilon, \underline{\boldsymbol{\theta}})}{\partial \underline{\boldsymbol{\theta}}} = (1-\epsilon)\mathbb{E}_H\Big[\frac{\partial \Psi(\mathbf{z}, \underline{\boldsymbol{\theta}})}{\partial \underline{\boldsymbol{\theta}}}\Big] + \epsilon \mathbb{E}_{\Delta_{\mathbf{z}_0}}\Big[\frac{\partial \Psi(\mathbf{z}, \underline{\boldsymbol{\theta}})}{\partial \underline{\boldsymbol{\theta}}}\Big]$$
$$+ \frac{\partial^2 \underline{q}_t(\underline{\boldsymbol{\theta}}; \underline{\lambda}_n)}{\partial [\underline{\boldsymbol{\theta}}]^2}, \quad (S139a)$$

$$\frac{\partial Q(\epsilon, \underline{\boldsymbol{\theta}})}{\partial \epsilon} = -\mathbb{E}_H[\Psi(\mathbf{z}, \underline{\boldsymbol{\theta}})] + \mathbb{E}_{\Delta_{\mathbf{z}_0}}[\Psi(\mathbf{z}, \underline{\boldsymbol{\theta}})]. \quad (S139b)$$

In order to find $\partial Q(\epsilon, \underline{\boldsymbol{\theta}})/\partial \underline{\boldsymbol{\theta}}$, it is necessary to calculate the partial derivatives $\partial^2 \underline{q}_t(\underline{\boldsymbol{\theta}}; \underline{\lambda}_n)/\partial [\underline{\boldsymbol{\theta}}]^2$ and $\partial \Psi(\mathbf{z}, \underline{\boldsymbol{\theta}})/\partial \underline{\boldsymbol{\theta}}$. This can be done as follows:

*1) Finding $\partial^2 \underline{q}_t(\underline{\boldsymbol{\theta}}; \underline{\lambda}_n)/\partial [\underline{\boldsymbol{\theta}}]^2$:* We begin with determining $\partial^2 \underline{q}_t(\underline{\boldsymbol{\theta}}; \underline{\lambda}_n)/\partial [\underline{\boldsymbol{\theta}}]^2$, which leads to

$$\frac{\partial^2 \underline{q}_t(\underline{\boldsymbol{\theta}}; \underline{\lambda}_n)}{\partial [\underline{\boldsymbol{\theta}}]^2} = \underline{\lambda}_n \begin{bmatrix} 0 & 0 & \cdots & 0 \\ 0 & l_t''(\underline{\beta}_1) & \cdots & 0 \\ \vdots & \vdots & \ddots & \vdots \\ 0 & 0 & \cdots & l_t''(\underline{\beta}_p) \end{bmatrix}_{(p+1)\times(p+1)} \quad (S140)$$

where

$$l_t''(\underline{\beta}_j) = \frac{2te^{t\underline{\beta}_j}}{(e^{t\underline{\beta}_j}+1)^2} \xrightarrow{t \to \infty} \begin{cases} 0, & \text{if } \underline{\beta}_j \neq 0. \\ +\infty, & \text{otherwise.} \end{cases} \quad (S141)$$

*2) Finding $\partial \Psi(\mathbf{z}, \underline{\boldsymbol{\theta}})/\partial \underline{\boldsymbol{\theta}}$:* By performing some calculus, we arrive at

$$\frac{\partial \Psi(\mathbf{z}, \underline{\boldsymbol{\theta}})}{\partial \underline{\boldsymbol{\theta}}} = \underbrace{\begin{bmatrix} -\frac{1}{\underline{s}_\sigma}\psi_0(\tilde{r}(\underline{\boldsymbol{\theta}}))\tilde{r}(\underline{\boldsymbol{\theta}}) & -\frac{1}{\underline{s}_\sigma}\psi_0(\tilde{r}(\underline{\boldsymbol{\theta}}))\mathbf{x}^T \\ -\mathbf{x}(\underline{s}_\sigma\frac{\partial \psi(\tilde{r}(\underline{\boldsymbol{\theta}}))}{\partial \underline{s}_\sigma} + \psi(\tilde{r}(\underline{\boldsymbol{\theta}}))) & -\mathbf{x}\underline{s}_\sigma\frac{\partial \psi(\tilde{r}(\underline{\boldsymbol{\theta}}))}{\partial \underline{\boldsymbol{\beta}}} \end{bmatrix}}_{(p+1)\times(p+1)}. \quad (S142)$$

This in turn requires finding the derivatives $\partial \psi(\tilde{r}(\underline{\boldsymbol{\theta}}))/\partial \underline{\boldsymbol{\beta}}$ and $\partial \psi(\tilde{r}(\underline{\boldsymbol{\theta}}))/\partial \underline{s}_\sigma$ as follows:

$$\frac{\partial \psi(\tilde{r}(\underline{\boldsymbol{\theta}}))}{\partial \underline{\boldsymbol{\beta}}} = \frac{\partial W(\tilde{r}(\underline{\boldsymbol{\theta}}))}{\partial \underline{\boldsymbol{\beta}}}\psi_0(\tilde{r}(\underline{\boldsymbol{\theta}}))$$
$$+ W(\tilde{r}(\underline{\boldsymbol{\theta}}))\frac{\partial \psi_0(\tilde{r}(\underline{\boldsymbol{\theta}}))}{\partial \tilde{r}(\underline{\boldsymbol{\theta}})}\frac{\partial \tilde{r}(\underline{\boldsymbol{\theta}})}{\partial \underline{\boldsymbol{\beta}}} + \frac{\partial \psi_1(\tilde{r}(\underline{\boldsymbol{\theta}}))}{\partial \tilde{r}(\underline{\boldsymbol{\theta}})}\frac{\partial \tilde{r}(\underline{\boldsymbol{\theta}})}{\partial \underline{\boldsymbol{\beta}}} \quad (S143a)$$

$$\frac{\partial \psi(\tilde{r}(\underline{\boldsymbol{\theta}}))}{\partial \underline{s}_\sigma} = \frac{\partial W(\tilde{r}(\underline{\boldsymbol{\theta}}))}{\partial \underline{s}_\sigma}\psi_0(\tilde{r}(\underline{\boldsymbol{\theta}}))$$
$$+ W(\tilde{r}(\underline{\boldsymbol{\theta}}))\frac{\partial \psi_0(\tilde{r}(\underline{\boldsymbol{\theta}}))}{\partial \tilde{r}(\underline{\boldsymbol{\theta}})}\frac{\partial \tilde{r}(\underline{\boldsymbol{\theta}})}{\partial \underline{s}_\sigma} + \frac{\partial \psi_1(\tilde{r}(\underline{\boldsymbol{\theta}}))}{\partial \tilde{r}(\underline{\boldsymbol{\theta}})}\frac{\partial \tilde{r}(\underline{\boldsymbol{\theta}})}{\partial \underline{s}_\sigma} \quad (S143b)$$

On the other hand, finding the partial derivatives $\partial \psi(\tilde{r}(\underline{\boldsymbol{\theta}}))/\partial \underline{\boldsymbol{\beta}}$ and $\partial \psi(\tilde{r}(\underline{\boldsymbol{\theta}}))/\partial \underline{s}_\sigma$ involves deriving expressions for the partial derivatives $\partial \tilde{r}(\underline{\boldsymbol{\theta}})/\partial \underline{\boldsymbol{\beta}}$, $\partial \tilde{r}(\underline{\boldsymbol{\theta}})/\partial \underline{s}_\sigma$, $\partial W(\tilde{r}(\underline{\boldsymbol{\theta}}))/\partial \underline{\boldsymbol{\beta}}$ and $\partial W(\tilde{r}(\underline{\boldsymbol{\theta}}))/\partial \underline{s}_\sigma$. By straightforward calculations, we obtain $\partial \tilde{r}(\underline{\boldsymbol{\theta}})/\partial \underline{\boldsymbol{\beta}}$ and $\partial \tilde{r}(\underline{\boldsymbol{\theta}})/\partial \underline{s}_\sigma$ as follows:

$$\frac{\partial \tilde{r}(\underline{\boldsymbol{\theta}})}{\partial \underline{\boldsymbol{\beta}}} = -\frac{\mathbf{x}^T}{\underline{s}_\sigma} \quad (S144a)$$

$$\frac{\partial \tilde{r}(\underline{\boldsymbol{\theta}})}{\partial \underline{s}_\sigma} = -\frac{\tilde{r}(\underline{\boldsymbol{\theta}})}{\underline{s}_\sigma} \quad (S144b)$$

Likewise, we find the partial derivatives of $W(\tilde{r}(\underline{\boldsymbol{\theta}}))$ in equation (S127) with respect to $\underline{\boldsymbol{\beta}}$ as follows:

$$\frac{\partial W(\tilde{r}(\underline{\boldsymbol{\theta}}))}{\partial \underline{\boldsymbol{\beta}}}$$
$$= \Bigg[\frac{\mathbb{E}_H[-(2\psi_1(\tilde{r}(\underline{\boldsymbol{\theta}})) - \frac{\partial \psi_1(\tilde{r}(\underline{\boldsymbol{\theta}}))}{\partial \tilde{r}(\underline{\boldsymbol{\theta}})}\tilde{r}(\underline{\boldsymbol{\theta}}) - \psi_1(\tilde{r}(\underline{\boldsymbol{\theta}})))\frac{\mathbf{x}^T}{\underline{s}_\sigma}]}{\mathbb{E}_H[\psi_0(\tilde{r}(\underline{\boldsymbol{\theta}}))\tilde{r}(\underline{\boldsymbol{\theta}})]^2}$$
$$\times \mathbb{E}_H[\psi_0(\tilde{r}(\underline{\boldsymbol{\theta}}))\tilde{r}(\underline{\boldsymbol{\theta}})]\Bigg]$$
$$- \Bigg[\frac{\mathbb{E}_H[-(\frac{\partial \psi_0(\tilde{r}(\underline{\boldsymbol{\theta}}))}{\partial \tilde{r}(\underline{\boldsymbol{\theta}})}\tilde{r}(\underline{\boldsymbol{\theta}}) + \psi_0(\tilde{r}(\underline{\boldsymbol{\theta}})))\frac{\mathbf{x}^T}{\underline{s}_\sigma}]}{\mathbb{E}_H[\psi_0(\tilde{r}(\underline{\boldsymbol{\theta}}))\tilde{r}(\underline{\boldsymbol{\theta}})]^2}$$
$$\times \mathbb{E}_H[2\rho_1(\tilde{r}(\underline{\boldsymbol{\theta}})) - \psi_1(\tilde{r}(\underline{\boldsymbol{\theta}}))\tilde{r}(\underline{\boldsymbol{\theta}})]\Bigg]$$
$$= \frac{1}{\underline{s}_\sigma}\Bigg[-\frac{\mathbb{E}_H[(\psi_1(\tilde{r}(\underline{\boldsymbol{\theta}})) - \frac{\partial \psi_1(\tilde{r}(\underline{\boldsymbol{\theta}}))}{\partial \tilde{r}(\underline{\boldsymbol{\theta}})}\tilde{r}(\underline{\boldsymbol{\theta}}))\mathbf{x}^T]}{\mathbb{E}_H[\psi_0(\tilde{r}(\underline{\boldsymbol{\theta}}))\tilde{r}(\underline{\boldsymbol{\theta}})]}$$
$$+ W(\tilde{r}(\underline{\boldsymbol{\theta}}))\frac{\mathbb{E}_H[(\frac{\partial \psi_0(\tilde{r}(\underline{\boldsymbol{\theta}}))}{\partial \tilde{r}(\underline{\boldsymbol{\theta}})}\tilde{r}(\underline{\boldsymbol{\theta}}) + \psi_0(\tilde{r}(\underline{\boldsymbol{\theta}})))\mathbf{x}^T]}{\mathbb{E}_H[\psi_0(\tilde{r}(\underline{\boldsymbol{\theta}}))\tilde{r}(\underline{\boldsymbol{\theta}})]}\Bigg]. \quad (S145)$$

and

$$
\begin{aligned}
\frac{\partial W(\tilde{r}(\underline{\boldsymbol{\theta}}))}{\partial \underline{s}_\sigma} &= \left[\frac{\mathbb{E}_H[-(2\psi_1(\tilde{r}(\underline{\boldsymbol{\theta}})) - \frac{\partial \psi_1(\tilde{r}(\underline{\boldsymbol{\theta}}))}{\partial \tilde{r}(\underline{\boldsymbol{\theta}})}\tilde{r}(\underline{\boldsymbol{\theta}}) - \psi_1(\tilde{r}(\underline{\boldsymbol{\theta}})))\frac{\tilde{r}(\underline{\boldsymbol{\theta}})}{\underline{s}_\sigma}]}{\mathbb{E}_H[\psi_0(\tilde{r}(\underline{\boldsymbol{\theta}}))\tilde{r}(\underline{\boldsymbol{\theta}})]^2}\right.\\
&\quad \times \mathbb{E}_H[\psi_0(\tilde{r}(\underline{\boldsymbol{\theta}}))\tilde{r}(\underline{\boldsymbol{\theta}})]\Bigg] \\
&\quad - \left[\frac{\mathbb{E}_H[-(\frac{\partial \psi_0(\tilde{r}(\underline{\boldsymbol{\theta}}))}{\partial \tilde{r}(\underline{\boldsymbol{\theta}})}\tilde{r}(\underline{\boldsymbol{\theta}}) + \psi_0(\tilde{r}(\underline{\boldsymbol{\theta}})))\frac{\tilde{r}(\underline{\boldsymbol{\theta}})}{\underline{s}_\sigma}]}{\mathbb{E}_H[\psi_0(\tilde{r}(\underline{\boldsymbol{\theta}}))\tilde{r}(\underline{\boldsymbol{\theta}})]^2}\right. \\
&\quad \times \mathbb{E}_H[2\rho_1(\tilde{r}(\underline{\boldsymbol{\theta}})) - \psi_1(\tilde{r}(\underline{\boldsymbol{\theta}}))\tilde{r}(\underline{\boldsymbol{\theta}})]\Bigg] \\
&= \frac{1}{\underline{s}_\sigma}\left[-\frac{\mathbb{E}_H[(\psi_1(\tilde{r}(\underline{\boldsymbol{\theta}})) - \frac{\partial \psi_1(\tilde{r}(\underline{\boldsymbol{\theta}}))}{\partial \tilde{r}(\underline{\boldsymbol{\theta}})}\tilde{r}(\underline{\boldsymbol{\theta}}))\tilde{r}(\underline{\boldsymbol{\theta}})]}{\mathbb{E}_H[\psi_0(\tilde{r}(\underline{\boldsymbol{\theta}}))\tilde{r}(\underline{\boldsymbol{\theta}})]}\right.\\
&\quad + W(\tilde{r}(\underline{\boldsymbol{\theta}}))\frac{\mathbb{E}_H[(\psi_0(\tilde{r}(\underline{\boldsymbol{\theta}})) + \frac{\partial \psi_0(\tilde{r}(\underline{\boldsymbol{\theta}}))}{\partial \tilde{r}(\underline{\boldsymbol{\theta}})}\tilde{r}(\underline{\boldsymbol{\theta}}))\tilde{r}(\underline{\boldsymbol{\theta}})]}{\mathbb{E}_H[\psi_0(\tilde{r}(\underline{\boldsymbol{\theta}}))\tilde{r}(\underline{\boldsymbol{\theta}})]}\Bigg].
\end{aligned}
$$
(S146)

Let us now return to equation (S139). Having obtained all the necessary terms and expressions required to find the expression for $\partial Q(\epsilon, \underline{\boldsymbol{\theta}})/\partial \underline{\boldsymbol{\theta}}$, we will now evaluate $\partial Q(\epsilon, \underline{\boldsymbol{\theta}})/\partial \underline{\boldsymbol{\theta}}$ and $\partial Q(\epsilon, \underline{\boldsymbol{\theta}})/\partial \epsilon$ at $(0, \underline{T}_t(0))$ to find $\partial \underline{T}_t(\epsilon)/\partial \epsilon|_{\epsilon=0}$. We first evaluate $\partial Q(\epsilon, \underline{\boldsymbol{\theta}})/\partial \epsilon$ at $(0, \underline{T}_t(0))$

$$
\begin{aligned}
\frac{\partial Q(0, \underline{T}_t(0))}{\partial \epsilon} &= -\mathbb{E}_H[\Psi(\mathbf{z}, \underline{T}_t(0))] + \Psi(\mathbf{z}_0, \underline{T}_t(0)) \\
&\stackrel{(i)}{=} \underbrace{-\mathbb{E}_H[\Psi(\mathbf{z}, \underline{T}_t(0))] - \frac{\partial \underline{q}_t(\underline{\boldsymbol{\theta}}; \underline{\lambda}_n)}{\partial \underline{\boldsymbol{\theta}}}|_{\underline{\boldsymbol{\theta}}=\underline{T}_t(0)}}_{=\mathbf{0}_{p+1},\text{ by equation (S136)}} \\
&\quad + \Psi(\mathbf{z}_0, \underline{T}_t(0)) + \frac{\partial \underline{q}_t(\underline{\boldsymbol{\theta}}; \underline{\lambda}_n)}{\partial \underline{\boldsymbol{\theta}}}|_{\underline{\boldsymbol{\theta}}=\underline{T}_t(0)}
\end{aligned}
$$
(S147)

where the equality $(i)$ follows from adding and subtracting $\partial \underline{q}_t(\underline{\boldsymbol{\theta}}; \underline{\lambda}_n)/\partial \underline{\boldsymbol{\theta}}$ evaluated at $\underline{\boldsymbol{\theta}} = \underline{T}_t(0)$. As a consequence, we have

$$
\frac{\partial Q(0, \underline{T}_t(0))}{\partial \epsilon} = \Psi(\mathbf{z}_0, \underline{T}_t(0)) + \frac{\partial \underline{q}_t(\underline{\boldsymbol{\theta}}; \underline{\lambda}_n)}{\partial \underline{\boldsymbol{\theta}}}|_{\underline{\boldsymbol{\theta}}=\underline{T}_t(0)} \quad \text{(S148)}
$$

Next, we shall evaluate $\partial Q(\epsilon, \underline{\boldsymbol{\theta}})/\partial \underline{\boldsymbol{\theta}} \in \mathbb{R}^{(p+1)\times(p+1)}$ at $(0, \underline{T}_t(0))$ and then invert it via the block matrix inversion lemma.

$$
\frac{\partial Q(0, \underline{T}_t(0))}{\partial \underline{\boldsymbol{\theta}}} = \begin{bmatrix} \text{Block 1} & \text{Block 2} \\ \text{Block 3} & \text{Block 4} \end{bmatrix}_{(p+1)\times(p+1)} \quad \text{(S149)}
$$

where

$$
\text{Block 1} = \overbrace{-\frac{1}{\underline{T}_{t,1}(0)}\mathbb{E}_H[\psi_0(\tilde{r}(\underline{T}_t(0)))\tilde{r}(\underline{T}_t(0))]}^{\text{scalar}} \quad \text{(S150a)}
$$

$$
\text{Block 2} = \overbrace{-\frac{1}{\underline{T}_{t,1}(0)}\mathbb{E}_H[\psi_0(\tilde{r}(\underline{T}_t(0)))\mathbf{x}^T]}^{(1\times p)\text{ row vector}} \quad \text{(S150b)}
$$

$$
\text{Block 3} = \underbrace{-\mathbb{E}_H[\mathbf{x}(\underline{T}_{t,1}(0)\frac{\partial \psi(\tilde{r}(\underline{T}_t(0)))}{\partial \underline{s}_\sigma} + \psi(\tilde{r}(\underline{T}_t(0))))]}_{(p\times 1)\text{ column vector}}
$$
(S150c)

$$
\text{Block 4} = \underbrace{-\mathbb{E}_H[\mathbf{x}\underline{T}_{t,1}(0)\frac{\partial \psi(\tilde{r}(\underline{T}_t(0)))}{\partial \underline{\boldsymbol{\beta}}}] + \underline{\boldsymbol{\Phi}}_t}_{(p\times p)\text{ matrix}} \quad \text{(S150d)}
$$

$$
\underline{\boldsymbol{\Phi}}_t = \mathbf{diag}\big(\underline{\lambda}_n[l_t''(\underline{T}_{t,2}(0)), \cdots, l_t''(\underline{T}_{t,p+1}(0))]\big). \quad \text{(S150e)}
$$

With the above results in hand, we derive $\text{IF}_{\underline{q}_t}(\mathbf{z}_0; H, \underline{T})$.

### C. Proof of step 3

We now obtain the influence function of the $\tau$-Lasso estimator by limiting that of the sequence of approximating estimators derived in the preceding step. To do so, we need to find the limit of $[\partial Q(0, \underline{T}_t(0))/\partial \underline{\boldsymbol{\theta}}]^{-1}$ and $\partial Q(0, \underline{T}_t(0))/\partial \epsilon$ as $t$ goes to infinity.

*1) Deriving $\lim_{t\to\infty}[\frac{\partial Q(0, \underline{T}_t(0))}{\partial \underline{\boldsymbol{\theta}}}]^{-1}$:* By Lemma 2 of [S14], we know that $\lim_{t\to\infty} \underline{T}_t(0) = \underline{T}(H) = \underline{\boldsymbol{\theta}}_\infty$. In conjunction with equation (S149) and the matrix inversion lemma [S15], we obtain

$$
\lim_{t\to\infty}[\frac{\partial Q(0, \underline{T}_t(0))}{\partial \underline{\boldsymbol{\theta}}}]^{-1}
= \underbrace{\begin{bmatrix} M^{-1} & \mathbf{0}_{(\underline{k}_s+1)\times(p-\underline{k}_s)} \\ \mathbf{0}_{(p-\underline{k}_s)\times(\underline{k}_s+1)} & \mathbf{0}_{(p-\underline{k}_s)\times(p-\underline{k}_s)} \end{bmatrix}}_{(p+1)\times(p+1)\text{ matrix}} \quad \text{(S151)}
$$

where

$$
M = \begin{bmatrix} \overbrace{M_{11}}^{\text{scalar}} & \overbrace{M_{12}}^{(1\times \underline{k}_s)\text{ row vector}} \\ \underbrace{M_{21}}_{(\underline{k}_s\times 1)\text{ column vector}} & \underbrace{M_{22}}_{(\underline{k}_s\times \underline{k}_s)\text{ matrix}} \end{bmatrix}_{(\underline{k}_s+1)\times(\underline{k}_s+1)}
$$
(S152)

with

$$
M_{11} = -\frac{1}{\underline{s}_\infty}\mathbb{E}_H[\psi_0(\tilde{r}(\underline{T}(H)))\tilde{r}(\underline{T}(H))], \quad \text{(S153a)}
$$

$$
M_{12} = -\frac{1}{\underline{s}_\infty}\mathbb{E}_H[\psi_0(\tilde{r}(\underline{T}(H)))\mathbf{x}_{\underline{\Gamma}}^T], \quad \text{(S153b)}
$$

$$
M_{21} = -\mathbb{E}_H[(\underline{s}_\infty\frac{\partial \psi(\tilde{r}(\underline{T}(H)))}{\partial \underline{s}_\sigma} + \psi(\tilde{r}(\underline{T}(H))))\mathbf{x}_{\underline{\Gamma}}], \quad \text{(S153c)}
$$

and $M_{22}$ referring to a $\underline{k}_s \times \underline{k}_s$ submatrix of $-(\mathbb{E}_H[\mathbf{x}\underline{s}_\infty \partial \psi(\tilde{r}(\underline{T}(H)))/\partial \underline{\boldsymbol{\beta}}])$ indexed by the set $\underline{\Upsilon} = \{1, \cdots, \underline{k}_s\} \times \{1, \cdots, \underline{k}_s\}$. $\mathbf{x}_{\underline{\Gamma}}$ denotes a subvector of elements indexed by $\underline{\Gamma} = \{1, \cdots, \underline{k}_s\}$.

*2) Finding* $\lim_{t\to\infty} \frac{\partial Q(0,\underline{T}_t(0))}{\partial \epsilon}$: Lemma 2 of [S14] implies that $\lim_{t\to\infty} \underline{T}_t(0) = \underline{T}(H) = \underline{\boldsymbol{\theta}}_\infty$. Besides, $\lim_{t\to\infty} l''_t(\underline{\beta}_{j,\infty}) \to 0$ for $j = 1,\cdots,\underline{k}_s$ and $\lim_{t\to\infty} l''_t(\underline{\beta}_{-j,\infty}) \to \infty$ elsewhere. We then continue our analysis by finding the expression for $\lim_{t\to\infty} \partial Q(0,\underline{T}_t(0))/\partial \epsilon$

$$\lim_{t\to\infty} \frac{\partial Q(0,\underline{T}_t(0))}{\partial \epsilon} = \underbrace{\Psi(\mathbf{z}_0,\underline{T}(H)) + \frac{\partial q(\boldsymbol{\theta};\lambda_n)}{\partial \underline{\boldsymbol{\theta}}}|_{\underline{\boldsymbol{\theta}}=\underline{T}(H)}}_{(p+1)\times 1 \text{ column vector}}. \tag{S154}$$

We can thereby complete the proof by deriving the influence function of the $\tau$-Lasso estimator as in equation (26) within the main body of the paper.

*Remark S10*: Note that we may reuse the symbols $M$, $M_{22}$, and $Q$ for similar or different purposes, meaning they are merely local to this theorem.

## S.XII. PROOF OF THEOREM 7

Following a similar line of reasoning to the framework introduced in [S14], we will derive the influence function of the adaptive $\tau$-Lasso estimator by leveraging the tools developed for computing the influence function of two-stage regularized *M*-estimators and adapting them to the adaptive $\tau$-Lasso formulation. Before proceeding with the proofs, we will introduce all the necessary notations to clarify the proof and the derivation of the influence function.

- We write $q(\boldsymbol{\theta},\underline{T}(H);\lambda_n)$ and $q_t(\boldsymbol{\theta},\underline{T}(H);\lambda_n)$ to denote the regularization term of the adaptive $\tau$-Lasso estimator expressed in the standard form of two-stage regularized *M*-estimators and its smooth approximation, respectively.
- We use $T(H;q_t) = [T_1(H;q_t),\cdots,T_{p+1}(H;q_t)]^T$ to represent a sequence of approximating two-stage regularized *M*-estimators that is generated by replacing the non-smooth $q(\boldsymbol{\theta},\underline{T}(H);\lambda_n)$ with a smooth $q_t(\boldsymbol{\theta},\underline{T}(H);\lambda_n)$ when the distribution of $\mathbf{z}$ is $H$.
- Likewise, we denote by $T(H_\epsilon;q_t)$ the sequence of approximating two-stage regularized *M*-estimators obtained by perturbing $\mathbf{z}$ with $\epsilon$-contamination of the underlying distribution $H$. To simplify the notation, we use the shorthand $T_t(\epsilon) = [T_{t,1}(\epsilon),\cdots,T_{t,p+1}(\epsilon)]^T$ for $T(H_\epsilon;q_t)$.
- $\text{IF}_{q_t}(\mathbf{z}_0;H,T)$ represents a sequence of influence functions derived for the adaptive $\tau$-Lasso estimator when the non-smooth $q(\boldsymbol{\theta},\underline{T}(H);\lambda_n)$ is replaced with a smooth $q_t(\boldsymbol{\theta},\underline{T}(H);\lambda_n)$.
- $\Gamma$ denotes the set of non-zero elements in $\boldsymbol{\beta}_\infty$.

*Remark S11*: The regularized *M*-estimator formulation of the $\tau$-Lasso involves the notations $\underline{T}(H;\underline{q}_t)$, its $\epsilon$-contaminated version $\underline{T}(H_\epsilon;\underline{q}_t)$, the shorthand for the $\epsilon$-contaminated version $\underline{T}_t(\epsilon)$, $\text{IF}_{\underline{q}}(\mathbf{z};H,\underline{T})$ and its approximation $\text{IF}_{\underline{q}_t}(\mathbf{z}_0;H,\underline{T})$, which correspond to their respective counterparts in the adaptive $\tau$-Lasso estimator. We also defined these notations in Theorem 6 for deriving the influence function of the $\tau$-Lasso estimator.

We would now measure the amount of change in the estimator $\boldsymbol{\theta}_\infty = T(H)$, caused by infinitesimal contamination. Doing so requires three steps:

- *Step 1:* Substituting the non-smooth penalty term $q$ by a sequence of smooth functions $q_t$ such that $\lim_{t\to\infty} q_t = q$.
- *Step 2:* Finding the influence function of the surrogate estimator with smooth penalty.
- *Step 3:* Deriving the influence function of the adaptive $\tau$-Lasso estimator by taking the limit as $t \to \infty$ of the influence function of the surrogate estimator obtained in the preceding step.

In the subsequent lines, we guide the reader through the technical details of the proof in three steps.

### A. Proof of step 1

Let us now revisit equation (S124). As stated above in step 1, we shall replace the non-smooth penalty function $q(\boldsymbol{\theta},\underline{T}(H);\lambda_n)$ in equation (S124), as defined by

$$q(\boldsymbol{\theta},\underline{T}(H);\lambda_n) = \lambda_n \sum_{j=1}^{p} \frac{|\beta_j|}{|\underline{\beta}_{j,\infty}|} \tag{S155}$$

with

$$q_t(\boldsymbol{\theta},\underline{T}(H);\lambda_n) = \lambda_n \sum_{j=1}^{p} \frac{l_t(\beta_j)}{l_t(\underline{\beta}_{j,\infty})}, \tag{S156}$$

such that

$$\lim_{t\to\infty} q_t(\boldsymbol{\theta},\underline{T}(H);\lambda_n) \to q(\boldsymbol{\theta},\underline{T}(H);\lambda_n). \tag{S157}$$

Doing so yields a sequence of approximating estimators $T(H;q_t)$ satisfying the following system of estimating equations:

$$\mathbb{E}_H[\Psi(\mathbf{z},T(H;q_t))] + \frac{\partial q_t(\boldsymbol{\theta},\underline{T}(H;\underline{q}_t);\lambda_n)}{\partial \boldsymbol{\theta}}\bigg|_{\boldsymbol{\theta}=T(H;q_t)} = \mathbf{0}_{p+1}. \tag{S158}$$

where

$$\frac{\partial q_t(\boldsymbol{\theta},\underline{T}(H;\underline{q}_t);\lambda_n)}{\partial \boldsymbol{\theta}}\bigg|_{\boldsymbol{\theta}=T(H;q_t)} = \begin{bmatrix} 0 \\ \lambda_n \frac{l'_t(T_2(H;q_t))}{l_t(\underline{T}_2(H;\underline{q}_t))} \\ \vdots \\ \lambda_n \frac{l'_t(T_{p+1}(H;q_t))}{l_t(\underline{T}_{p+1}(H;\underline{q}_t))} \end{bmatrix}. \tag{S159}$$

### B. Proof of step 2

We will then obtain the influence function of the approximating estimators described above. Alternatively, we shall measure the influence of contamination on the value of the functional $T(H;q_t)$. Formally speaking, we take the derivative of the functional $T(H_\epsilon;q_t)$ w.r.t. $\epsilon$ and set $\epsilon = 0$. In order to do so, we shall first define $T(H_\epsilon;q_t)$, an estimating functional that satisfies

$$\mathbb{E}_{H_\epsilon}[\Psi(\mathbf{z},T(H_\epsilon;q_t))]$$
$$+ \frac{\partial q_t(\boldsymbol{\theta},\underline{T}(H_\epsilon;\underline{q}_t);\lambda_n)}{\partial \boldsymbol{\theta}}\bigg|_{\boldsymbol{\theta}=T(H_\epsilon;q_t)} = \mathbf{0}_{p+1}, \tag{S160}$$

resulting from the $\epsilon$-contamination of $H$ by the distribution $\Delta_{\mathbf{z}_0}$ as given by $H_\epsilon = (1-\epsilon)H + \epsilon\Delta_{\mathbf{z}_0}$. Substituting $H_\epsilon = (1-\epsilon)H + \epsilon\Delta_{\mathbf{z}_0}$ into the above equation leads to

$$(1-\epsilon)\mathbb{E}_H\left[\Psi(\mathbf{z}, T(H_\epsilon; q_t))\right] + \epsilon\mathbb{E}_{\Delta_{\mathbf{z}_0}}\left[\Psi(\mathbf{z}, T(H_\epsilon; q_t))\right]$$
$$+ \frac{\partial q_t(\boldsymbol{\theta}, \underline{T}(H_\epsilon;\underline{q}_t)\,;\lambda_n)}{\partial \boldsymbol{\theta}}|_{\boldsymbol{\theta}=T(H_\epsilon;q_t)} = \mathbf{0}_{p+1}. \tag{S161}$$

To streamline the notation, we adopt the shorthand $q_t'(T(H_\epsilon;q_t), \underline{T}(H_\epsilon;\underline{q}_t)\,;\lambda_n)$ for

$$\frac{\partial q_t(\boldsymbol{\theta}, \underline{T}(H_\epsilon;\underline{q}_t)\,;\lambda_n)}{\partial \boldsymbol{\theta}}|_{\boldsymbol{\theta}=T(H_\epsilon;q_t)}. \tag{S162}$$

To proceed with the derivation of the influence function of the surrogate estimator, we first rewrite the expectations within equation (S161) as integrals. We will then take the derivative of the resulting equation w.r.t. $\epsilon$ and evaluate it at $\epsilon = 0$ as follows:

$$\underbrace{\frac{\partial}{\partial \epsilon}\left[(1-\epsilon)\int \Psi(\mathbf{z}, T(H_\epsilon; q_t))dH\right]_{\epsilon=0}}_{\text{first term}}$$
$$+ \underbrace{\frac{\partial}{\partial \epsilon}\left[\epsilon \int \Psi(\mathbf{z}, T(H_\epsilon; q_t))d\Delta_{\mathbf{z}_0}\right]_{\epsilon=0}}_{\text{second term}}$$
$$+ \underbrace{\frac{\partial}{\partial \epsilon}\left[q_t'(T(H_\epsilon;q_t), \underline{T}(H_\epsilon;\underline{q}_t)\,;\lambda_n)\right]_{\epsilon=0}}_{\text{third term}} = \mathbf{0}_{p+1}. \tag{S163}$$

To simplify matters, we will differentiate each term separately as follows:

*1) Finding the first term:* By the chain rule, equation (S158), and appropriately adding and subtracting terms, we find that

$$\frac{\partial}{\partial \epsilon}\left[(1-\epsilon)\int \Psi(\mathbf{z}, T(H_\epsilon; q_t))dH\right]_{\epsilon=0}$$
$$= \left(\int \Psi'(\mathbf{z}, T(H; q_t))dH\right) \times \frac{\partial T(H_\epsilon; q_t)}{\partial \epsilon}|_{\epsilon=0}$$
$$+ q_t'(T(H;q_t), \underline{T}(H;\underline{q}_t)\,;\lambda_n)$$
$$\underbrace{- \int \Psi(\mathbf{z}, T(H; q_t))dH - q_t'(T(H;q_t), \underline{T}(H;\underline{q}_t)\,;\lambda_n)}_{=\mathbf{0}_{p+1},\text{ by equation (S158)}}$$
$$= \left(\int \Psi'(\mathbf{z}, T(H; q_t))dH\right) \times \frac{\partial T(H_\epsilon; q_t)}{\partial \epsilon}|_{\epsilon=0}$$
$$+ q_t'(T(H;q_t), \underline{T}(H;\underline{q}_t)\,;\lambda_n). \tag{S164}$$

Note that $\Psi'(\mathbf{z}, T(H; q_t))$ stands for $\partial \Psi(\mathbf{z}, T(H_\epsilon; q_t))/\partial T(H_\epsilon; q_t))$ evaluated at $\epsilon = 0$.

*2) Finding the second term:* By taking derivatives and some simple algebra, we obtain

$$\frac{\partial}{\partial \epsilon}\left[\epsilon \int \Psi(\mathbf{z}, T(H_\epsilon; q_t))d\Delta_{\mathbf{z}_0}\right]_{\epsilon=0} = \Psi(\mathbf{z}_0, T(H; q_t)). \tag{S165}$$

*3) Finding the third term:* Before we proceed, let us recall that $T_t(\epsilon)$ and $\underline{T}_t(\epsilon)$ are shorthands for $T(H_\epsilon; q_t)$ and $\underline{T}(H_\epsilon;\underline{q}_t)$, respectively. Returning now to the third term of equation (S163), we will reduce the notational overhead by replacing $T(H_\epsilon; q_t)$ with $T_t(\epsilon)$ and $\underline{T}(H_\epsilon;\underline{q}_t)$ with $\underline{T}_t(\epsilon)$. By the chain rule and some simple algebra, we then obtain the following expressions

$$\frac{\partial}{\partial \epsilon}\left[q_t'(T_t(\epsilon), \underline{T}_t(\epsilon)\,;\lambda_n)\right]_{\epsilon=0}$$
$$= \frac{\partial q_t'(T_t(\epsilon), \underline{T}_t(\epsilon)\,;\lambda_n)}{\partial T_t(\epsilon)}|_{\epsilon=0} \times \frac{\partial T_t(\epsilon)}{\partial \epsilon}|_{\epsilon=0}$$
$$+ \frac{\partial q_t'(T_t(\epsilon), \underline{T}_t(\epsilon)\,;\lambda_n)}{\partial \underline{T}_t(\epsilon)}|_{\epsilon=0} \times \frac{\partial \underline{T}_t(\epsilon)}{\partial \epsilon}|_{\epsilon=0}$$
$$= \frac{\partial}{\partial \epsilon}\left(\begin{bmatrix} 0 \\ \lambda_n \frac{l_t'(\underline{T}_{t,2}(\epsilon))}{l_t(\underline{T}_{t,2}(\epsilon))} \\ \vdots \\ \lambda_n \frac{l_t'(\underline{T}_{t,p+1}(\epsilon))}{l_t(\underline{T}_{t,p+1}(\epsilon))} \end{bmatrix}\right)_{\epsilon=0}$$
$$= \lambda_n\left(\begin{bmatrix} 0 & 0 & \cdots & 0 \\ 0 & \frac{l_t''(\underline{T}_{t,2}(0))}{l_t(\underline{T}_{t,2}(0))} & \cdots & 0 \\ \vdots & \vdots & \ddots & 0 \\ 0 & 0 & \cdots & \frac{l_t''(\underline{T}_{t,p+1}(0))}{l_t(\underline{T}_{t,p+1}(0))} \end{bmatrix}\right.$$
$$\times \frac{\partial \underline{T}_t(\epsilon)}{\partial \epsilon}|_{\epsilon=0}$$
$$- \begin{bmatrix} 0 & 0 & \cdots & 0 \\ 0 & \frac{l_t'(\underline{T}_{t,2}(0))l_t'(\underline{T}_{t,2}(0))}{(l_t(\underline{T}_{t,2}(0)))^2} & \cdots & 0 \\ \vdots & \vdots & \ddots & 0 \\ 0 & 0 & \cdots & \frac{l_t'(\underline{T}_{t,p+1}(0))l_t'(\underline{T}_{t,p+1}(0))}{(l_t(\underline{T}_{t,p+1}(0)))^2} \end{bmatrix}$$
$$\left.\times \frac{\partial \underline{T}_t(\epsilon)}{\partial \epsilon}|_{\epsilon=0}\right). \tag{S166}$$

Recalling the definition of influence function by equation (21) in the main body of the paper, we can derive the ultimate formula for the third term by plugging in the following expressions into equation (S166).

$$\text{IF}_{q_t}(\mathbf{z}_0\,;H,T) = \frac{\partial T_t(\epsilon)}{\partial \epsilon}|_{\epsilon=0},$$
$$\text{IF}_{\underline{q}_t}(\mathbf{z}_0\,;H,\underline{T}) = \frac{\partial \underline{T}_t(\epsilon)}{\partial \epsilon}|_{\epsilon=0}. \tag{S167}$$

Having computed all three terms, we can now write expression for the influence function of the surrogate estimator for the adaptive $\tau$-Lasso estimator and thereby complete the proof of step 2. Substituting the computed terms back into equation (S163), isolating terms involving $\text{IF}_{q_t}(\mathbf{z}_0\,;H,T)$, and moving

the remaining terms to the right-hand side yields

$$\left(\int \Psi^{'}(\mathbf{z}, T(H;q_t)) dH \right.$$

$$+ \underbrace{\begin{bmatrix} 0 & 0 & \cdots & 0 \\ 0 & \lambda_n \frac{l_t^{''}(T_{t,2}(0))}{l_t(\underline{T}_{t,2}(0))} & \cdots & 0 \\ \vdots & \vdots & \ddots & 0 \\ 0 & 0 & \cdots & \lambda_n \frac{l_t^{''}(T_{t,p+1}(0))}{l_t(\underline{T}_{t,p+1}(0))} \end{bmatrix}}_{\mathbf{\Lambda}_t} \Bigg)$$

$$\times \mathrm{IF}_{q_t}(\mathbf{z}_0\ ; H, T)$$

$$= -\Big( \Psi(\mathbf{z}_0, T(H;q_t)) + q^{'}_t(T(H;q_t), \underline{T}(H;\underline{q}_t)\ ; \lambda_n)$$

$$- \underbrace{\begin{bmatrix} 0 & 0 & \cdots & 0 \\ 0 & \lambda_n \frac{l_t^{'}(T_{t,2}(0)) l_t^{'}(\underline{T}_{t,2}(0))}{(l_t(\underline{T}_{t,2}(0)))^2} & \cdots & 0 \\ \vdots & \vdots & \ddots & 0 \\ 0 & 0 & \cdots & \lambda_n \frac{l_t^{'}(T_{t,p+1}(0)) l_t^{'}(\underline{T}_{t,p+1}(0))}{(l_t(\underline{T}_{t,p+1}(0)))^2} \end{bmatrix}}_{\mathbf{\Phi}_t}$$

$$\times \mathrm{IF}_{\underline{q}_t}(\mathbf{z}_0\ ; H, \underline{T}) \Big) \quad (\mathrm{S}168)$$

and hence conclude that

$$\mathrm{IF}_{q_t}(\mathbf{z}_0\ ; H, T) = -\Big( \mathbb{E}_H[\Psi^{'}(\mathbf{z}, T(H;q_t))] + \mathbf{\Lambda}_t \Big)^{-1}$$
$$\times \Big( \Psi(\mathbf{z}_0, T(H;q_t)) + q^{'}_t(T(H;q_t), \underline{T}(H;\underline{q}_t)\ ; \lambda_n)$$
$$- \mathbf{\Phi}_t \times \mathrm{IF}_{\underline{q}_t}(\mathbf{z}_0\ ; H, \underline{T}) \Big). \quad (\mathrm{S}169)$$

### C. Proof of step 3

In order to complete the proof, it only remains to take the limit of the influence function $\mathrm{IF}_{q_t}(\mathbf{z}_0\ ; H, T)$ as $t \to \infty$. To do so, we shall determine the limits of its constituent components $T(H;q_t)$, $\underline{T}(H;\underline{q}_t)$, $\mathbf{\Phi}_t$, $\mathbf{\Lambda}_t$, $\Psi(\mathbf{z}_0, T(H;q_t))$, $\Psi^{'}(\mathbf{z}, T(H;q_t))$, $q^{'}_t(T(H;q_t), \underline{T}(H;\underline{q}_t)\ ; \lambda_n)$ and $\mathrm{IF}_{\underline{q}_t}(\mathbf{z}_0\ ; H, \underline{T})$ as $t$ approaches infinity, as outlined below.

*1) Taking* $\lim_{t\to\infty} T(H;q_t)$: By Lemma 4 of [S14], we know that $\lim_{t\to\infty} T(H;q_t) = T(H) = \boldsymbol{\theta}_\infty$.

*2) Finding* $\lim_{t\to\infty} \underline{T}(H;\underline{q}_t)$: Likewise, using Lemma 2 of [S14], we have $\lim_{t\to\infty} \underline{T}(H;\underline{q}_t) = \underline{T}(H) = \underline{\boldsymbol{\theta}}_\infty$.

*3) Taking* $\lim_{t\to\infty} \mathbf{\Phi}_t$: By $\lim_{t\to\infty} T(H;q_t) = T(H)$, $\lim_{t\to\infty} \underline{T}(H;\underline{q}_t) = \underline{T}(H)$ and relations of equations (S130) and (S131), we get

$$\lim_{t\to\infty} \frac{l_t^{'}(T_{t,j}(0)) l_t^{'}(\underline{T}_{t,j}(0))}{(l_t(\underline{T}_{t,j}(0)))^2}$$
$$= \begin{cases} \frac{\mathrm{sgn}(T_{t,j}(0))\mathrm{sgn}(\underline{T}_{t,j}(0))}{|\underline{T}_{t,j}(0)|^2} & \text{for } j = 2,\cdots, \underline{k}_s+1 \\ 0 & \text{elsewhere} \end{cases}. \quad (\mathrm{S}170)$$

Combining this with the definition of $\mathbf{\Phi}_t$ given by equation (S168), we find that $\lim_{t\to\infty} \mathbf{\Phi}_t = \underbrace{\mathbf{diag}\Big(\mathbf{\Phi}, \mathbf{0}_{(p-\underline{k}_s)}\Big)}_{(p+1)\times(p+1) \text{ matrix}}$ where

$$\mathbf{\Phi} = \underbrace{\begin{bmatrix} 0 \\ \lambda_n \frac{\mathrm{sgn}(\beta_{1,\infty})\mathrm{sgn}(\underline{\beta}_{1,\infty})}{|\underline{\beta}_{1,\infty}|^2} \\ \vdots \\ \lambda_n \frac{\mathrm{sgn}(\beta_{\underline{k}_s,\infty})\mathrm{sgn}(\underline{\beta}_{\underline{k}_s,\infty})}{|\underline{\beta}_{\underline{k}_s,\infty}|^2} \end{bmatrix}}_{(\underline{k}_s+1)\times 1 \text{ column vector}}. \quad (\mathrm{S}171)$$

*4) Finding* $\lim_{t\to\infty} \mathbf{\Lambda}_t$: Using $\lim_{t\to\infty} T(H;q_t) = T(H)$, $\lim_{t\to\infty} \underline{T}(H;\underline{q}_t) = \underline{T}(H)$ and relations of equations (S130) and (S132), we obtain

$$\lim_{t\to\infty} \frac{l_t^{''}(T_{t,j}(0))}{l_t(\underline{T}_{t,j}(0))} = \begin{cases} 0, & \text{for } j = 1,\cdots, \underline{k}_s. \\ +\infty, & \text{elsewhere.} \end{cases} \quad (\mathrm{S}172)$$

This result when combined with the definition of $\mathbf{\Lambda}_t$ given by equation (S168) yields $\lim_{t\to\infty} \mathbf{\Lambda}_t = \mathbf{\Lambda}$ where

$$\mathbf{\Lambda} = \begin{bmatrix} \mathbf{0}_{(k_s+1)\times(k_s+1)} & \mathbf{0}_{(k_s+1)\times(p-k_s)} \\ \mathbf{0}_{(p-k_s)\times(k_s+1)} & +\infty \times \mathbf{I}_{p-k_s} \end{bmatrix}. \quad (\mathrm{S}173)$$

*5) Taking* $\lim_{t\to\infty} \Psi(\mathbf{z}_0, T(H;q_t))$: By the composition theorem for limits [S16] and $\lim_{t\to\infty} T(H;q_t) = T(H)$, we conclude that $\lim_{t\to\infty} \Psi(\mathbf{z}_0, T(H;q_t)) = \Psi(\mathbf{z}_0, \lim_{t\to\infty} T(H;q_t)) = \Psi(\mathbf{z}_0, T(H))$.

*6) Finding* $\lim_{t\to\infty} \Psi^{'}(\mathbf{z}, T(H;q_t))$: Applying the composition theorem for limits and using $\lim_{t\to\infty} T(H;q_t) = T(H)$, we get $\lim_{t\to\infty} \Psi^{'}(\mathbf{z}, T(H;q_t)) = \Psi^{'}(\mathbf{z}, \lim_{t\to\infty} T(H;q_t)) = \Psi^{'}(\mathbf{z}, T(H))$.

*7) Finding* $\lim_{t\to\infty} q^{'}_t(T(H;q_t), \underline{T}(H;\underline{q}_t)\ ; \lambda_n)$: Using the composition theorem for limits, $\lim_{t\to\infty} T(H;q_t) = T(H)$, $\lim_{t\to\infty} \underline{T}(H;\underline{q}_t) = \underline{T}(H)$ and relations of equations (S130) and (S131), we find that

$$\lim_{t\to\infty} q^{'}_t(T(H;q_t), \underline{T}(H;\underline{q}_t)\ ; \lambda_n) = q^{'}(T(H), \underline{T}(H)\ ; \lambda_n)$$
$$= \begin{bmatrix} 0 \\ \lambda_n \frac{\mathrm{sgn}(\beta_{1,\infty})}{|\underline{\beta}_{1,\infty}|} \\ \vdots \\ \lambda_n \frac{\mathrm{sgn}(\beta_{p,\infty})}{|\underline{\beta}_{p,\infty}|} \end{bmatrix}. \quad (\mathrm{S}174)$$

*8) Finding* $\lim_{t\to\infty} \mathrm{IF}_{\underline{q}_t}(\mathbf{z}_0\ ; H, \underline{T})$: By the proof of theorem 6, quantifying the influence function of $\tau$-Lasso estimator, we have $\lim_{t\to\infty} \mathrm{IF}_{\underline{q}_t}(\mathbf{z}_0\ ; H, \underline{T}) = \mathrm{IF}(\mathbf{z}_0\ ; H, \underline{T})$.

Putting together the pieces, we find the limiting expression for $\text{IF}_{q_t}(\mathbf{z}_0\,;H,T)$, that is,

$$\begin{aligned}\text{IF}(\mathbf{z}_0\,;H,T) &= \lim_{t\to\infty}\text{IF}_{q_t}(\mathbf{z}_0\,;H,T) \\ &= -\underbrace{\begin{bmatrix} N^{-1} & \mathbf{0}_{(k_s+1)\times(p-k_s)} \\ \mathbf{0}_{(p-k_s)\times(k_s+1)} & \mathbf{0}_{(p-k_s)\times(p-k_s)} \end{bmatrix}}_{(p+1)\times(p+1)\text{ matrix}} \\ &\quad \times \Bigg(\underbrace{\Psi(\mathbf{z}_0,T(H)) + q^{'}(T(H),\underline{T}(H)\,;\lambda_n)}_{(p+1)\times 1\text{ column vector}} \\ &\quad - \underbrace{\mathbf{diag}\Big(\boldsymbol{\Phi},\mathbf{0}_{(p-\underline{k}_s)}\Big)}_{(p+1)\times(p+1)\text{ matrix}} \times \underbrace{\text{IF}(\mathbf{z}_0\,;H,\underline{T})}_{(p+1)\times 1\text{ column vector}}\Bigg).\end{aligned}$$
(S175)

We can obtain the submatrix $N$ in equation (S175) by modifying the matrix $M$ defined by equation (S152). First, each occurrence of $\underline{T}(H)$, $\partial\underline{\boldsymbol{\beta}}$, $\partial\underline{s}_\sigma$, $\underline{\Gamma}$, and $\underline{s}_\infty$ is replaced with $T(H)$, $\partial\boldsymbol{\beta}$, $\partial s_\sigma$, $\Gamma$, and $s_\infty$, respectively. Next, we fix the matrix dimension to $k_s+1 \times k_s+1$. Thus, we have derived the influence function of the adaptive $\tau$-Lasso estimator, given in equation (29) within the main body of the paper, thereby completing the proof.

*Remark S12*: Notice that the set of regression coefficients estimated to be zero in the $\tau$-Lasso estimator will not appear in $\boldsymbol{\beta}_\infty$.



\end{document}